\documentclass{article}

\usepackage[preprint]{neurips2025}

\usepackage[utf8]{inputenc} % allow utf-8 input
\usepackage[T1]{fontenc}    % use 8-bit T1 fonts
\usepackage{hyperref}       % hyperlinks
\usepackage{url}            % simple URL typesetting
\usepackage{booktabs}       % professional-quality tables
\usepackage{amsfonts}       % blackboard math symbols
\usepackage{nicefrac}       % compact symbols for 1/2, etc.
\usepackage{microtype}      % microtypography
\usepackage{arydshln}

\usepackage{microtype}
\usepackage{graphicx}
\usepackage{amsmath}
\usepackage{amssymb}
\usepackage{mathtools}
\usepackage{amsthm}

\usepackage[capitalize,noabbrev]{cleveref}

\usepackage{bm}
\usepackage{amsthm}
\usepackage{amssymb}
\usepackage{amsmath}
\newtheorem{lemma}{Lemma}
\newtheorem{theorem}{Theorem}
\newtheorem{definition}{Definition}

\usepackage{booktabs}
\usepackage{multirow}
\usepackage{tabularx}
\usepackage{subcaption}

\usepackage{pgfplots}
\usepackage{graphicx}
\usepgfplotslibrary{fillbetween}
\pgfplotsset{compat=1.18}

\definecolor{softred}{RGB}{250,100,100}
\definecolor{softgreen}{RGB}{56,118,29}
\definecolor{softblue}{RGB}{100,150,200}
\definecolor{softgray}{RGB}{150,150,150}
\newcommand{\textred}[1]{{\color{softred}#1}}
\newcommand{\textblue}[1]{{\color{softblue}#1}}

\newcommand{\textgreen}[1]{{\color{softgreen}#1}}
\usepackage{xspace}
\newcommand{\ourmethod}{\textsc{GradS}\xspace}

\title{Multi-Layer Attention is the Amplifier of Demonstration Effectiveness}

\author{
    \textbf{Dingzirui Wang$^{1}$ \quad Xuanliang Zhang$^{1}$ \quad Keyan Xu$^{1}$ \quad Qingfu Zhu$^{1}$} \\ \textbf{Wanxiang Che$^{1}$ \quad Yang Deng$^{2}$}\\
    $^{1}$Harbin Institute of Technology \quad $^{2}$Singapore Management University \\
    \texttt{\{dzrwang, xuanliangzhang, kyxu, qfzhu, car\}@ir.hit.edu.cn} \\
    \texttt{ydeng@smu.edu.sg}
}

\usepackage{bibentry}
\begin{document}

    \maketitle
    
    \begin{abstract}
        % 当前，有许多工作研究了ICL有效的机理，来启发相关的方法设计
        Numerous studies have investigated the underlying mechanisms of in-context learning (ICL) effectiveness to inspire the design of related methods. 
        % 然而，现有的工作都基于ICL中提供的示例有效，但许多工作表明，并非所有的示例都有效，导致ICL无法带来性能提升
        However, existing work predominantly assumes the effectiveness of the demonstrations provided within ICL, while many research indicates that not all demonstrations are effective, failing to yielding any performance improvement during ICL.
        % 因此在本文中，我们探讨了什么原因会导致示例无效，以及模型为什么会忽视无效的示例
        Therefore, in this paper, we investigate the reasons behind demonstration ineffectiveness. 
        % 我们基于梯度流和线性子注意力模型开展研究
        Our analysis is based on gradient flow and linear self-attention models. 
        % 我们令梯度流为零，推导得出了示例无效是因为其包含的信息已经被模型学过，或者和用户查询不相关
        By setting the gradient flow to zero, we deduce that a demonstration becomes ineffective if its information has either been learned by the model or is irrelevant to the user query.
        % 然后我们证明了，对于多层模型，随着层数增大，示例间的有效性也在不断放大，从而模型会去忽略无效的示例而更关注有效的示例
        Furthermore, we demonstrate that in multi-layer models, the disparity in effectiveness among demonstrations is amplified with layer increasing, causing the model to focus more on effective ones.
        % 然后，考虑到现有的示例选取方法主要关注示例和用户查询间的相似性而忽略了模型是否学习了示例中的信息，我们提出了\ourmethod，基于梯度流来进行示例选取
        Considering that current demonstration selection methods primarily focus on the relevance to the user query while overlooking the information that the model has already assimilated, we propose a novel method called \ourmethod, which leverages gradient flow for demonstration selection. 
        % 我们将示例对给定用户查询的梯度流大小作为选取依据，确保选取的示例的有效性
        We use the magnitude of the gradient flow of the demonstration with respect to a given user query as the criterion, thereby ensuring the effectiveness of the chosen ones.
        % 我们在四个主流LLMs和五个主流数据集上对我们的结论和方法进行了验证
        We validate our derivation and \ourmethod on four prominent LLMs across five mainstream datasets. 
        % 实验结果表明，随着模型层数增大，示例间有效性的差异也被不断放大，证明了我们的推导结论
        The experimental results confirm that the disparity in effectiveness among demonstrations is magnified as the model layer increases, substantiating our derivations. 
        % 此外，相较于最好的aselines，\ourmethod带来了平均1.0%的性能提升，证明了我们方法的有效性
        Moreover, \ourmethod achieves a relative improvement of $6.8\%$ on average over the strongest baselines, demonstrating its effectiveness.
    \end{abstract}

    \section{Introduction}
        % In-Context Learning (ICL)是一种有效增强大模型性能的方法
In-Context Learning (ICL) is an effective method for enhancing the performance of Large Language Models (LLMs), being widely adapted to various tasks \cite{brown-2020-gpt3,dong-etal-2024-icl-survey}. 
% 通过在输入中提供用户查询相关的示例，来引导模型生成正确的推理过程，从而提高推理性能
By providing demonstrations relevant to the user query in the input, it guides the reasoning of LLMs, thereby improving inference performance. 
% 有许多前人工作研究了ICL的内部机理，来解释大模型如何获得ICL的能力，并指导ICL方法的设计
Recent years have witnessed many efforts on investigating the internal mechanisms of ICL to explain how LLMs acquire this ability for guiding the design of ICL methods \cite{zhou-etal-2024-mechanism-survey}. 
% 例如，\citep{}讨论了ICL的收敛性与收敛速度，而\citep{}主要关注执行ICL时模型内部不同结构的功能
For example, recent works \cite{zhang-etal-2024-trained,mahankali2024one,smart2025incontext} discuss the convergence and convergence speed of ICL, while some other works \cite{olsson2022incontextlearninginductionheads,li2024how,chen2024how} study the function of each module of Transformer~\cite{vaswani-etal-2017-transformer} during ICL.

% 而随着基座模型性能的不断增强，模型表现出一种现象，即存在ineffective的示例无法在ICL时增强推理性能
As the performance of LLMs continues to improve, a phenomenon emerges that there exist ineffective demonstrations, which cannot enhance reasoning performance during ICL \cite{deepseekai2025deepseekr1,wang2025lcs}.
% 但现有的ICL机理研究都基于给定的示例有效的情况开展研究，限制了探索如何增强ICL的性能
However, existing research on the mechanisms of ICL is predominantly based on the assumption that the given demonstrations are effective, which limits the exploration of how to enhance the performance of ICL.
% 因此在本文中，我们主要讨论：
Therefore, in this paper, we aim to answer the following research questions (RQs):
% 1. 什么会导致demonstration变得无效
\textit{1) What makes the demonstration ineffective?}
% 2. LLM是如何处理示例的有效性的
\textit{2) How LLMs handle the demonstration effectiveness?}
% 3. 如何选取大模型不会忽略的示例，来增强ICL性能
and \textit{3) How to select effective demonstrations to enhance ICL performance?}

\begin{figure}[t]
    \centering
    \small
    \includegraphics[width=0.6\linewidth]{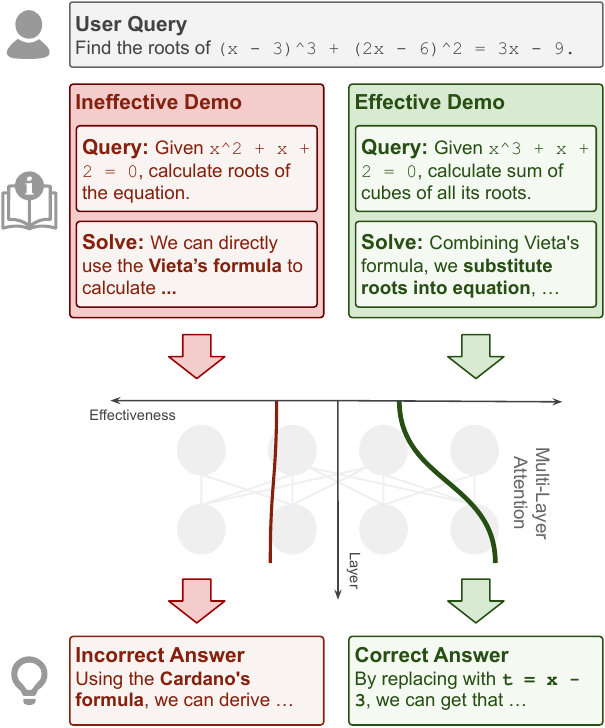}
    \caption{
        The gradient flow comparison of effective (\textgreen{green}) and ineffective demonstrations (\textred{red}).
        % 对于无效的demonstration，随着模型层数的增加，有效性一直在较低的水平
        For the ineffective demonstration, with the increase of model layers, the effectiveness remains at a low level.
        % 而对于有效的demonstration，有效性随着模型层数增大而被显著放大
        About the effectiveness, the effectiveness is significantly amplified as the layer increases.
    }
    \label{fig:motivation}
\end{figure}

% 遵循前人工作，我们主要采用梯度流方法来进行研究：示例对生成答案的梯度流贡献越大，则认为示例越有效
Following prior works \cite{wang-etal-2023-label,liu2025iterative}, we employ the gradient flow approach for our investigation: the greater the gradient flow from the demonstration to the generated answer, the more effective the demonstration is considered.
% 我们首先研究了单层线性自注意力模型中，决定梯度流大小的因素，从而给出了导致示例无效的原因为：示例中的信息已经被大模型学过，或者示例和用户查询无关
We first study the factors that determine the magnitude of the gradient flow in a single-layer linear self-attention model, thereby providing the reasons for an ineffective demonstration: the information in the demonstration has already been learned by LLMs, or the demonstration is irrelevant to the user query (RQ1).
% 然后我们提出，在多层模型中，层数的增大会放大示例间有效性的差异，从而LLM能够去更关注有效性更高的示例中的信息
We then prove that in a multi-layer linear self-attention model, \textbf{the difference in effectiveness among demonstrations is amplified along the increase of the number of layers}, resulting in the ignoration of information from less effective demonstrations (RQ2), as shown in Figure~\ref{fig:motivation}.
% 具体来说，如果示例A比示例B更有效，那随着层数的增大，二者对应的梯度流间的比值也会越来越大，即梯度流间的大小差异越来越大
Specifically, if a demonstration is more effective than the other, the ratio of their corresponding gradient flows increases with the number of layers, meaning the disparity in their magnitudes grows.

% 考虑到现有的示例选取方法主要关注示例和查询间的相似性，可能会选取无效的示例
Considering that existing demonstration selection methods mainly focus on the relevance between the demonstration and the query, which could select ineffective demonstrations \cite{wei2024larger,chen-etal-2023-many,wang2025vsynthesis}.
% 因此，基于上述讨论，我们提出基于梯度流的示例选取方法，对于每个查询，选取能使得梯度流最大的示例
Therefore, based on the above discussion, we propose \ourmethod, which selects the demonstration that maximizes the gradient flow with the given query (RQ3).
% 通过梯度流的大小来度量LLMs是否会使用到给定示例中的信息，从而确保提供的示例都在推理过程中有效，从而增强ICL的性能
Measure whether LLMs use the information in a given demonstration through the gradient flow, so as to ensure that the demonstrations provided are effective in the reasoning process, thus enhancing the performance of ICL.

% 为了验证我们结论和方法的有效性，我们在五个主流数据集和四个主流模型上进行实验
To validate our conclusions and the effectiveness of \ourmethod, we conduct experiments on five mainstream datasets and four mainstream models. 
% 分析实验表明，有效和无效示例间的梯度流比值会随着模型层数的增大而不断增大，证明了多层Transformer确实是示例有效性的放大器
Experimental analysis shows that during ICL, the gradient flow ratio between effective and ineffective demonstrations increases with the number of model layers, proving that the multi-layer Transformer indeed acts as an amplifier of demonstration effectiveness.
% 此外，实验结果表明，相较于最好的示例选取baseline，我们的方法平均带来了$xx%$的性能提升，证明了我们方法的有效性
Furthermore, the experimental results of \ourmethod indicate that compared to the best demonstration selection baseline, \ourmethod achieves a relative improvement of $6.8\%$ on average, proving its effectiveness.

% 综上所述，我们的贡献包括：
In summary, our contributions include:
\begin{itemize}
    % 我们分析了在ICL时，一个示例无效是因为它的信息已经被LLMs学过了，或者和给定的用户查询无关
    \item We argue that a demonstration in ICL is ineffective when its information is already learned by the LLMs or is irrelevant to the given user query.
    % 解释了大模型忽略无效示例是因为多层模型会放大示例有效性的作用
    \item We theoretically and empirically analyze the internal mechanism of LLMs that the multi-layer structure amplifies the ICL effectiveness between the effective demonstrations and the ineffective ones.
    % 提出了一种基于梯度流的示例选取方法，来确保选取高有效性示例
    \item We propose \ourmethod, a gradient-flow-based demonstration selection method that ensures the selection of highly effective demonstrations.
\end{itemize}

    \section{Analysis}
        \begin{table*}[ht]
    \small
    \centering
    \begin{tabular}{@{}p{0.28\textwidth} p{0.55\textwidth} l@{}}
    \toprule
    \textbf{RQ} & \textbf{Finding} & \textbf{Evidence} \\
    \midrule
    \textbf{RQ1}: What makes the demonstration ineffective? 
    & A demonstration is ineffective if the information it contains has already been learned by LLMs or is irrelevant to the user query. 
    & Equation~\ref{equ:single_layer_grad_flow} \\
    \midrule
    \textbf{RQ2}: How LLMs handle the demonstration effectiveness?
    & With deeper layers, the gradient flow disparity between effective and ineffective demonstrations widens, prioritizing to learn from the effective one.
    & Theorem~\ref{the:multi_layer_amplifier} \\
    \bottomrule
\end{tabular}

    \caption{
        The main research question (RQ), findings, and corresponding evidence of our analysis.
    }
    \label{tab:findings}
\end{table*}

% 在这一部分，我们从梯度流的角度，讨论模型在ICL中如何忽略无效的示例
In this section, we discuss why and how LLMs ignore ineffective demonstrations in ICL from a gradient flow perspective.
% 我们首先给出一些必要的数学符号和概念的定义
We first provide definitions for necessary mathematical notations and concepts.
% 然后，我们讨论单层线性自注意力网络上的梯度流，并讨论大模型如何判别不相关示例
Then, we discuss the gradient flow in a single-layer linear self-attention network and why LLMs distinguish ineffective demonstrations.
% 之后，我们讨论多层线性自注意力网络上的梯度流，并讨论大模型如何忽略不相关的示例
Subsequently, we discuss the gradient flow in a multi-layer linear self-attention network and how LLMs ignore ineffective demonstrations.
The main findings of our analysis are summarized in Table~\ref{tab:findings}.
% 这一节所有的证明见Appendix
All proofs in this section are provided in Appendix~\ref{app:proof}, and the calculation of the gradient flow follows \cite{wang-etal-2023-label}.

\subsection{Preliminary}
    % 在本文中，我们主要讨论1-shot
    In this paper, we primarily focus on the 1-shot setting.
    % 我们记$d = (d_x d_y)$表示示例，其中$d_x \in \mathbb{R}^{e}, d_y \in \mathbb{R}^{e}$分别表示示例的输入和输出序列，其中$e$表示嵌入维度
    Following \cite{zhang-etal-2024-trained}, we denote a demonstration as $d = \begin{pmatrix} d_x & d_y \end{pmatrix}$, where $d_x,d_y \in \mathbb{R}^{e}$ represent the input and output embedding vector of the demonstration, respectively. 
    Here, $e$ is the embedding dimension.
    % 记$q = (q_x 0)$表示用户查询，$q_x \in \mathbb{R}^{e}$表示查询的输入，最后的0表示输入中并没有提供查询对应的答案
    We denote a user query as $q = \begin{pmatrix} q_x & q_y \end{pmatrix}$, where $q_x \in \mathbb{R}^{e}$ is the query input embedding vector, and we set $q_y = 0$ indicates that the corresponding answer to the query is not provided in the input.
    % 记$E = (d q)$表示完整的网络输入
    We denote the complete network input as $E = \begin{pmatrix} d & q \end{pmatrix}$.
    % 在本文中，我们用$\Vert M \Vert$表示给定矩阵的1-范数
    In this paper, we use $\Vert M \Vert$ to denote the Frobenius norm~\cite{wu-etal-2022-maximum} of a given matrix $M$.

    % 遵循前人工作，我们定义线性自注意力网络为：
    Following the previous work~\cite{zhang-etal-2024-trained}, we define the linear self-attention network (LSA) as:
    \begin{equation}
        f_{LSA}(E;\theta) = E + W^{PV}E \cdot \frac{E^TW^{KQ}E}{\rho}
        \label{equ:lsa}
    \end{equation}
    % 其中$W^{PV} \in \mathbb{R}^{e \times e}$表示projection matrix和value matrix合并后的矩阵，$W^{KQ} \in \mathbb{R}^{e \times e}$表示key matrix和query matrix合并后的矩阵，具体的定义与\cite中一致
    $W^{PV} \in \mathbb{R}^{e \times e}$ is the combined projection and value matrix of Attention, and $W^{KQ} \in \mathbb{R}^{e \times e}$ is the combined key and query matrix, where the specific definitions are consistent with \cite{zhang-etal-2024-trained}.
    % $\rho$表示正则化系数，遵循\cite，本文中我们取$\rho = 1$
    $\rho$ is the normalization coefficient, where we set $\rho = 1$ in this paper following \cite{zhang-etal-2024-trained}.
    % 记$\theta = \{W^{PV}, W^{KQ}, \rho\}$表示所有的网络参数
    We denote all network parameters as $\theta = \{W^{PV}, W^{KQ}, \rho\}$.
    % 可以发现，Equation~\ref{}是将Transformer中的单层注意力网络中的激活函数替换为线性函数后的结果
    It can be observed that Equation~\ref{equ:lsa} is the result of replacing the activation function in a single-layer attention network of a Transformer with a linear function.
    % 记网络预测的答案为\hat{q_y}，即f_{LSA}(E;\theta)输出的最后一列对应的向量
    Specifically, we denote $\hat{q_y}(E;\theta)$ as the predicted answer of the given $E$ and $\theta$, abbreviated as $\hat{q_y}$, which is the vector corresponding to the last column of the output from $f_{LSA}(E;\theta)$.

    % 遵循前人工作，对于一个多元函数f和其相关的一个自变量x，我们定义f关于x的梯度流大小为f关于x的偏导数$\nabla_x f = \frac{\partial f}{\partial x}$
    Following previous work \cite{wang-etal-2023-label}, for a multivariate function $f$ and one of its independent variables $x$, we define the magnitude of the gradient flow of $f$ with respect to $x$ as the partial derivative of $f$ with respect to $x$, which is $\nabla_x f = \frac{\partial f}{\partial x}$.
    % 梯度流的大小度量了x的变化对f的影响
    The magnitude of the gradient flow measures the influence of the change in $x$ on $f$.
    % 具体到ICL任务上，输出$\hat{q_y}$关于输入示例$d$的梯度流反应了示例对答案的贡献，从而能反应示例中有多少信息参与到了答案生成
    Specifically, in the context of ICL, the gradient flow of the output $\hat{q_y}$ with respect to an input demonstration $d$ reflects the contribution of that demonstration to the answer. 
    Consequently, it can indicate how much information from the demonstration is utilized in generating the answer.

\subsection{Single-Layer Linear Self-Attention Network}
    % 我们首先分析单层网络上的梯度流
    We first analyze the gradient flow in a single-layer network.
    % 可以证明，输入的示例对答案生成的梯度流贡献为：
    It can be shown that the contribution of the input demonstration to the gradient flow for answer generation is:
    \begin{equation}
            \nabla_{d} \hat{q_y} = (W^{PV} d)_y (q^{\top} W^{KQ})^{\top} + (d^{\top}W^{KQ}q)W^{PV}_y
        \label{equ:single_layer_grad_flow}
    \end{equation}
    % Equation决定了示例中有多少的信息被用于答案的生成
    Equation~\ref{equ:single_layer_grad_flow} formally defines the gradient flow value of the single layer LSA, which presents how much information from the demonstration is used for answer generation.
    % 可以发现，在给定查询$q$的情况下，决定了Equation大小的因素可以分为两类:
    We can see that, given a query $q$, the factors determining the magnitude of the equation can be divided into two categories:
    % (i) 示例和用户查询间的相似性($-$)
    \textit{(i)} The similarity between the demonstration and the user query ($d^{\top}W^{KQ}q$).
    % (ii) 模型本身对是否已经学习了示例中的信息($-$)
    \textit{(ii)} Whether the model has already learned the information in the demonstration ($W^{PV} d$).
    % 因此，LLM判断是否使用了给定示例的信息，主要取决于示例和查询间的相似性，以及示例中的信息是否已经被模型学会了
    Therefore, the determination of whether to use the information from a given demonstration mainly depends on the similarity between the demonstration and the query, and whether the information in the demonstration has already been learned by the model.

    % 基于上述讨论，我们提出使用Equation~\ref中的参数，来将示例的有效性形式化地定义如下：
    Based on the discussion above, as the base of the following discussion, we propose to use the parameters in Equation~\ref{equ:single_layer_grad_flow}, formally define the effectiveness of demonstrations with the given query and model as follows:
    \begin{definition}[Demonstration Effectiveness]
        Given a query $q$ and model parameters $\theta$, if two demonstrations $d_1$ and $d_2$ satisfy that:
        $$\Vert W^{PV} d_1 \Vert \geq \Vert W^{PV} d_2 \Vert$$
        $$\Vert d_1^{\top}W^{KQ}q \Vert \geq \Vert d_2^{\top}W^{KQ}q \Vert$$
        then we say that $d_1$ is more effective than $d_2$ with respect to $q$ and $\theta$, denoted as:
        $$d_1 \succcurlyeq_{q;\theta} d_2$$
        \label{def:demonstration_effectivness}
    \end{definition}
    % 在Definition中，我们要求一个示例更有效当且仅当它包含更多模型没学过的知识，并且和查询更相似
    In Definition~\ref{def:demonstration_effectivness}, we require a demonstration to be more effective than the other if it contains more knowledge that the model has not learned, and it is more similar to the query.
    % 如果只有其中一项成立，就很难比较示例间的有效性，因为无法判断模型没学过的知识对性能影响更大，还是和查询间的相似性影响更大
    If only one of these conditions holds, it is difficult to compare the effectiveness of the demonstration because it is not possible to determine whether the knowledge or the similarity to the query has a greater impact on performance.
    % 我们在Figure~\ref中用实验验证了两个因素对ICL性能的影响
    We experimentally evaluate the impact of two factors on ICL performance in Figure~\ref{fig:decision_boundary}.

\subsection{Multi-Layer Linear Self-Attention Network}
    % 下面，我们讨论多层线性自注意力网络
    Next, we discuss the gradient flow of the multi-layer linear self-attention network.
    % 令L表示网络的总层数，记第l层的输入为E^{(l)}，网络参数为\theta^{(l)}
    Let $L$ be the total number of layers in the network. We denote the input to the $l$-th layer as $E^{(l)} = \begin{pmatrix} d^{(l)} & q^{(l)} \end{pmatrix}$ and its parameters as $\theta^{(l)}$, the corresponding predicted answer of $l$-th layer is $\hat{q}^{(l)}$.
    % 由于第l-1层的输出为第l层的输入：
    Since in the multi-layer LSA, the output of $(l-1)$-th layer is the input of the $l$-th layer:
    \begin{equation}
        E^{(l)} = f_{LSA}(E^{(l-1)}; \theta^{(l-1)})
    \end{equation}
    % 根据链式法则可知：
    According to the chain rule, we can derive that:
    \begin{equation}
        \frac{\partial \hat{q}^{(L)}_y}{\partial d^{(0)}} = \frac{\partial \hat{q}^{(L)}_y}{\partial E^{(L-1)}} \times \frac{\partial E^{(L-1)}}{\partial E^{(L-2)}} \times \cdots \times \frac{\partial E^{(1)}}{\partial d^{(0)}}
        \label{equ:multi_layer_chain_rule}
    \end{equation}
    % 即整个模型的梯度流为各个层的梯度流的乘积
    That is, the gradient flow of the whole model is the product of the gradient flow of each layer.

    % 由于网络结构较为复杂，因此关于多层模型我们只给出定性的分析
    Due to the complexity of the network structure, we only provide a qualitative analysis for the multi-layer model.
    % 基于Equation和Corollary，我们可以知道Equation中每一项的大小都和示例的有效性正相关，即：
    Based on Equation~\ref{equ:single_layer_grad_flow}, we know that the magnitude of each term in the chain rule is positively correlated with the demonstration effectiveness, i.e.:

    \begin{lemma}
        Let an $L$‑layer linear self‑attention (LSA) network be given.  
        For every layer $l=1,\dots,L$, assume there exist \emph{strictly increasing} functions
        \[
        g_l:\mathbb{R}_{\ge0}\!\to\!\mathbb{R}_{\ge0},
        \qquad
        h_l:\mathbb{R}_{\ge0}\!\to\!\mathbb{R}_{\ge0},
        \]
        such that for \emph{every} demonstration $d$ and query $q$, the following equalities hold:
        \begin{align}
            \Vert W^{PV, (l)} d^{(l)} \Vert &= g_l\!\bigl(\Vert W^{PV, (l-1)} d^{(l-1)} \Vert\bigr) \label{eq:C-matrix-PV} \\
            \Vert (d^{(l)})^{\top}W^{KQ, (l)}q^{(l)} \Vert &= h_l\!\bigl(\Vert (d^{(l-1)})^{\top}W^{KQ, (l-1)}q^{(l-1)} \Vert \bigr) \label{eq:C-matrix-KQ}
        \end{align}

        \noindent
        If two demonstrations satisfy
        \(d_1 \succcurlyeq_{q;\theta^{(0)}} d_2\) at the input layer, then for \emph{every}
        $l=1,\dots,L$, we shave
        \(
        d^{(l)}_1 \succcurlyeq_{q;\theta^{(l)}} d^{(l)}_2.
        \)

        \label{lem:multi_layer_flow}
    \end{lemma}

    % Lemma~\ref{lem:multi_layer_flow}中的条件假设了模型每一层对输入的作用效果是一致的
    The condition in Lemma~\ref{lem:multi_layer_flow} assumes that each layer of the model has a consistent effect on its input.
    % 基于Theorem可知，对于每一层输入的示例有效性越大，其相应的梯度流越大，对应的输出的有效性也越大
    Based on Lemma~\ref{lem:multi_layer_flow}, we know that for each layer, a more effective input demonstration results in a larger corresponding gradient flow on all layers.
    % 值得注意的是，对于相同的示例d，Theorem无法保证l层的梯度流均大于l-1层
    It is worth noting that for the same demonstration $d$, Lemma~\ref{lem:multi_layer_flow} does not guarantee that the gradient flow at layer $l$ is always greater than that at layer $l-1$.
    % 这是因为，梯度流的大小还取决于每一层的参数\theta^{(l)}，因此无法保证梯度流是递增的
    This is because the magnitude of the gradient flow also depends on the parameters $\theta^{(l)}$ of each layer, so it cannot be guaranteed that the gradient flow is monotonically increasing cross each layer.

    % 进一步我们可以推知，由于梯度流本身是偏导数，更大的梯度流会导致后续层的输出变化也会越大
    Based on Lemma~\ref{lem:multi_layer_flow}, we can deduce that since the gradient flow is a partial derivative, a larger gradient flow leads to a greater change in the output of subsequent layers.
    % 因此随着模型层数的增大，有效和无效示例间梯度流的差异也会越来越大：
    Therefore, as the number of model layers increases, the difference in gradient flow between effective and ineffective demonstrations also becomes larger.

    \begin{theorem}[Multi-Layer Attention is the Amplifier of Demonstration Effectiveness]
        For a given user query $q$ and a model $\theta$, let $d_1$ and $d_2$ be two demonstrations with corresponding inputs $E_1$ and $E_2$. 
        If the condition of Lemma~\ref{lem:multi_layer_flow} holds, for any $L \geq l_1 > l_2 \geq 1$, we can draw that the following inequalities hold:
        $$\frac{\Vert \nabla_{d^{(0)}} \hat{q_y}^{(l_1)}(E_1;\theta) \Vert}{\Vert \nabla_{d^{(0)}} \hat{q_y}^{(l_1)}(E_2;\theta) \Vert} \geq \frac{\Vert \nabla_{d^{(0)}} \hat{q_y}^{(l_2)}(E_1;\theta) \Vert}{\Vert \nabla_{d^{(0)}} \hat{q_y}^{(l_2)}(E_2;\theta) \Vert}$$
        \label{the:multi_layer_amplifier}
    \end{theorem}

    % Corollary表明，随着层数增大，示例间累计的梯度流之间的差距也会越来越大
    Theorem~\ref{the:multi_layer_amplifier} shows that as the number of layers increases, the gap between the cumulative gradient flows of different demonstrations also becomes increasingly large.
    % 这表明，多层Transformer模型起到了放大示例有效性的作用
    This indicates that the multi-layer LSA acts as an amplifier for demonstration effectiveness.

    \section{Methodology}
        % 在这一节，我们介绍\ourmethod，我们基于梯度流的选取方法，选取给定查询梯度流最大的示例作为选取结果
In this section, we introduce \ourmethod, our demonstration selection method based on gradient flow, selecting the demonstration with the largest flow to the given query as the selection result.
The prompt we used is shown in Appendix~\ref{app:prompt}.
% 考虑到Equation，在实际选取时，我们选用最后一层的梯度流作为选取指标，因为示例间的差异被充分放大，能更好地区分有效和无效的示例
Considering Theorem~\ref{the:multi_layer_amplifier}, in actual selection, we use the gradient flow of the last layer as the selection metric. 
This is because the differences between demonstrations are sufficiently amplified, allowing for a better distinction between effective and ineffective demonstrations.

% 然而，对于每个用户查询，在模型上进行完整推理得到梯度流会导致计算效率很低
However, performing a full inference pass on the model to obtain the gradient flow for each user query results in low computational efficiency. 
% 从Equation中可以发现，在计算梯度流，示例和用户查询的计算相对独立
From Equation~\ref{equ:single_layer_grad_flow}, it can be observed that in the calculation of the gradient flow, the computations for the demonstration and the user query are relatively independent. 
% 因此，我们可以首先分别计算示例和用户查询的编码向量，然后使用计算得到的结果再通过矩阵运算得到梯度流大小
Therefore, we can first compute the encoded vectors for the demonstrations and the user query separately, and then use these results to calculate the magnitude of the gradient flow through matrix operations. 
% 通过这种方式，可以显著降低计算开销，从而增强示例选取的效率
This approach significantly reduces the computational overhead, thereby enhancing the efficiency of demonstration selection.

% 具体来说，对于给定的示例池，我们会首先将每个示例分别作为输入，抽取最后一层的编码结果作为$\hat{d}$
Specifically, for a given demonstration pool, we first input each demonstration individually to extract the encoding result of the final layer as $\hat{d}$. 
% 然后，对于每个用户查询，我们同样将其作为输入，抽取最后一层的编码结果作为$\hat{q}$
Then, for each user query, we also input it to extract the encoding result of the final layer as $\hat{q}$. 
% 之后，将计算得到的$\hat{d}, \hat{q}$代入Equation~\ref{}，得到$\partial_d \hat{E}^{(1)}_{-1}$
Subsequently, the computed $\hat{d}$ and $\hat{q}$ are substituted into Equation~\ref{equ:single_layer_grad_flow} to obtain $\partial_d \hat{q}^{(L)}_y$. 
% 由于$\partial_d \hat{E}^{(1)}_{-1}$为一个$1 \times 2$的矩阵，包含了示例的输入和输出两部分的梯度流，考虑到输入和输出两部分同等重要，我们使用$\partial_d \hat{E}^{(1)}_{-1}$的1-范数作为示例选取的指标，来平衡两部分的重要性
Since $\partial_d \hat{q}^{(L)}_y$ is a $2 \times e$ matrix, containing the gradient flows for both the input and output parts of the demonstration, and considering that both parts are equally important, we use $\Vert \partial_d \hat{q}^{(L)}_y \Vert$ as the metric for demonstration selection to balance their importance.

\paragraph{Efficiency of \ourmethod}
    % \ourmethod的计算开销主要分为两部分，离线计算每个示例的编码结果，以及在线查询示例并进行推理
    The computational cost of \ourmethod is mainly divided into two parts: the offline computation of the encoding result for each demonstration, and the online retrieval of the demonstration and subsequent inference for a query. 
    % 记D={d_1, \dots, d_n}表示整个示例池，$\mathcal{M}_{\theta}$(x)表示给定模型在输入为$x$时的计算开销，查询q对应的示例检索结果为$d_q$
    Let $D=\{d_1, \dots, d_n\}$ represent the entire demonstration pool, and let $\mathcal{M}_{\theta}(x)$ denote the computational cost of the model $\mathcal{M}_{\theta}$ for a given input $x$. 
    For a query $q$, let the retrieved demonstration be $d_q$.

    % 那\ourmethod的离线处理的时间复杂度为： $O(\sum^{n}_i \mathcal{M}_{\theta}(d_i))$
    The time complexity of the offline processing for \ourmethod is: $O(\sum^{n}_{i=1} \mathcal{M}_{\theta}(d_i))$.
    % 尽管离线处理开销相对较高，但预处理示例编码是离线完成的，因此并不会对用户在线查询产生影响 
    Although the offline processing cost is relatively high, the pre-computation of demonstration encodings is done offline and thus does not affect the online user query process.

    % 而在线处理的时间复杂度为:
    The time complexity for online processing is:
    \begin{equation}
        O(\mathcal{M}_{\theta}(q) + 4 e^2 + \mathcal{M}_{\theta}(q + d_q))
        \label{equ:online_time_complexity}
    \end{equation}
    % Equation中，第一项为编码用户查询，第二项为计算Equation，第三项为根据检索示例生成用户查询的答案
    In Equation~\ref{equ:online_time_complexity}, the first term represents encoding the user query, the second term corresponds to calculating Equation~\ref{equ:single_layer_grad_flow}, and the third term is for generating the answer to the user query based on the retrieved demonstration. 
    % 考虑到\mathcal{M}_{\theta}与输入长度正相关，而计算Equation的复杂度要显著低于使用完整模型推理的复杂度\mathcal{M}_{\theta}，在线处理的时间复杂度即为：
    Considering that $\mathcal{M}_{\theta}$ is positively correlated with the input length and the complexity of calculating Equation~\ref{equ:single_layer_grad_flow} is significantly lower than that of a full model inference $\mathcal{M}_{\theta}$, the online processing time complexity simplifies to:
    \begin{equation}
        O(\mathcal{M}_{\theta}(q + d_q))
        \label{equ:offline_time_complexity}
    \end{equation}
    % 即与直接进行1-shot推理的时间复杂度一致，证明了\ourmethod的高效性
    This is equivalent to the time complexity of direct 1-shot inference, which demonstrates the high efficiency of \ourmethod.

    \section{Experiment}
        % 我们的实验主要分布三部分：
Our experiments are primarily divided into three parts:
% 1. 介绍实验的setting以及baselines
\textit{(i)} Introduction of the experimental settings and baselines.
% 2. 验证Theorem及其相关的结论
\textit{(ii)} Verification of Theorem~\ref{the:multi_layer_amplifier} and its related corollaries.
% 3. 验证\ourmethod的有效性，以及不同因素对\ourmethod的影响
\textit{(iii)} Validation of effectiveness and the impact of different factors on \ourmethod.

\subsection{Experiment Setup}
    \paragraph{Dataset}
        % 为了充分验证我们的结论和方法的有效性，我们在五个主流数据集上进行了实验，涵盖了不同的任务和领域，包括：
        To thoroughly validate our analytic conclusions and the effectiveness of our proposed method, we conduct experiments on five mainstream datasets that span various tasks and domains, including:
        % (i) 数学：GSM8K, MATH
        \textit{(i)} math: GSM8K~\cite{cobbe2021gsm8k} and MATH~\cite{hendrycks2021math};
        % (ii) 推理：ARC-Challenge, MMLU-Pro
        \textit{(ii)} reasoning: ARC-Challenge~\cite{yadav-etal-2019-arc} and MMLU-Pro~\cite{wang2024mmlupro};
        % (iii) 情感分析：Amazon Review
        \textit{(iii)} sentiment analysis: Amazon~Review~\cite{ni-etal-2019-amazon}.
        % 关于上述数据集的详细介绍见Appendix
        Detailed descriptions of these datasets are provided in Appendix~\ref{app:dataset}.
        % 在各个数据集上，我们均适用Exact Match（EM）作为评价指标
        We employ Exact Match (EM) as the evaluation metric across all datasets.

    \paragraph{Model}
        % 我们在四个模型上对本文的结论和方法进行验证，包括：Llama2-7b, Llama3.1-8b, Deepseek-R1-Distilled-Llama3.1-8b(Llama-R1-8b), Qwen3-8b\footnote{我们使用Qwen3的推理模式，来充分地评测}
        We validate our discovery and method on four LLMs: Llama2-7b~\cite{touvron2023llama2}, Llama3.1-8b~\cite{grattafiori2024llama3}, Deepseek-R1-Distilled-Llama3.1-8b (Llama-R1-8b)~\cite{deepseekai2025deepseekr1}, and Qwen3-8b~\cite{yang2025qwen3technicalreport}\footnote{We utilize the thinking mode of Qwen3 for a comprehensive evaluation.}.
        % 我们选用的模型涵盖了不同规模、系列和长思维链的大模型，从而充分验证我们的结论，并比较不同的模型来观察模型的改变对我们结论的影响
        Our selection of models encompasses various scales, series, and capabilities for generating long chains of thought (Long-CoT) \cite{chen2025reasoningerasurveylong}. 
        This diverse set allows for a robust verification of our conclusions and enables a comparative study on the influence of model characteristics on our findings.

    \paragraph{Baseline}
        % 我们将我们的方法和三类基线比较
        We compare our method against three classes of demonstration synthesis baselines following \cite{wang2025vsynthesis}: 
        % (i) 基于gram的方法（BM25）
        (i) n-gram-based methods (BM25) \cite{robertson-etal-2009-bm25,li-etal-2023-unified}, 
        % (ii) 基于向量相似度的方法（Cosine）
        (ii) vector similarity-based methods (Cosine) \cite{yang-etal-2023-representative}, and 
        % (iii) 基于大模型的方法（MMR、MoD）
        (iii) LLM-based methods (MMR~\cite{ye-etal-2023-complementary}, MoD~\cite{wang2024mixture}). 
        % 关于各个baseline的详细介绍和设置，见Appendix
        Detailed introductions and configurations for each baseline are provided in Appendix~\ref{app:baseline}.

    \paragraph{Implementation Detail}
        % 关于机理分析的实验，我们采用1-shot，来和分析中的setting对齐
        For the mechanism analysis experiments, we employ a 1-shot setting to align with the setup in our theoretical analysis. 
        % 关于\ourmethod的验证，遵循前人工作，我们采用3-shot，来确保最终的性能之间是可比较的
        For the validation of \ourmethod, we follow previous works \cite{wang2025vsynthesis} and adopt a 3-shot setting to ensure comparability of the final performance.
        % 遵循\cite，我们设置最大生成长度为$32768$，每个问题均只sample一个答案
        Following \cite{deepseekai2025deepseekr1}, we set the maximum generation length to $32768$, and for each question, we sample a single answer.
        % 我们的实验使用一张A100-80G，平均每个数据集上的检索和推理实验约为一小时
        Our experiments are performed on a single A100-80G GPU, with the selection and inference for each dataset taking approximately one hour on average.

\subsection{Experiment of Mechanism Analysis}
    % 若示例已经被模型学习或和用户查询无关，则梯度流为零
    \subsubsection{The Factors Making Demonstrations Ineffective}
        \begin{figure}
            \small
            \centering
            \includegraphics[width=0.95\linewidth]{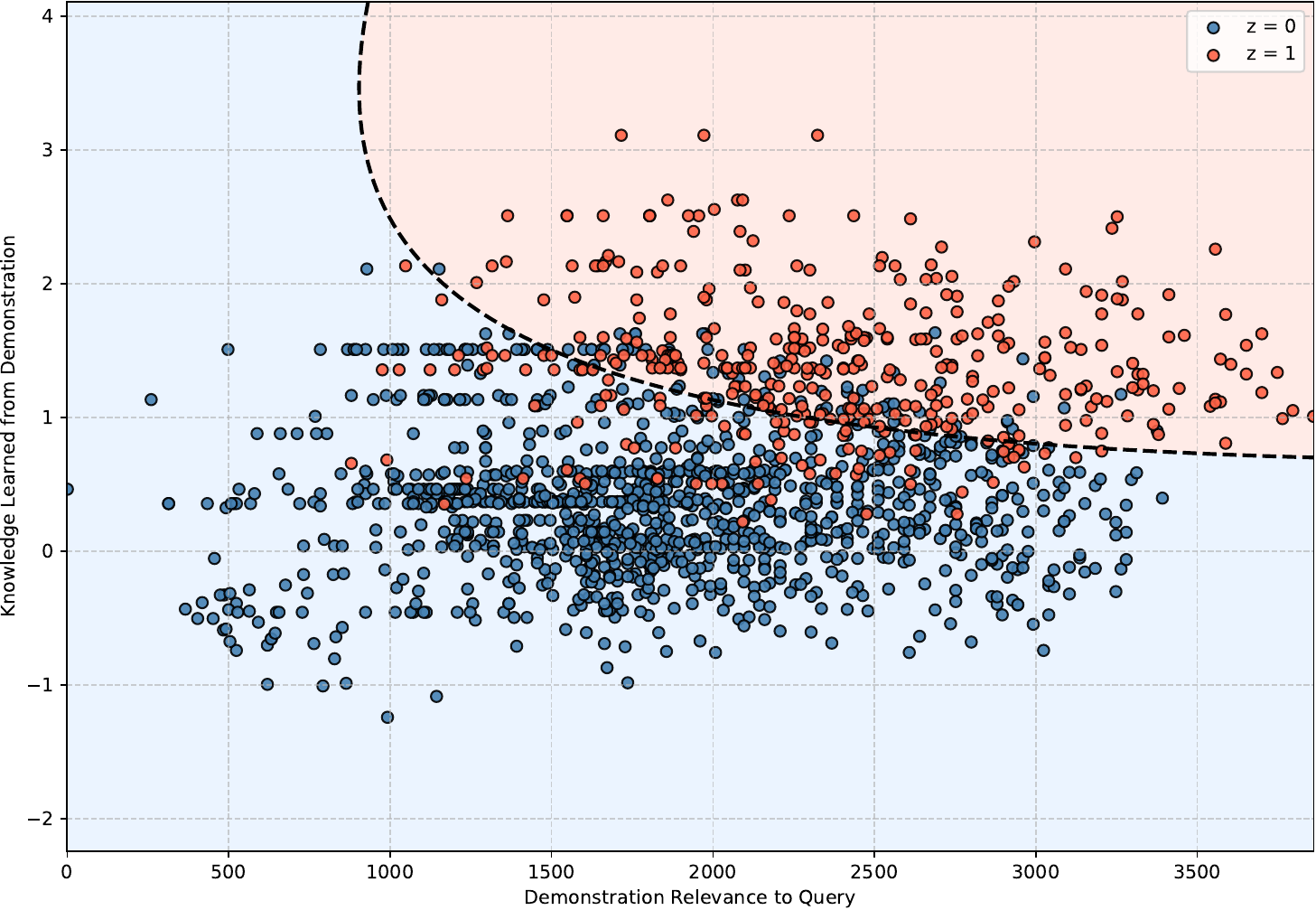}
            \caption{
                % 在所有数据集上，使用Llama3,1-8b，ICL性能随着示例相似性（X轴）和示例中未被模型学习过的知识（Y轴）的变化情况
                The ICL performance of Llama3.1-8b across all datasets with different demonstration relevance (X-axis) and unlearned knowledge within the demonstration (Y-axis). 
                % 红色点表示正确的点，蓝色点表示做错的点，图中的虚线表示决策边界
                \textred{Red} points denote correct prediction and \textblue{blue} points incorrect predictions. 
                The dashed line represents the decision boundary generated with polynomial logistic regression \cite{sklearn_api}.
            }
            \label{fig:decision_boundary}
        \end{figure}
    
        % 为了验证关于Equation的讨论中，梯度流为零的条件，我们统计了模型的性能随着$-$和$-$的变化情况
        To verify the condition for low gradient flow in the discussion regarding Equation~\ref{equ:single_layer_grad_flow}, we record the statistics of the model performance as a function of the changes in the demonstration relevance ($d^{\top}W^{KQ}q$) and learned knowledge ($W^{PV} d$).
        % 实验结果如图所示，从图中可以发现：
        The experimental results are shown in Figure~\ref{fig:decision_boundary}. 
        From the figure, we can observe that:
        % (i) 随着示例的relevance和从示例中学到的知识的增大，做对的数据点也逐渐增多，证明了Equation中的结论，即ICL的性能与这两个因素正相关
        \textit{(i)} As the relevance of the demonstrations and the knowledge learned from them increase, the number of correctly solved data points also increases, which verifies the conclusion from Equation~\ref{equ:single_layer_grad_flow} that the performance of ICL is positively correlated with these two factors.
        % (ii) 在示例的relevance和从示例中学到的知识超过一定阈值后，ICL才开始正确地求解给定的用户查询，表明只有提供足够多的有效信息才能帮助模型进行正确地推理
        \textit{(ii)} ICL only begins to correctly solve the given user queries after the relevance of the demonstrations and the knowledge learned from them surpass a certain threshold, which indicates that a sufficient amount of effective information that is relevant to the user query is necessary to enable the model to perform correct reasoning.

    \subsubsection{Gradient Flow is Amplified as the Layer Increasing}
        \begin{figure*}[ht]
            \small
            \centering
            \input{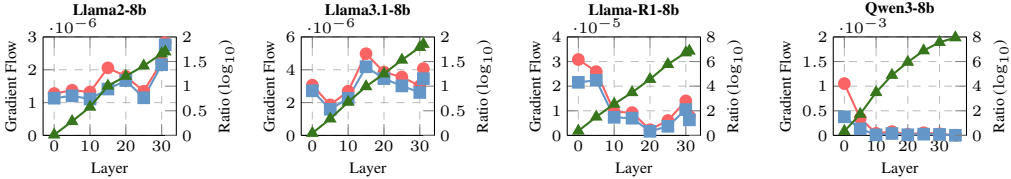}
            \caption{
                The average gradient flow (left y-axis) and the ratio $\frac{\Vert \nabla_{d^{(0)}} \hat{q_y}^{(l_1)}(E_1;\theta) \Vert}{\Vert \nabla_{d^{(0)}} \hat{q_y}^{(l_1)}(E_2;\theta) \Vert}$ of Theorem~\ref{the:multi_layer_amplifier} (right y-axis) cross different datasets under the $i$-th layer of each model, where $E_1$ denotes the input of the effective demonstrations and $E_2$ denotes the ineffective ones.
                \textred{Red} points denote the average gradient flow of the effective demonstrations, \textblue{blue} points denote the ineffective demonstrations, and \textgreen{green} points denote the ratio.
            }
            \label{fig:grad_flow_avg}
        \end{figure*}

        % 为了验证Theorem和Corollary的正确性，我们计算不同示例上的梯度流情况
        To validate the correctness of Lemma~\ref{lem:multi_layer_flow} and Theorem~\ref{the:multi_layer_amplifier}, we compute the gradient flow on different settings.
        % 对于给定的数据集，我们首先选取出zero-shot做错的数据
        For a given dataset, we first select the data points that are incorrectly predicted under the 0-shot setting.
        % 然后在这些数据上，我们分别选取出one-shot做对和做错的数据，作为demonstration有效和无效的数据，统计他们在每一层的梯度流的平均值
        From this subset, we then separate the data into two groups based on the 1-shot performance: those that become correctly predicted (\textit{effective demonstrations}) and those that remain incorrectly predicted (\textit{ineffective demonstrations}). 
        We then compute the average gradient flow for each layer across these two groups. 
        % 筛选出的有效和无效数据数量见Appendix
        The number of selected effective and ineffective data points is detailed in Appendix~\ref{app:scale_effective_ineffective}.

        % 实验结果如图所示，从图中可以发现：
        The experimental results are presented in Figure~\ref{fig:grad_flow_avg}. 
        From the figure, we can observe the following:
        % (i) 在各个setting下，有效示例在每一层的平均梯度流都强于无效示例，证明了Theorem的结论。而且，有效示例和无效示例间的梯度流比值也随着层数增大而增大，证明了Corollary
        \textit{(i)} Across all settings, the average gradient flow of effective demonstrations is stronger than that of ineffective demonstrations at every layer, which validates the conclusion of Lemma~\ref{lem:multi_layer_flow}. 
        Furthermore, the ratio of gradient flow between effective and ineffective demonstrations increases with the layer depth, which supports Theorem~\ref{the:multi_layer_amplifier}.
        % (ii) 长思维链模型的梯度流强度以及比值要显著强于其他模型，表明这类模型能更好地捕获示例中的信息，并且能更好地区分有效和无效示例间的差异，对示例中包含的有效信息更加敏感
        \textit{(ii)} The magnitude of the gradient flow, as well as the ratio, is significantly more pronounced in the Long-CoT model compared to other models. 
        This suggests that such models are more adept at capturing information from demonstrations and can better distinguish between effective and ineffective demonstrations, showing a higher sensitivity to the useful information contained within the demonstrations.
        % (iii) 不同模型上，都出现了在特定层梯度流显著增大的情况，表明模型学习示例中的信息主要集中在特定的层，而这些特定层在不同模型间存在一定的差异，但都在最后几层表现出强梯度流，表明模型会在输出时尤其关注示例中的信息
        \textit{(iii)} In all models, there are specific layers where the gradient flow exhibits a significant increase. 
        This indicates that the learning from demonstrations is primarily concentrated in these particular layers. 
        While the specific layers differ across models, a strong gradient flow is consistently observed in the final few layers, suggesting that the model pays special attention to the information in the demonstrations when generating the output.

\subsection{Experiment of \ourmethod}
    \subsubsection{\ourmethod is More Effective than Baselines}
        \begin{table}[ht]
            \small
            \centering
            \resizebox{0.48\textwidth}{!}{
\setlength{\tabcolsep}{1mm}
\begin{tabular}{cl|ccccc}
    \toprule
    \textbf{Model} & \textbf{Method} & \textbf{GSM8K} & \textbf{MATH} & \textbf{ARC-C} & \textbf{MMLU-Pro} & \textbf{Amazon} \\ 
    \midrule
    \multirow{6}{*}{\rotatebox{90}{Llama2-7b}} & Zero & $12.7$ & $5.0$ & $34.6$ & $14.5$ & $28.5$ \\
     & BM25 & $25.7$ & $8.4$ & $44.8$ & $17.1$ & $32.0$ \\
     & Cosine & $25.1$ & $7.2$ & $45.1$ & $17.5$ & $37.0$ \\
     & MMR & $24.9$ & $8.0$ & $45.7$ & $17.7$ & $33.5$ \\
     & MoD & $25.1$ & $7.2$ & $45.1$ & $17.5$ & $37.0$ \\
     & \ourmethod & $\textbf{26.7}$ & $\textbf{9.6}$ & $\textbf{46.5}$ & $\textbf{18.8}$ & $\textbf{37.5}$ \\
    \midrule
    \multirow{6}{*}{\rotatebox{90}{Llama3.1-8b}} & Zero & $83.7$ & $47.0$ & $82.0$ & $50.4$ & $63.5$ \\
     & BM25 & $84.5$ & $44.2$ & $84.6$ & $52.7$ & $69.5$ \\
     & Cosine & $85.0$ & $47.0$ & $85.2$ & $52.3$ & $69.0$ \\
     & MMR & $84.2$ & $45.0$ & $85.3$ & $52.0$ & $66.5$ \\
     & MoD & $85.0$ & $47.0$ & $85.2$ & $51.9$ & $69.0$ \\
     & \ourmethod & $\textbf{85.6}$ & $\textbf{48.2}$ & $\textbf{86.3}$ & $\textbf{56.0}$ & $\textbf{70.0}$ \\
    \midrule
    \multirow{6}{*}{\rotatebox{90}{Llama-R1-8b}} & Zero & $86.1$ & $75.4$ & $83.5$ & $58.2$ & $61.5$ \\
     & BM25 & $80.1$ & $74.2$ & $84.9$ & $52.7$ & $65.0$ \\
     & Cosine & $86.0$ & $74.6$ & $84.9$ & $55.9$ & $65.5$ \\
     & MMR & $85.6$ & $72.8$ & $84.4$ & $59.0$ & $66.0$ \\
     & MoD & $86.0$ & $74.6$ & $84.4$ & $56.9$ & $65.5$ \\
     & \ourmethod & $\textbf{87.2}$ & $\textbf{75.6}$ & $\textbf{85.7}$ & $\textbf{59.5}$ & $\textbf{67.0}$ \\
    \midrule
    \multirow{6}{*}{\rotatebox{90}{Qwen3-8b}} & Zero & $93.9$ & $76.2$ & $89.2$ & $64.9$ & $62.5$ \\
     & BM25 & $93.2$ & $76.8$ & $92.0$ & $64.0$ & $65.0$ \\
     & Cosine & $93.9$ & $77.4$ & $90.4$ & $65.0$ & $68.5$ \\
     & MMR & $93.1$ & $77.4$ & $90.8$ & $65.5$ & $74.5$ \\
     & MoD & $93.9$ & $78.2$ & $90.9$ & $64.8$ & $69.0$ \\
     & \ourmethod & $\textbf{94.2}$ & $\textbf{79.4}$ & $\textbf{92.6}$ & $\textbf{68.5}$ & $\textbf{75.0}$ \\
    \bottomrule
\end{tabular}
}
            \caption{
                The performance of \ourmethod compared with different baselines.
                ARC-C denotes ARC-Challenge, Amazon denotes Amazon Review.
                The best result under each setting is marked in \textbf{bold}.
            }
            \label{tab:performance_main}
        \end{table}
    
        % 为了验证\ourmethod的有效性，我们将我们的方法和其他baselines进行了比较，实验结果如Table所示
        To validate the effectiveness of \ourmethod, we conduct a comparative analysis against several baselines, with the experimental results presented in Table~\ref{tab:performance_main}. 
        % 相较于各个setting最好的baselines，我们的方法带来了平均1.0%的性能提升，证明了我们方法的有效性
        The result shows that \ourmethod achieves a relative improvement of $6.8\%$ on average over the best-performing baseline cross different setting, which substantiates its efficacy and generalizability.
        % 除此之外，从表中我们还可以发现：
        Furthermore, a deeper analysis of the results reveals several key observations:

        % (i) 从方法的角度，在很多setting下，MMR和MoD的性能并没有优于BM25这类简单的方法，甚至低于zero-shot。这表明，基于示例和用户查询相似度的方法并不能保证带来ICL的性能提升，因为模型可能已经学过了示例中的信息，而示例中的无关信息反而会误导模型的推理。而\ourmethod确保选取的示例中的信息能被模型充分使用，从而确保了性能
        \textit{(i) From a methodological perspective}: In many settings, the performance of MMR and MoD does not exceed that of simpler methods like BM25, and in some cases, is even inferior to the zero-shot approach. 
        This suggests that methods based on the similarity between demonstrations and the user query do not guarantee an enhancement in ICL performance due to that the model could have already been exposed to the information present in the demonstrations, and irrelevant information within these demonstrations could consequently mislead the model reasoning process. 
        In contrast, \ourmethod ensures that the information in the selected demonstrations is effectively utilized by the model, thereby securing the effectiveness of the selected demonstrations.

        % (ii) 从模型的角度，\ourmethod在Llama3.1-8b上带来的性能提升最显著，这是因为该模型没有使用长思维链，无法有效使用示例中的有效信息，因此相较于Llama-R1-8b和Qwen3-8b性能更依赖示例质量。而在Llama2-7b上\ourmethod带来的带来的性能提升相对较弱，这是因为Llama2中的知识较少，因此示例和用户查询间的相关性对ICL的性能影响更显著，因此\ourmethod与其他基于相关性的baseline之间性能差异并不大
        \textit{(ii) From a model perspective}: The most significant performance improvement from \ourmethod is observed on Llama3.1-8b. 
        This is likely because this model does not employ Long-CoT, rendering it less capable of effectively leveraging the useful information within the demonstrations. 
        Consequently, its performance is more dependent on the quality of the demonstrations compared to Llama-R1-8b and Qwen3-8b. 
        Conversely, the performance gain on Llama2-7b is relatively modest. 
        We attribute this to the limited knowledge base of Llama2 \cite{wang-etal-2025-language}, which makes the relevance between the demonstrations and the user query a more critical factor for ICL performance. 
        As a result, the performance gap between \ourmethod and other relevance-based baselines is less pronounced.

        % (iii) 从数据集的角度看，我们的方法在MATH和MMLU-Pro上带来的性能提升更显著，这是因为这两个数据集相较于其他数据集对特定知识的要求更高，因此更依赖示例中提供模型不会的知识，而\ourmethod相较于其他baseline可以更有效选取出包含模型所需知识的示例，从而取得了更显著的性能提升
        \textit{(iii) From a dataset perspective}: \ourmethod yielded more substantial performance gains on the MATH and MMLU-Pro datasets. 
        This is because these two datasets, compared to others, demand a higher degree of specialized knowledge. 
        Therefore, they are more reliant on the demonstrations to provide knowledge that the model lacks.
        \ourmethod, being more effective at selecting demonstrations that contain the requisite knowledge for the model, achieves a more significant performance improvement in these contexts.

    % 随着选取层数的增大，我们方法的有效性越来越强
    \subsubsection{The Performance of \ourmethod is Positively Correlated to Model Layer}
        \begin{table*}
            \small
            \centering
            \begin{tabular}{ll|cccccccc}
    \toprule
    \multirow{2}{*}{\textbf{Model}} & \multirow{2}{*}{\textbf{Dataset}} & \multicolumn{8}{c}{\textbf{Layer}} \\
     &  & \textbf{0} & \textbf{5} & \textbf{10} & \textbf{15} & \textbf{20} & \textbf{25} & \textbf{30} & \textbf{31} \\
    \midrule
     & GSM8K & $83.2$ & $84.4$ & $84.9$ & $85.1$ & $85.2$ & $85.6$ & $85.2$ & $\textbf{85.6}$ \\
    \cmidrule{2-10}
     & MATH & $44.4$ & $44.8$ & $45.8$ & $45.2$ & $46.2$ & $47.6$ & $46.8$ & $\textbf{48.2}$ \\
    \cmidrule{2-10}
    \multirow{-3}{*}{Llama3.1-8b} & ARC-Challenge & $82.1$ & $82.3$ & $82.1$ & $82.8$ & $83.4$ & $84.0$ & $85.1$ & $\textbf{86.3}$ \\
    \cmidrule{2-10}
     & MMLU-Pro & $50.8$ & $51.5$ & $53.6$ & $54.8$ & $54.8$ & $55.6$ & $\textbf{56.2}$ & $56.0$ \\
    \bottomrule
\end{tabular}
            \caption{
                The performance of \ourmethod using the gradient flow of different layer, where $31$ is the last layer of the model.
                The best result under each setting is marked in \textbf{bold}.
            }
            \label{tab:performance_cross_layer}
        \end{table*}
    
        % 为了证明Theorem，随着层数增大，有效和无效示例间的梯度流差异越来越大，从而模型能更好地选取示例，我们实验了\ourmethod使用不同层的梯度流进行示例选取的性能
        To validate Theorem~\ref{the:multi_layer_amplifier} that the disparity in gradient flow between effective and ineffective demonstrations increases with network layer, thereby enhancing the demonstration selection capability, we conduct the experiment on the performance of \ourmethod using gradient flows from different layers for demonstration selection. 
        % 实验结果如图所示，从图中我们可以发现：
        The experimental results are presented in Table~\ref{tab:performance_cross_layer}.
        From the table, we can observe the following:
        % (i) 在各个数据集上，随着层数的增大，模型的性能整体呈上升趋势，证明了Corollary的结论，即随着层数的增大，示例间有效性的差异也被放大，从而能更好地选出有效的示例
        \textit{(i)} Across all datasets, there is a general upward trend in model performance as the number of layers increases, which confirms the conclusion of Theorem~\ref{the:multi_layer_amplifier} that as network layer increases, the difference in effectiveness among demonstrations is amplified, which enables a better selection of effective demonstrations.
        % (ii) 然而，\ourmethod的性能并不随层数的增大而单调上升，这是因为我们实验的模型要比理论分析更加复杂，例如残差流的存在会降低多层transformer对梯度流的放大作用，导致模型误选不正确的示例，导致性能下降
        \textit{(ii)} However, the performance of \ourmethod does not monotonically increase with the number of layers since the models used in our experiments are more complex than those in our theoretical analysis. For instance, the presence of residual streams \cite{he-etal-2015-resnet} can diminish the amplifying effect of multi-layer transformers on the gradient flow, which could lead the model to erroneously select ineffective demonstrations, resulting in a decline in performance.

    \section{Related Works}
        \subsection{In-Context Learning}
    % ICL是一种有效增强LLM性能的方法，在输入中提供用户查询相关的示例来增强推理性能
    ICL is an effective method for enhancing the performance of LLMs by providing demonstrations related to the user query in the input to improve reasoning performance \cite{brown-2020-gpt3,dong-etal-2024-icl-survey}.
    % 现有的ICL工作可以分为三类：如何获取示例、如何选取示例、如何使用示例
    Existing ICL research can be categorized into three areas: how to acquire demonstrations, how to select demonstrations, and how to utilize demonstrations.
    % 为了降低人工标注示例的开销，人们提出了许多使用LLM来合成示例的方法，依赖已有的数据、相关数据、相似任务数据等资源来合成目标任务相关的示例
    To reduce the cost of manually labeling demonstrations, many methods have been proposed to synthesize demonstrations using LLMs \cite{long-etal-2024-llms}, relying on resources such as existing demonstrations \cite{su-etal-2024-demonstration,he-etal-2024-self}, related information \cite{chen-etal-2023-self,wang2025vsynthesis,chen2025maple}, and data from similar tasks \cite{wang2025incontext} to generate demonstrations relevant to the target task.
    % 示例选取主要关注如何从示例池中选取和目标相关的示例，早期的工作主要依赖基于gram的方法，而最新的工作则依赖模型增强对示例、查询中的信息处理
    Demonstration selection primarily focuses on how to select demonstrations from a pool that are relevant to the target query \cite{peng-etal-2024-revisiting,wang2025demonstration}. 
    Early work mainly relies on gram-based methods \cite{li-etal-2023-unified}, while recent work leverages model-enhanced processing of information within demonstrations and queries \cite{yang-etal-2023-representative,ye-etal-2023-complementary,wang2024mixture}.
    % 示例处理主要关注如何更好地使用检索得到的示例，例如调整示例的顺序，将示例编码为向量直接注入到模型中，在降低ICL推理开销的同时，进一步增强ICL性能
    Demonstration utilization focuses on how to make better use of the selected demonstrations, for example, by adjusting the order of the demonstrations \cite{lu-etal-2022-fantastically,pham2025rapid} or encoding them into vectors and directly injecting them into the model \cite{li2024incontext,wang2025elicit,li2025implicit}, thereby further enhancing ICL performance while reducing its inference overhead.

    % 然而，现有的示例选取方法主要关注示例和查询间的相关性，忽视了inference模型可能会忽视提供的示例现象，由于示例中的信息已经被模型学习过了，导致选取的示例并不能对ICL有帮助
    However, existing demonstration selection methods primarily focus on the relevance between the demonstration and the query, overlooking the phenomenon that the demonstrations could be ineffective due to that the information in the demonstrations has already been learned by the model.
    % 因此我们提出了\ourmethod，基于梯度流，确保选取的示例在推理时能对推理提供足够大的信息贡献，从而增强ICL有效性
    Therefore, we propose \ourmethod, which is based on gradient flow, to ensure that the selected demonstrations provide a sufficiently large information contribution during inference, thereby enhancing the performance of ICL.

\subsection{Mechanism of In-Context Learning}
    % 理解ICL的内部机理，能够帮助我们更好地增强ICL性能，同时理解模型的推理机制
    Understanding the internal mechanism of ICL can help us better enhance its performance and comprehend the model reasoning process \cite{zhou-etal-2024-mechanism-survey}.
    % 现有的ICL机理工作可以分为四类：理论推导、模型架构、训练数据和推理分析
    Existing work on the mechanism of ICL can be divided into four categories: theoretical derivation, model architecture, training data, and inference analysis.
    % 理论推导工作主要关注如何使用数学推导来证明ICL的有效性，例如ICL的收敛性以及收敛速度，并且其中有很多工作基于LSA开展分析
    Theoretical derivation work primarily focuses on using mathematical proofs to demonstrate the effectiveness of ICL, such as its convergence \cite{wies2023the,huang2024incontext,yang2024incontext} and convergence rate \cite{smart2025incontext,fu2024transformers,huang2025transformers,vladymyrov2024linear}, among which many works adapt the research based on LSA \cite{zhang-etal-2024-trained,mahankali2024one,lu2024incontext}.
    % 模型结构主要讨论了Transformer架构中不同模块在ICL中的作用，例如Attention在ICL中起到了主要作用，MLP层主要起到了辅助ICL的作用
    The study of model architecture mainly discusses the roles of different modules within Transformer architecture in ICL. 
    For example, the Attention mechanism plays a major role in ICL \cite{olsson2022incontextlearninginductionheads,chen2024how,oko-etal-2024-pretrained}, while the MLP layers primarily serve an auxiliary function \cite{li2024how,nguyen2025differential}.
    % 训练数据分析了ICL能力如何在训练过程中涌现出来的，现有的工作认为ICL能力主要来自训练任务的多样性，并且在训练过程中，模型逐渐从in-weight learning过渡到in-context learning
    Training data analysis examines how the ICL ability emerges during the training process. 
    Current work suggests that the ICL ability mainly originates from the diversity of training tasks \cite{raventos2023pretraining,wibisono2024unstructured,zhang2025understanding,goddard2025when,zhang2025training}, and during training, the model gradually transitions from in-weight learning to in-context learning \cite{singh2025strategy,park2025competition}.
    % 推理分析依靠观察ICL中推理的现象进行分析，包括不同因素对ICL的影响、内部计算信息的变化等
    Inference analysis relies on observing phenomena during ICL inference, including the influence of different factors on ICL and changes in internal computational information \cite{bigelow2024icl_random_binary,shi2024why_larger_models_different,lin2024dual_operating_modes,long2024does_icl_really_learn,sia2024where_icl_happen,zhao2024probing_decision_boundaries}.

    % 然而，现有的ICL机理研究在分析时均假设给定的示例和用户查询相关
    However, existing research on the ICL mechanism assumes that the given demonstrations are effective.
    % 但在实际推理时，会出现ICL时忽略提供的示例的现象，这一现象依然有待研究
    In actual inference, there is the phenomenon that demonstrations are ineffective, leading no performance improvement \cite{deepseekai2025deepseekr1,wang2025lcs}.
    % 因此在本文中，我们研究了忽略示例的现象，给出忽视示例的原因包括示例无关或示例已经被模型学习，并且提出了多层transformer是示例有效性的放大器，从而去忽视无关的示例
    Therefore, in this paper, we investigate the ineffective demonstrations, provide reasons for it including the demonstration being irrelevant or its information having already been learned by the model, and propose that multi-layer transformers act as amplifiers of demonstration effectiveness.

    \section{Conclusion}
        % 在本文中，我们主要讨论了ICL中的无效示例的现象
        In this paper, we primarily discuss the phenomenon of ineffective demonstrations in ICL.
        % 我们基于LSA，首先证明了一个示例无效是因为其中的信息已经被模型学习过，或者和用户查询无关
        Based on LSA, we first discuss that a demonstration is ineffective because the information it contains has already been learned by the model or is irrelevant to the user query. 
        % 然后我们证明了，在多层LSA中，随着模型层数增大，示例的有效性也会被方法，从而模型会去更关注相关示例中的信息
        We then demonstrate that in multi-layer LSA, as the number of model layers increases, the distinction in effectiveness among demonstrations is amplified, leading the model to focus more on the information within relevant demonstrations.
        % 为了在ICL时选用有效示例，基于上述讨论，我们提出了\ourmethod，通过将梯度流作为度量指标选取示例，确保选取结果的有效性，从而增强ICL性能
        To select effective demonstrations for ICL, based on the above discussion, we propose \ourmethod, which uses gradient flow as a metric to select demonstrations. This ensures the effectiveness of the selected results, thereby enhancing ICL performance.
        % 为了验证上述结论，我们在四个主流LLMs和五个主流数据集上进行了验证
        To validate these conclusions, we conduct experiments on four mainstream LLMs and five mainstream datasets, covering various tasks and domains.
        % 首先分析实验表明，随着模型层数的增加，示例间有效性的差异也被逐渐增大，证明了我们理论推导的结论
        First, under all experimental settings, analytical experiments show that as the number of model layers increases, the difference in effectiveness among demonstrations is progressively magnified, which corroborates our theoretical derivations. 
        % 其次，\ourmethod相较于现有的示例选取方法带来了平均$6.8\%$的相对性能提升，证明了我们方法的有效性
        Second, \ourmethod achieves an average relative performance improvement of $6.8\%$ compared to existing demonstration selection methods, proving the effectiveness of our method.

    \clearpage
    \bibliography{neurips2025}

\begin{thebibliography}{10}

\bibitem{grattafiori2024llama3}
et~al Aaron~Grattafiori, Abhimanyu~Dubey.
\newblock The llama 3 herd of models, 2024.

\bibitem{yang2025qwen3technicalreport}
et~al An~Yang, Anfeng~Li.
\newblock Qwen3 technical report, 2025.

\bibitem{bigelow2024icl_random_binary}
Eric~J. Bigelow, Ekdeep~S. Lubana, Robert~P. Dick, Hidenori Tanaka, and Tomer~D. Ullman.
\newblock In-context learning dynamics with random binary sequences.
\newblock In {\em International Conference on Learning Representations (ICLR)}, 2024.
\newblock OpenReview (ICLR 2024).

\bibitem{sklearn_api}
Lars Buitinck, Gilles Louppe, Mathieu Blondel, Fabian Pedregosa, Andreas Mueller, Olivier Grisel, Vlad Niculae, Peter Prettenhofer, Alexandre Gramfort, Jaques Grobler, Robert Layton, Jake VanderPlas, Arnaud Joly, Brian Holt, and Ga{\"{e}}l Varoquaux.
\newblock {API} design for machine learning software: experiences from the scikit-learn project.
\newblock In {\em ECML PKDD Workshop: Languages for Data Mining and Machine Learning}, pages 108--122, 2013.

\bibitem{chen-etal-2023-many}
Jiuhai Chen, Lichang Chen, Chen Zhu, and Tianyi Zhou.
\newblock How many demonstrations do you need for in-context learning?
\newblock In Houda Bouamor, Juan Pino, and Kalika Bali, editors, {\em Findings of the Association for Computational Linguistics: EMNLP 2023}, pages 11149--11159, Singapore, December 2023. Association for Computational Linguistics.

\bibitem{chen2025reasoningerasurveylong}
Qiguang Chen, Libo Qin, Jinhao Liu, Dengyun Peng, Jiannan Guan, Peng Wang, Mengkang Hu, Yuhang Zhou, Te~Gao, and Wanxiang Che.
\newblock Towards reasoning era: A survey of long chain-of-thought for reasoning large language models, 2025.

\bibitem{chen-etal-2023-self}
Wei-Lin Chen, Cheng-Kuang Wu, Yun-Nung Chen, and Hsin-Hsi Chen.
\newblock Self-{ICL}: Zero-shot in-context learning with self-generated demonstrations.
\newblock In Houda Bouamor, Juan Pino, and Kalika Bali, editors, {\em Proceedings of the 2023 Conference on Empirical Methods in Natural Language Processing}, pages 15651--15662, Singapore, December 2023. Association for Computational Linguistics.

\bibitem{chen2024how}
Xingwu Chen, Lei Zhao, and Difan Zou.
\newblock How transformers utilize multi-head attention in in-context learning? a case study on sparse linear regression.
\newblock In {\em The Thirty-eighth Annual Conference on Neural Information Processing Systems}, 2024.

\bibitem{chen2025maple}
Zihan Chen, Song Wang, Zhen Tan, Jundong Li, and Cong Shen.
\newblock {MAPLE}: Many-shot adaptive pseudo-labeling for in-context learning.
\newblock In {\em Forty-second International Conference on Machine Learning}, 2025.

\bibitem{cobbe2021gsm8k}
Karl Cobbe, Vineet Kosaraju, Mohammad Bavarian, Mark Chen, Heewoo Jun, Lukasz Kaiser, Matthias Plappert, Jerry Tworek, Jacob Hilton, Reiichiro Nakano, Christopher Hesse, and John Schulman.
\newblock Training verifiers to solve math word problems, 2021.

\bibitem{deepseekai2025deepseekr1}
DeepSeek-AI.
\newblock Deepseek-r1: Incentivizing reasoning capability in llms via reinforcement learning, 2025.

\bibitem{dong-etal-2024-icl-survey}
Qingxiu Dong, Lei Li, Damai Dai, Ce~Zheng, Jingyuan Ma, Rui Li, Heming Xia, Jingjing Xu, Zhiyong Wu, Baobao Chang, Xu~Sun, Lei Li, and Zhifang Sui.
\newblock A survey on in-context learning.
\newblock In Yaser Al-Onaizan, Mohit Bansal, and Yun-Nung Chen, editors, {\em Proceedings of the 2024 Conference on Empirical Methods in Natural Language Processing}, pages 1107--1128, Miami, Florida, USA, November 2024. Association for Computational Linguistics.

\bibitem{fu2024transformers}
Deqing Fu, Tian qi~Chen, Robin Jia, and Vatsal Sharan.
\newblock Transformers learn to achieve second-order convergence rates for in-context linear regression.
\newblock In {\em The Thirty-eighth Annual Conference on Neural Information Processing Systems}, 2024.

\bibitem{goddard2025when}
Chase Goddard, Lindsay~M. Smith, Vudtiwat Ngampruetikorn, and David~J. Schwab.
\newblock When can in‑context learning generalize out of task distribution?
\newblock In {\em International Conference on Machine Learning (ICML)}, 2025.
\newblock OpenReview (ICML 2025).

\bibitem{he-etal-2015-resnet}
Kaiming He, Xiangyu Zhang, Shaoqing Ren, and Jian Sun.
\newblock Deep residual learning for image recognition.
\newblock In {\em 2016 IEEE Conference on Computer Vision and Pattern Recognition (CVPR)}, pages 770--778, 2016.

\bibitem{he-etal-2024-self}
Wei He, Shichun Liu, Jun Zhao, Yiwen Ding, Yi~Lu, Zhiheng Xi, Tao Gui, Qi~Zhang, and Xuanjing Huang.
\newblock Self-demos: Eliciting out-of-demonstration generalizability in large language models.
\newblock In Kevin Duh, Helena Gomez, and Steven Bethard, editors, {\em Findings of the Association for Computational Linguistics: NAACL 2024}, pages 3829--3845, Mexico City, Mexico, June 2024. Association for Computational Linguistics.

\bibitem{hendrycks2021math}
Dan Hendrycks, Collin Burns, Saurav Kadavath, Akul Arora, Steven Basart, Eric Tang, Dawn Song, and Jacob Steinhardt.
\newblock Measuring mathematical problem solving with the {MATH} dataset.
\newblock In {\em Thirty-fifth Conference on Neural Information Processing Systems Datasets and Benchmarks Track (Round 2)}, 2021.

\bibitem{huang2025transformers}
Jianhao Huang, Zixuan Wang, and Jason~D. Lee.
\newblock Transformers learn to implement multi-step gradient descent with chain of thought.
\newblock In {\em The Thirteenth International Conference on Learning Representations}, 2025.

\bibitem{huang2024incontext}
Yu~Huang, Yuan Cheng, and Yingbin Liang.
\newblock In-context convergence of transformers, 2024.

\bibitem{touvron2023llama2}
et~al Hugo~Touvron, Louis~Martin.
\newblock Llama 2: Open foundation and fine-tuned chat models, 2023.

\bibitem{li2024incontext}
Dongfang Li, zhenyu liu, Xinshuo Hu, Zetian Sun, Baotian Hu, and Min Zhang.
\newblock In-context learning state vector with inner and momentum optimization.
\newblock In {\em The Thirty-eighth Annual Conference on Neural Information Processing Systems}, 2024.

\bibitem{li2024how}
Hongkang Li, Meng Wang, Songtao Lu, Xiaodong Cui, and Pin-Yu Chen.
\newblock How do nonlinear transformers learn and generalize in in-context learning?
\newblock In {\em Forty-first International Conference on Machine Learning}, 2024.

\bibitem{li-etal-2023-unified}
Xiaonan Li, Kai Lv, Hang Yan, Tianyang Lin, Wei Zhu, Yuan Ni, Guotong Xie, Xiaoling Wang, and Xipeng Qiu.
\newblock Unified demonstration retriever for in-context learning.
\newblock In Anna Rogers, Jordan Boyd-Graber, and Naoaki Okazaki, editors, {\em Proceedings of the 61st Annual Meeting of the Association for Computational Linguistics (Volume 1: Long Papers)}, pages 4644--4668, Toronto, Canada, July 2023. Association for Computational Linguistics.

\bibitem{li2025implicit}
Zhuowei Li, Zihao Xu, Ligong Han, Yunhe Gao, Song Wen, Di~Liu, Hao Wang, and Dimitris~N. Metaxas.
\newblock Implicit in-context learning.
\newblock In {\em The Thirteenth International Conference on Learning Representations}, 2025.

\bibitem{lightman2024lets}
Hunter Lightman, Vineet Kosaraju, Yuri Burda, Harrison Edwards, Bowen Baker, Teddy Lee, Jan Leike, John Schulman, Ilya Sutskever, and Karl Cobbe.
\newblock Let's verify step by step.
\newblock In {\em The Twelfth International Conference on Learning Representations}, 2024.

\bibitem{lin2024dual_operating_modes}
Ziqian Lin and Kangwook Lee.
\newblock Dual operating modes of in‑context learning.
\newblock In {\em Proceedings of the 41st International Conference on Machine Learning (ICML)}, pages 30135--30188. PMLR, 2024.
\newblock PMLR (ICML 2024).

\bibitem{liu2025iterative}
Yiting Liu and Zhi-Hong Deng.
\newblock Iterative vectors: In-context gradient steering without backpropagation.
\newblock In {\em Forty-second International Conference on Machine Learning}, 2025.

\bibitem{long-etal-2024-llms}
Lin Long, Rui Wang, Ruixuan Xiao, Junbo Zhao, Xiao Ding, Gang Chen, and Haobo Wang.
\newblock On {LLM}s-driven synthetic data generation, curation, and evaluation: A survey.
\newblock In Lun-Wei Ku, Andre Martins, and Vivek Srikumar, editors, {\em Findings of the Association for Computational Linguistics: ACL 2024}, pages 11065--11082, Bangkok, Thailand, August 2024. Association for Computational Linguistics.

\bibitem{long2024does_icl_really_learn}
Quanyu Long, Yin Wu, Wenya Wang, and Sinno~Jialin Pan.
\newblock Does in-context learning really learn? rethinking how large language models respond and solve tasks via in-context learning.
\newblock In {\em Conference on Learning and Modeling (COLM)}, 2024.
\newblock OpenReview (COLM 2024).

\bibitem{lu-etal-2022-fantastically}
Yao Lu, Max Bartolo, Alastair Moore, Sebastian Riedel, and Pontus Stenetorp.
\newblock Fantastically ordered prompts and where to find them: Overcoming few-shot prompt order sensitivity.
\newblock In Smaranda Muresan, Preslav Nakov, and Aline Villavicencio, editors, {\em Proceedings of the 60th Annual Meeting of the Association for Computational Linguistics (Volume 1: Long Papers)}, pages 8086--8098, Dublin, Ireland, May 2022. Association for Computational Linguistics.

\bibitem{lu2024incontext}
Yue Lu, Mary Letey, Jacob~A Zavatone-Veth, Anindita Maiti, and Cengiz Pehlevan.
\newblock In-context learning by linear attention: Exact asymptotics and experiments.
\newblock In {\em NeurIPS 2024 Workshop on Mathematics of Modern Machine Learning}, 2024.

\bibitem{mahankali2024one}
Arvind~V. Mahankali, Tatsunori Hashimoto, and Tengyu Ma.
\newblock One step of gradient descent is provably the optimal in-context learner with one layer of linear self-attention.
\newblock In {\em The Twelfth International Conference on Learning Representations}, 2024.

\bibitem{nguyen2025differential}
Alex Nguyen and Gautam Reddy.
\newblock Differential learning kinetics govern the transition from memorization to generalization during in-context learning.
\newblock In {\em The Thirteenth International Conference on Learning Representations}, 2025.

\bibitem{ni-etal-2019-amazon}
Jianmo Ni, Jiacheng Li, and Julian McAuley.
\newblock Justifying recommendations using distantly-labeled reviews and fine-grained aspects.
\newblock In Kentaro Inui, Jing Jiang, Vincent Ng, and Xiaojun Wan, editors, {\em Proceedings of the 2019 Conference on Empirical Methods in Natural Language Processing and the 9th International Joint Conference on Natural Language Processing (EMNLP-IJCNLP)}, pages 188--197, Hong Kong, China, November 2019. Association for Computational Linguistics.

\bibitem{oko-etal-2024-pretrained}
Kazusato Oko, Yujin Song, Taiji Suzuki, and Denny Wu.
\newblock Pretrained transformer efficiently learns low-dimensional target functions in-context.
\newblock In {\em NeurIPS}, 2024.

\bibitem{olsson2022incontextlearninginductionheads}
Catherine Olsson, Nelson Elhage, Neel Nanda, Nicholas Joseph, Nova DasSarma, Tom Henighan, Ben Mann, Amanda Askell, Yuntao Bai, Anna Chen, Tom Conerly, Dawn Drain, Deep Ganguli, Zac Hatfield-Dodds, Danny Hernandez, Scott Johnston, Andy Jones, Jackson Kernion, Liane Lovitt, Kamal Ndousse, Dario Amodei, Tom Brown, Jack Clark, Jared Kaplan, Sam McCandlish, and Chris Olah.
\newblock In-context learning and induction heads, 2022.

\bibitem{park2025competition}
Core~Francisco Park, Ekdeep~Singh Lubana, and Hidenori Tanaka.
\newblock Competition dynamics shape algorithmic phases of in-context learning.
\newblock In {\em The Thirteenth International Conference on Learning Representations}, 2025.

\bibitem{peng-etal-2024-revisiting}
Keqin Peng, Liang Ding, Yancheng Yuan, Xuebo Liu, Min Zhang, Yuanxin Ouyang, and Dacheng Tao.
\newblock Revisiting demonstration selection strategies in in-context learning.
\newblock In Lun-Wei Ku, Andre Martins, and Vivek Srikumar, editors, {\em Proceedings of the 62nd Annual Meeting of the Association for Computational Linguistics (Volume 1: Long Papers)}, pages 9090--9101, Bangkok, Thailand, August 2024. Association for Computational Linguistics.

\bibitem{pham2025rapid}
Kha Pham, Hung Le, Man Ngo, and Truyen Tran.
\newblock Rapid selection and ordering of in-context demonstrations via prompt embedding clustering.
\newblock In {\em The Thirteenth International Conference on Learning Representations}, 2025.

\bibitem{raventos2023pretraining}
Allan Ravent{\'o}s, Mansheej Paul, Feng Chen, and Surya Ganguli.
\newblock Pretraining task diversity and the emergence of non‑bayesian in‑context learning for regression.
\newblock In {\em Advances in Neural Information Processing Systems (NeurIPS)}, 2023.
\newblock OpenReview (NeurIPS 2023).

\bibitem{robertson-etal-2009-bm25}
Stephen Robertson and Hugo Zaragoza.
\newblock The probabilistic relevance framework: Bm25 and beyond.
\newblock {\em Found. Trends Inf. Retr.}, 3(4):333–389, April 2009.

\bibitem{shi2024why_larger_models_different}
Zhenmei Shi, Zhouyan Xu, Junyi Wei, and Yingyu Liang.
\newblock Why larger language models do in‑context learning differently?
\newblock In {\em International Conference on Machine Learning (ICML)}, 2024.
\newblock OpenReview (ICML 2024).

\bibitem{sia2024where_icl_happen}
Suzanna Sia, David Mueller, and Kevin Duh.
\newblock Where does in-context learning happen in large language models?
\newblock In {\em NeurIPS (Poster)}, 2024.
\newblock OpenReview (NeurIPS 2024).

\bibitem{singh2025strategy}
Aaditya~K Singh, Ted Moskovitz, Sara Dragutinovi{\'c}, Felix Hill, Stephanie~C.Y. Chan, and Andrew~M Saxe.
\newblock Strategy coopetition explains the emergence and transience of in-context learning.
\newblock In {\em Forty-second International Conference on Machine Learning}, 2025.

\bibitem{smart2025incontext}
Matthew Smart, Alberto Bietti, and Anirvan~M. Sengupta.
\newblock In-context denoising with one-layer transformers: Connections between attention and associative memory retrieval.
\newblock In {\em Forty-second International Conference on Machine Learning}, 2025.

\bibitem{su-etal-2024-demonstration}
Yi~Su, Yunpeng Tai, Yixin Ji, Juntao Li, Yan Bowen, and Min Zhang.
\newblock Demonstration augmentation for zero-shot in-context learning.
\newblock In Lun-Wei Ku, Andre Martins, and Vivek Srikumar, editors, {\em Findings of the Association for Computational Linguistics: ACL 2024}, pages 14232--14244, Bangkok, Thailand, August 2024. Association for Computational Linguistics.

\bibitem{brown-2020-gpt3}
et~al Tom B.~Brown, Benjamin~Mann.
\newblock Language models are few-shot learners, 2020.

\bibitem{vaswani-etal-2017-transformer}
Ashish Vaswani, Noam Shazeer, Niki Parmar, Jakob Uszkoreit, Llion Jones, Aidan~N Gomez, \L~ukasz Kaiser, and Illia Polosukhin.
\newblock Attention is all you need.
\newblock In I.~Guyon, U.~Von Luxburg, S.~Bengio, H.~Wallach, R.~Fergus, S.~Vishwanathan, and R.~Garnett, editors, {\em Advances in Neural Information Processing Systems}, volume~30. Curran Associates, Inc., 2017.

\bibitem{vladymyrov2024linear}
Max Vladymyrov, Johannes von Oswald, Mark Sandler, and Rong Ge.
\newblock Linear transformers are versatile in-context learners.
\newblock In {\em ICML 2024 Workshop on In-Context Learning}, 2024.

\bibitem{wang2025incontext}
Dingzirui Wang, Xuanliang Zhang, Qiguang Chen, Longxu Dou, Xiao Xu, Rongyu Cao, YINGWEI MA, Qingfu Zhu, Wanxiang Che, Binhua Li, Fei Huang, and Yongbin Li.
\newblock In-context transfer learning: Demonstration synthesis by transferring similar tasks, 2025.

\bibitem{wang2025vsynthesis}
Dingzirui Wang, Xuanliang Zhang, Keyan Xu, Qingfu Zhu, Wanxiang Che, and Yang Deng.
\newblock V-synthesis: Task-agnostic synthesis of consistent and diverse in-context demonstrations from scratch via v-entropy, 2025.

\bibitem{wang2025lcs}
Dingzriui Wang, Xuanliang Zhang, Keyan Xu, Qingfu Zhu, Wanxiang Che, and Yang Deng.
\newblock Learning-to-context slope: Evaluating in-context learning effectiveness beyond performance illusions, 2025.

\bibitem{wang2025elicit}
Futing Wang, Jianhao Yan, Yue Zhang, and Tao Lin.
\newblock {ELICIT}: {LLM} augmentation via external in-context capability.
\newblock In {\em The Thirteenth International Conference on Learning Representations}, 2025.

\bibitem{wang-etal-2023-label}
Lean Wang, Lei Li, Damai Dai, Deli Chen, Hao Zhou, Fandong Meng, Jie Zhou, and Xu~Sun.
\newblock Label words are anchors: An information flow perspective for understanding in-context learning.
\newblock In Houda Bouamor, Juan Pino, and Kalika Bali, editors, {\em Proceedings of the 2023 Conference on Empirical Methods in Natural Language Processing}, pages 9840--9855, Singapore, December 2023. Association for Computational Linguistics.

\bibitem{wang-etal-2025-language}
Shumin Wang, Yuexiang Xie, Bolin Ding, Jinyang Gao, and Yanyong Zhang.
\newblock Language adaptation of large language models: An empirical study on {LL}a{MA}2.
\newblock In Owen Rambow, Leo Wanner, Marianna Apidianaki, Hend Al-Khalifa, Barbara~Di Eugenio, and Steven Schockaert, editors, {\em Proceedings of the 31st International Conference on Computational Linguistics}, pages 7195--7208, Abu Dhabi, UAE, January 2025. Association for Computational Linguistics.

\bibitem{wang2024mixture}
Song Wang, Zihan Chen, Chengshuai Shi, Cong Shen, and Jundong Li.
\newblock Mixture of demonstrations for in-context learning.
\newblock In {\em The Thirty-eighth Annual Conference on Neural Information Processing Systems}, 2024.

\bibitem{wang2025demonstration}
Xubin Wang, Jianfei Wu, Yuan Yichen, Deyu Cai, Mingzhe Li, and Weijia Jia.
\newblock Demonstration selection for in-context learning via reinforcement learning.
\newblock In {\em Forty-second International Conference on Machine Learning}, 2025.

\bibitem{wang2024mmlupro}
Yubo Wang, Xueguang Ma, Ge~Zhang, Yuansheng Ni, Abhranil Chandra, Shiguang Guo, Weiming Ren, Aaran Arulraj, Xuan He, Ziyan Jiang, Tianle Li, Max Ku, Kai Wang, Alex Zhuang, Rongqi Fan, Xiang Yue, and Wenhu Chen.
\newblock {MMLU}-pro: A more robust and challenging multi-task language understanding benchmark.
\newblock In {\em The Thirty-eight Conference on Neural Information Processing Systems Datasets and Benchmarks Track}, 2024.

\bibitem{wei2024larger}
Jerry Wei, Jason Wei, Yi~Tay, Dustin Tran, Albert Webson, Yifeng Lu, Xinyun Chen, Hanxiao Liu, Da~Huang, Denny Zhou, and Tengyu Ma.
\newblock Larger language models do in-context learning differently, 2024.

\bibitem{wibisono2024unstructured}
Kevin~Christian Wibisono and Yixin Wang.
\newblock From unstructured data to in‑context learning: Exploring what tasks can be learned and when.
\newblock In {\em Advances in Neural Information Processing Systems (NeurIPS)}, 2024.
\newblock OpenReview (NeurIPS 2024).

\bibitem{wies2023the}
Noam Wies, Yoav Levine, and Amnon Shashua.
\newblock The learnability of in-context learning.
\newblock In {\em Thirty-seventh Conference on Neural Information Processing Systems}, 2023.

\bibitem{wu-etal-2022-maximum}
Yuan Wu, Diana Inkpen, and Ahmed El{-}Roby.
\newblock Maximum batch frobenius norm for multi-domain text classification.
\newblock In {\em {IEEE} International Conference on Acoustics, Speech and Signal Processing, {ICASSP} 2022, Virtual and Singapore, 23-27 May 2022}, pages 3763--3767. {IEEE}, 2022.

\bibitem{yadav-etal-2019-arc}
Vikas Yadav, Steven Bethard, and Mihai Surdeanu.
\newblock Quick and (not so) dirty: Unsupervised selection of justification sentences for multi-hop question answering.
\newblock In Kentaro Inui, Jing Jiang, Vincent Ng, and Xiaojun Wan, editors, {\em Proceedings of the 2019 Conference on Empirical Methods in Natural Language Processing and the 9th International Joint Conference on Natural Language Processing (EMNLP-IJCNLP)}, pages 2578--2589, Hong Kong, China, November 2019. Association for Computational Linguistics.

\bibitem{yang2024incontext}
Tong Yang, Yu~Huang, Yingbin Liang, and Yuejie Chi.
\newblock In-context learning with representations: Contextual generalization of trained transformers.
\newblock In {\em ICML 2024 Workshop on Theoretical Foundations of Foundation Models}, 2024.

\bibitem{yang-etal-2023-representative}
Zhao Yang, Yuanzhe Zhang, Dianbo Sui, Cao Liu, Jun Zhao, and Kang Liu.
\newblock Representative demonstration selection for in-context learning with two-stage determinantal point process.
\newblock In Houda Bouamor, Juan Pino, and Kalika Bali, editors, {\em Proceedings of the 2023 Conference on Empirical Methods in Natural Language Processing}, pages 5443--5456, Singapore, December 2023. Association for Computational Linguistics.

\bibitem{ye-etal-2023-complementary}
Xi~Ye, Srinivasan Iyer, Asli Celikyilmaz, Veselin Stoyanov, Greg Durrett, and Ramakanth Pasunuru.
\newblock Complementary explanations for effective in-context learning.
\newblock In Anna Rogers, Jordan Boyd-Graber, and Naoaki Okazaki, editors, {\em Findings of the Association for Computational Linguistics: ACL 2023}, pages 4469--4484, Toronto, Canada, July 2023. Association for Computational Linguistics.

\bibitem{zhang-etal-2024-trained}
Ruiqi Zhang, Spencer Frei, and Peter~L. Bartlett.
\newblock Trained transformers learn linear models in-context.
\newblock {\em Journal of Machine Learning Research}, 25(49):1--55, 2024.

\bibitem{zhang2025understanding}
Xingxuan Zhang, Haoran Wang, Jiansheng Li, Yuan Xue, Shikai Guan, Renzhe Xu, Hao Zou, Han Yu, and Peng Cui.
\newblock Understanding the generalization of in‑context learning in transformers: An empirical study.
\newblock In {\em International Conference on Learning Representations (ICLR)}, 2025.
\newblock OpenReview (ICLR 2025).

\bibitem{zhang2025training}
Yedi Zhang, Aaditya~K Singh, Peter~E. Latham, and Andrew~M Saxe.
\newblock Training dynamics of in-context learning in linear attention.
\newblock In {\em Forty-second International Conference on Machine Learning}, 2025.

\bibitem{zhao2024probing_decision_boundaries}
Siyan Zhao, Tung Nguyen, and Aditya Grover.
\newblock Probing the decision boundaries of in‑context learning in large language models.
\newblock In {\em NeurIPS (Poster)}, 2024.
\newblock OpenReview (NeurIPS 2024).

\bibitem{zhou-etal-2024-mechanism-survey}
Yuxiang Zhou, Jiazheng Li, Yanzheng Xiang, Hanqi Yan, Lin Gui, and Yulan He.
\newblock The mystery of in-context learning: A comprehensive survey on interpretation and analysis.
\newblock In Yaser Al-Onaizan, Mohit Bansal, and Yun-Nung Chen, editors, {\em Proceedings of the 2024 Conference on Empirical Methods in Natural Language Processing}, pages 14365--14378, Miami, Florida, USA, November 2024. Association for Computational Linguistics.

\end{thebibliography}
    \bibliographystyle{plain}

    \clearpage
    \appendix
    \section{Appendix}
        \subsection{Proof}
    \label{app:proof}

    \subsubsection{Proof of Equation~\ref{equ:single_layer_grad_flow}}
        \begin{proof}
            Based on Equation~\ref{equ:lsa} and \cite{zhang-etal-2024-trained}, the predicted answer of LSA is:
            \[
            \hat{y}_{query} = ((w_{21}^{PV})^{\top} \quad w_{22}^{PV}) \cdot \left(\frac{EE^{\top}}{N}\right) \cdot \begin{pmatrix}W_{11}^{KQ}\\ (w_{21}^{KQ})^{\top}\end{pmatrix} q_x,
            \]
            where $N$ is the number of demonstrations.
            For $N=1$, the matrix product $EE^\top$ can be expressed compactly as $EE^\top = dd^\top + qq^\top$. Substituting this into the equation gives:
            \[
            \hat{y}_{query} = ((w_{21}^{PV})^{\top} \quad w_{22}^{PV}) (dd^{\top} + qq^{\top}) \begin{pmatrix}W_{11}^{KQ}\\ (w_{21}^{KQ})^{\top}\end{pmatrix} q_x
            \]
            We can expand this expression and separate the terms that depend on $d$:
            \begin{align*}
                \hat{y}_{query} = & \underbrace{((w_{21}^{PV})^{\top} \quad w_{22}^{PV}) (dd^{\top}) \begin{pmatrix}W_{11}^{KQ}\\ (w_{21}^{KQ})^{\top}\end{pmatrix} q_x}_{\text{Term depending on } d} + \\
                & \underbrace{((w_{21}^{PV})^{\top} \quad w_{22}^{PV}) (qq^{\top}) \begin{pmatrix}W_{11}^{KQ}\\ (w_{21}^{KQ})^{\top}\end{pmatrix} q_x}_{\text{Constant w.r.t. } d}
            \end{align*}
            The term depending on $d$ can be rewritten using the property of scalar products:
            \begin{align*}
                & ((w_{21}^{PV})^{\top} \quad w_{22}^{PV}) d d^{\top} \begin{pmatrix}W_{11}^{KQ}\\ (w_{21}^{KQ})^{\top}\end{pmatrix} q_x \\
                & = \left( ((w_{21}^{PV})^{\top} \quad w_{22}^{PV}) d \right) \left( d^{\top} \begin{pmatrix}W_{11}^{KQ}\\ (w_{21}^{KQ})^{\top}\end{pmatrix} q_x \right)
            \end{align*}
            This is a quadratic form of the vector $d$. To compute its gradient, we use the vector calculus identity $\nabla_z((a^\top z)(b^\top z)) = a(b^\top z) + b(a^\top z)$. We set:
            \begin{align*}
                a &= \begin{pmatrix} w_{21}^{PV} \\ w_{22}^{PV} \end{pmatrix} \\
                b &= \begin{pmatrix} W_{11}^{KQ} q_x \\ (w_{21}^{KQ})^{\top} q_x \end{pmatrix}
            \end{align*}
            The second term in the expression for $\hat{y}_{query}$ is constant with respect to $d$, so its gradient is zero. Applying the identity to the first term yields the gradient of $\hat{y}_{query}$ with respect to $d$:
            \begin{align*}
                    \nabla_{d} \hat{y}_{query} = & \begin{pmatrix} w_{21}^{PV} \\ w_{22}^{PV} \end{pmatrix} \left( \begin{pmatrix} d_x \\ d_y \end{pmatrix}^{\top} \begin{pmatrix} W_{11}^{KQ} q_x \\ (w_{21}^{KQ})^{\top} q_x \end{pmatrix} \right) + \\
                    & \begin{pmatrix} W_{11}^{KQ} q_x \\ (w_{21}^{KQ})^{\top} q_x \end{pmatrix} \left( \begin{pmatrix} w_{21}^{PV} \\ w_{22}^{PV} \end{pmatrix}^{\top} \begin{pmatrix} d_x \\ d_y \end{pmatrix} \right) \\
                    = &  \begin{pmatrix} w_{21}^{PV} \\ w_{22}^{PV} \end{pmatrix} \left( d_x^{\top} W_{11}^{KQ} q_x + d_y (w_{21}^{KQ})^{\top} q_x \right) + \\
                    & \begin{pmatrix} W_{11}^{KQ} q_x \\ (w_{21}^{KQ})^{\top} q_x \end{pmatrix} \left( (w_{21}^{PV})^{\top} d_x + d_y w_{22}^{PV} \right)
            \end{align*}
            This completes the derivation.
        \end{proof}

    \subsection{Proof of Lemma~\ref{lem:multi_layer_flow}}
        \begin{proof}
            We prove by induction on the layer index $l$.
            
            \paragraph{Base case ($l=0$).}
            The two inequalities in the statement are satisfied by assumption.
            
            \paragraph{Induction step.}
            Assume the inequalities hold for some layer $l-1$ $(1\le l\le L)$, i.e.
            \begin{align*}
                & \bigl\|W^{PV,(l-1)}d_1^{(l-1)}\bigr\| \;\ge\; \bigl\|W^{PV,(l-1)}d_2^{(l-1)}\bigr\| \\
                & \bigl\|\bigl(d_1^{(l-1)}\bigr)^{\!\top}W^{KQ,(l-1)}q^{(l-1)}\bigr\| \ge \\
                & \bigl\|\bigl(d_2^{(l-1)}\bigr)^{\!\top}W^{KQ,(l-1)}q^{(l-1)}\bigr\|.
            \end{align*}
            Apply \eqref{eq:C-matrix-PV} to both demonstrations and use the strict monotonicity of $g_l$:
            \begin{align*}
                & \bigl\|W^{PV,(l)}d_1^{(l)}\bigr\| \\
                & = g_l\!\bigl(\bigl\|W^{PV,(l-1)}d_1^{(l-1)}\bigr\|\bigr) \\
                & \ge g_l\!\bigl(\bigl\|W^{PV,(l-1)}d_2^{(l-1)}\bigr\|\bigr) \\
                & = \bigl\|W^{PV,(l)}d_2^{(l)}\bigr\|.
            \end{align*}
            Similarly, apply \eqref{eq:C-matrix-KQ} and the strict monotonicity of $h_l$:
            \begin{align*}
                & \bigl\|\bigl(d_1^{(l)}\bigr)^{\!\top}W^{KQ,(l)}q^{(l)}\bigr\| \\
                & = h_l\!\bigl(\bigl\|\bigl(d_1^{(l-1)}\bigr)^{\!\top}W^{KQ,(l-1)}q^{(l-1)}\bigr\|\bigr) \\
                & \ge h_l\!\bigl(\bigl\|\bigl(d_2^{(l-1)}\bigr)^{\!\top}W^{KQ,(l-1)}q^{(l-1)}\bigr\|\bigr) \\
                & = \bigl\|\bigl(d_2^{(l)}\bigr)^{\!\top}W^{KQ,(l)}q^{(l)}\bigr\|.
            \end{align*}
            Thus the claim holds for layer $l$.  
            By induction it holds for all layers $l=0,\dots,L$.
        \end{proof}

    \subsubsection{Proof of Theorem~\ref{the:multi_layer_amplifier}}
        \begin{proof}
            Based on Equation~\ref{equ:multi_layer_chain_rule}, we can derive:
            \begin{align*}
                & \frac{\partial \hat{q}_y^{(l_1)}}{\partial d^{(0)}} / \frac{\partial \hat{q}_y^{(l_2)}}{\partial d^{(0)}} \\
                & = \left(\frac{\partial \hat{q}_y^{(l_1)}}{\partial E^{(l_1-1)}} \times \frac{\partial E^{(1)}}{\partial d^{(0)}} \times \prod_{i=2}^{l_1} \frac{\partial E^{(i)}}{\partial E^{(i-1)}} \right) / \\
                & \quad \left(\frac{\partial \hat{q}_y^{(l_2)}}{\partial E^{(l_2-1)}} \times \frac{\partial E^{(1)}}{\partial d^{(0)}} \times  \prod_{i=2}^{l_2} \frac{\partial E^{(i)}}{\partial E^{(i-1)}} \right) \\
                & =\left( \frac{\partial \hat{q}_y^{(l_1)}}{\partial E^{(l_1-1)}} / \frac{\partial \hat{q}_y^{(l_2)}}{\partial E^{(l_2-1)}} \right) \times \prod_{i=l_2 + 1}^{l_1} \frac{\partial E^{(i)}}{\partial E^{(i-1)}}
            \end{align*}

            % 考虑f_LSA关于输入E的全微分，我们可以得到
            Consider the total derivative of $f_{\mathrm{LSA}}$:
            \begin{align*}
                \mathrm{d}f_{\mathrm{LSA}} = & \mathrm{d}E + W^{PV}\,\mathrm{d}E\,E^{\top}W^{KQ}E + \\
                & W^{PV}E \bigl( E^{\top}W^{KQ}\,\mathrm{d}E + \mathrm{d}E^{\top}W^{KQ}E \bigr)
            \end{align*}
            % 可以发现，各个微分项的系数与Definition~\ref{}中的一致，因此由$d^{(l)}_1 \succcurlyeq_{q;\theta} d^{(l)}_2$，我们知道
            It can be observed that the coefficients of the differential terms are consistent with those in Definition~\ref{def:demonstration_effectivness}. 
            Therefore, from $d^{(l)}_1 \succcurlyeq_{q^{(l)};\theta^{(l)}} d^{(l)}_2$, we know that:
            $$\frac{\partial E^{(l)}}{\partial E^{(l-1)}}(E_1;\theta) \geq \frac{\partial E^{(l)}}{\partial E^{(l-1)}}(E_2;\theta), \forall l \in \{l_2+1, \dots, l_1\}$$
            Therefore, we can conclude that:
            $$\frac{\Vert \nabla_{d^{(0)}} \hat{q_y}^{(l_1)}(E_1;\theta) \Vert}{\Vert \nabla_{d^{(0)}} \hat{q_y}^{(l_1)}(E_2;\theta) \Vert} \geq \frac{\Vert \nabla_{d^{(0)}} \hat{q_y}^{(l_2)}(E_1;\theta) \Vert}{\Vert \nabla_{d^{(0)}} \hat{q_y}^{(l_2)}(E_2;\theta) \Vert}$$
        \end{proof}

\subsection{Prompt}
    \label{app:prompt}

    \begin{table}[t]
        \centering
        \small
        \begin{tabular}{p{0.9\linewidth}}
            \toprule
            \textbf{Prompt of Inference} \\
            \midrule
            \{task\} \\
            Below are some examples \\
            \\
            --- \\
            \\
            \{demo\} \\
            \\
            --- \\
            \\
            Based on the above instruction and examples, solve the following problem. \\
            \{question\} \\
            \bottomrule
        \end{tabular}
        \caption{
            The prompt of the inference.
        }
        \label{tab:prompt_synthesis}
    \end{table}

    In this section, we present the inference prompt of our main experiment, as shown in Table~\ref{tab:prompt_synthesis}.
    The task definition we used is same to the previous works \cite{deepseekai2025deepseekr1,grattafiori2024llama3}.

\subsection{Dataset}
    \label{app:dataset}

    \begin{table}[ht]
        \centering
        \small
        \begin{tabular}{l|cc}
            \toprule
            \textbf{Dataset} & \textbf{Test Set} & \textbf{Demonstration} \\
            \midrule
            GSM8K & $1319$ & $7473$ \\
            MATH & $500$ & $7496$ \\
            ARC-Challenge & $1172$ & $1119$ \\
            MMLU-Pro & $1000$ & $70$ \\
            Amazon Review & $200$ & $1800$ \\
            \bottomrule
        \end{tabular}
        \caption{
            The scales of test set and demonstrations of each dataset.
        }
        \label{tab:dataset_scale}
    \end{table}

    In this section, we detail the datasets used in our study. 
    Table~\ref{tab:dataset_scale} summarizes the scale of the test set and demonstrations for each.

    \paragraph{GSM8K}
        GSM8K~\cite{cobbe2021gsm8k} is a high-quality collection of elementary school-level math problems. 
        We utilize its training set directly as the demonstration pool.

    \paragraph{MATH}
        The MATH dataset~\cite{hendrycks2021math} consists of challenging high school competition-level math problems in fields like algebra, probability, and geometry. 
        Following the approach of \cite{lightman2024lets}, we evaluate \ourmethod on a random sample of $500$ problems. 
        The demonstrations are drawn from the official training set.

    \paragraph{ARC-Challenge}
        The ARC-Challenge~\cite{yadav-etal-2019-arc} is a question-answering dataset with difficult, science-focused questions. 
        For this dataset, the training set is used as our demonstration pool.
    
    \paragraph{MMLU-Pro}
        MMLU-Pro~\cite{wang2024mmlupro} serves as a multi-task benchmark for the comprehensive evaluation of LLMs on professional domain knowledge and complex reasoning. 
        As the dataset is only divided into validation and test sets, we use the validation set as our demonstration pool and conduct evaluations on the test set.

    \paragraph{Amazon Review}
        The Amazon Review dataset~\cite{ni-etal-2019-amazon}, containing a vast amount of user ratings and reviews, is widely used for research in sentiment analysis and recommender systems. 
        Due to the immense size, we select the \textit{Health and Personal Care} category for testing, while using the \textit{All Beauty}, \textit{Digital Music}, and \textit{Software} categories to form the demonstration pool.

\subsection{Baseline}
    \label{app:baseline}

    \paragraph{BM25}
        BM25 is a classic sparse retrieval method based on the probabilistic relevance framework, serving as an extension of TF-IDF. 
        In the context of ICL, it treats the test query as a search query and the candidate demonstrations as documents. 
        It ranks demonstrations by scoring them based on the frequency and distribution of query terms, without considering word order or deep semantics. 
        The top-K scored demonstrations are then selected as the in-context demonstrations. 
        This ``bag-of-words'' approach is known for its computational efficiency and serves as a strong baseline.

    \paragraph{Cosine Similarity}
        Cosine Similarity is a popular and effective strategy for ICL demonstration selection that aims to find demonstrations semantically similar to the test query. 
        This method involves encoding the test query and candidate demonstrations into high-dimensional vectors (embeddings) using a pre-trained sentence encoder, such as BERT. 
        The semantic relevance between the query and each demonstration is then measured by computing the cosine similarity of their respective vectors. 
        The demonstrations with the highest similarity scores are selected to form the context. This dense retrieval approach generally captures semantic nuances better than sparse methods like BM25, but its performance depends on the quality of the underlying embedding model.

    \paragraph{MMR}
        Maximal Marginal Relevance (MMR) is a selection strategy designed to balance the relevance of demonstrations to the query with the diversity within the selected set. 
        The rationale is that selecting only the nearest neighbors (most relevant examples) can result in a set of overly similar demonstrations, which may limit the variety of reasoning processes shown to the model. 
        MMR addresses this by iteratively selecting demonstrations that maximize a combined score of relevance to the query and dissimilarity from the examples already chosen. 
        This approach aims to create a set of demonstrations that are not only relevant but also complementary, using diversity as a proxy for complementarity to improve ICL performance.

    \paragraph{MoD}
        Mixture of Demonstrations (MoD) is a framework designed to overcome the challenges of a large search space and suboptimal retriever optimization in ICL demonstration selection. 
        The core idea is to partition the entire demonstration pool into distinct groups, typically using K-means clustering on sentence embeddings. 
        Each group is governed by a dedicated ``expert'', a unique retriever model trained specifically for that partition. 
        During inference, these experts collaboratively retrieve demonstrations for a given query, with the final set being an aggregation of examples selected by the most relevant experts. 
        This ``mixture of experts'' approach reduces the search complexity while ensuring the selected demonstrations are diverse and effective.

\subsection{Additional Experiment Results}
    \subsubsection{Scale of Effective and Ineffective Data}
        \label{app:scale_effective_ineffective}
        \begin{table}[ht]
            \small
            \centering
            \begin{tabular}{llrr}
    \toprule
    \textbf{Model} & \textbf{Dataset} & \textbf{Effective} & \textbf{Ineffective} \\
    \midrule
    \multirow{5}{*}{\textbf{Llama2-7b}} & GSM8K & $225$ & $1030$ \\
    & MATH & $18$ & $452$ \\
    & ARC-Challenge & $355$ & $674$ \\
    & MMLU-Pro & $108$ & $843$ \\
    & Amazon & $38$ & $134$ \\
    \midrule
    \multirow{5}{*}{\textbf{Llama3.1-8b}} & GSM8K & $84$ & $210$ \\
    & MATH & $52$ & $280$ \\
    & ARC-Challenge & $86$ & $212$ \\
    & MMLU-Pro & $174$ & $466$ \\
    & Amazon & $26$ & $63$ \\
    \midrule
    \multirow{5}{*}{\textbf{Llama-R1-8b}} & GSM8K & $41$ & $519$ \\
    & MATH & $27$ & $146$ \\
    & ARC-Challenge & $77$ & $198$ \\
    & MMLU-Pro & $203$ & $550$ \\
    & Amazon & $20$ & $102$ \\
    \midrule
    \multirow{5}{*}{\textbf{Qwen3-8b}} & GSM8K & $33$ & $99$ \\
    & MATH & $31$ & $111$ \\
    & ARC-Challenge & $28$ & $110$ \\
    & MMLU-Pro & $251$ & $368$ \\
    & Amazon & $23$ & $77$ \\
    \bottomrule
\end{tabular}
            \caption{
                The number of effective and ineffective demonstrations under each setting.
            }
            \label{tab:demonstration_count}
        \end{table}
    
        The number of effective and ineffective demonstrations we sampled is shown in Table~\ref{tab:demonstration_count}.

    \subsubsection{Detailed Gradient Flow under Each Setting}
        \label{app:detailed_gradient_flow}
        \begin{figure}
    \small
    \centering
    \begin{tikzpicture}
    \begin{axis}[
        % --- 左侧 Y 轴 ---
        axis y line*=left,
        xlabel={Layer},
        % 修复 2: 手动在 Y 轴标签中添加比例因子
        ylabel={Gradient Flow}, 
        ymin=0, ymax=4e-6,
        width=0.9\linewidth,
        grid=major,
        grid style=dashed,
        % 修复 2: 手动设置刻度位置和标签文本
        ytick={0, 1e-6, 2e-6, 3e-6, 4e-6},
        yticklabels={0.0, 1.0, 2.0, 3.0, 4.0},
        yticklabel style={/pgf/number format/fixed, /pgf/number format/precision=1},
    ]
        % 为了让图例能够引用，我们只绘制曲线，图例在第二个axis环境中统一处理
        \addplot[softred, mark=*, sharp plot, thick] coordinates {
            (0, 1.3042571739501968e-06)
            (5, 1.3748078711248493e-06)
            (10, 8.811462954377198e-07)
            (15, 1.337741105975534e-06)
            (20, 2.177433805025123e-06)
            (25, 1.6552984947067034e-06)
            (30, 2.395923606352928e-06)
            (31, 3.51724693629947e-06)
        };
        \addplot[softblue, mark=square*, sharp plot, thick] coordinates {
            (0, 1.3303741178347745e-06)
            (5, 1.3170297863463877e-06)
            (10, 8.401473159293678e-07)
            (15, 1.278038830349829e-06)
            (20, 2.116395790454277e-06)
            (25, 1.5362733269853532e-06)
            (30, 2.2861196062953547e-06)
            (31, 3.2999101012511326e-06)
        };
    \end{axis}

    \begin{axis}[
        % --- 右侧 Y 轴 ---
        % 修复 3: "axis y line*=right" 是正确的命令，确保Y轴和标签在右侧
        axis y line*=right,
        axis x line=none, % 不重复绘制X轴
        ylabel={Ratio ($\log_{10}$)},
        ymin=-0.2, ymax=0.6,
        width=0.9\linewidth,
        % --- 图例 ---
        legend style={
            at={(0.5, -0.2)}, % 稍微调高位置
            anchor=north,
            legend columns=3,
            legend cell align=left,
        }
    ]
        % --- 修复 1: 手动为所有图例条目指定样式 ---
        % 我们需要一个“占位”的addplot，因为addlegendimage必须在addplot之后
        % 这个占位图本身并不会被画出来
        \addplot[draw=none, forget plot] coordinates {(0,0)};
        
        % 手动添加第一个图例，直接指定样式
        \addlegendimage{softred, mark=*, sharp plot, thick}
        \addlegendentry{Effective}
        
        % 手动添加第二个图例，直接指定样式
        \addlegendimage{softblue, mark=square*, sharp plot, thick}
        \addlegendentry{Ineffective}
        
        % 绘制第三条曲线，并为其添加图例
        \addplot[softgreen, mark=triangle*, sharp plot, thick] coordinates {
            (0, -0.00861055298141214)
            (5, 0.06734676857440718)
            (10, 0.07377168059166883)
            (15, 0.16639236316050712)
            (20, 0.2504378806639889)
            (25, 0.38994671610418946)
            (30, 0.5259583571679174)
            (31, 0.5536591085376052)
        };
        \addlegendentry{Ratio} % 这个会自动匹配上一条曲线（绿色）

    \end{axis}
\end{tikzpicture}
    \caption{
        The average gradient flow (left y-axis) and the ratio between the effective and the ineffective (right y-axis) under each layer of Llama2-7b on GSM8K.
    }
    \label{fig:grad_flow_llama2_7b_gsm8k}
\end{figure}
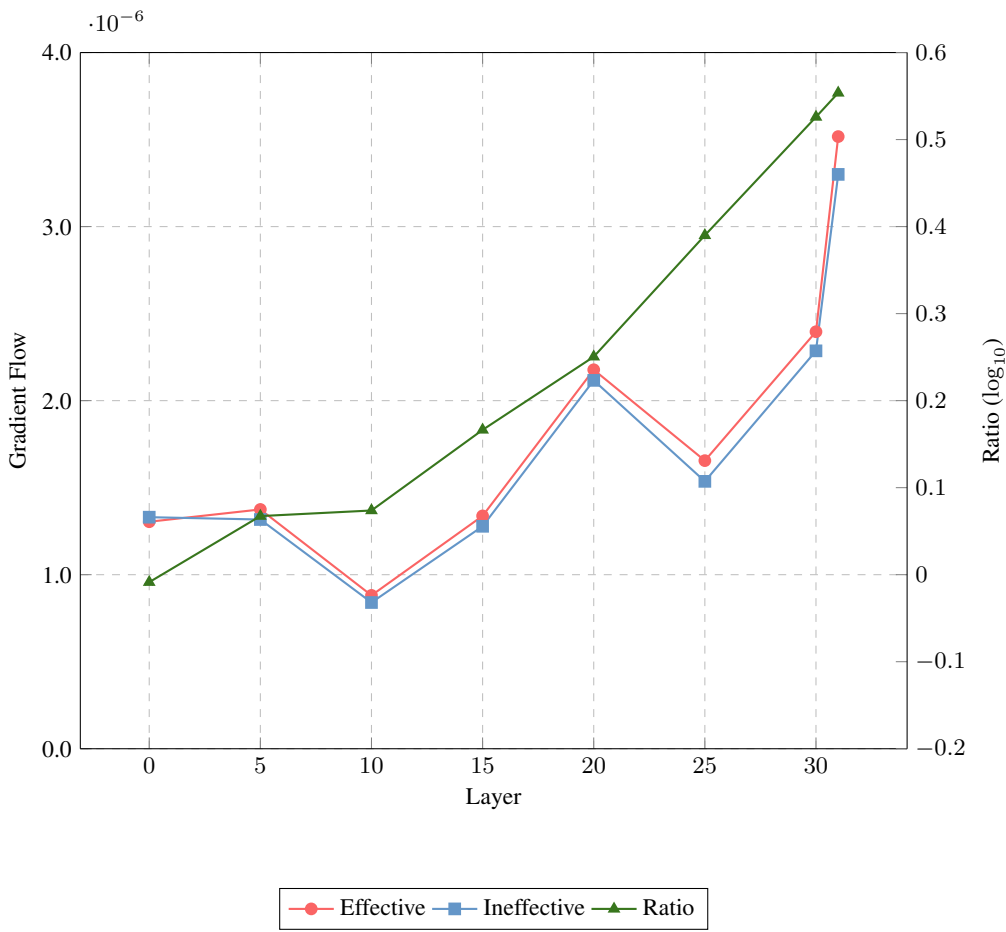

\begin{figure}
    \small
    \centering
    \begin{tikzpicture}
    \begin{axis}[
        % --- 左侧 Y 轴 ---
        axis y line*=left,
        xlabel={Layer},
        ylabel={Gradient Flow}, 
        ymin=0, ymax=5e-6, % 调整 Ymax
        width=0.9\linewidth,
        grid=major,
        grid style=dashed,
        % 调整刻度
        ytick={0, 1e-6, 2e-6, 3e-6, 4e-6, 5e-6},
        yticklabels={0.0, 1.0, 2.0, 3.0, 4.0, 5.0},
        yticklabel style={/pgf/number format/fixed, /pgf/number format/precision=1},
    ]
        % 更新 "Effective" 数据
        \addplot[softred, mark=*, sharp plot, thick] coordinates {
            (0, 1.7904324077716114e-06)
            (5, 2.1442633908463904e-06)
            (10, 1.9083678211018196e-06)
            (15, 3.7126875298554296e-06)
            (20, 2.348637593740932e-06)
            (25, 1.7109578800363486e-06)
            (30, 2.9289985832292587e-06)
            (31, 4.042236412260536e-06)
        };
        % 更新 "Ineffective" 数据
        \addplot[softblue, mark=square*, sharp plot, thick] coordinates {
            (0, 1.4822970655359313e-06)
            (5, 1.7926364646304552e-06)
            (10, 1.5515075107780184e-06)
            (15, 2.302106618761019e-06)
            (20, 2.1878700985358724e-06)
            (25, 1.432831680858726e-06)
            (30, 2.8288279278272e-06)
            (31, 4.1487771787582326e-06)
        };
    \end{axis}

    \begin{axis}[
        % --- 右侧 Y 轴 ---
        axis y line*=right,
        axis x line=none, % 不重复绘制X轴
        ylabel={Ratio ($\log_{10}$)},
        ymin=-0.2, ymax=2.0, % 显著调整 Ymax
        width=0.9\linewidth,
        % --- 图例 ---
        legend style={
            at={(0.5, -0.2)},
            anchor=north,
            legend columns=3,
            legend cell align=left,
        }
    ]
        % 占位图
        \addplot[draw=none, forget plot] coordinates {(0,0)};
        
        % 图例条目
        \addlegendimage{softred, mark=*, sharp plot, thick}
        \addlegendentry{Effective}
        
        \addlegendimage{softblue, mark=square*, sharp plot, thick}
        \addlegendentry{Ineffective}
        
        % 更新 "Ratio" 数据
        \addplot[softgreen, mark=triangle*, sharp plot, thick] coordinates {
            (0, -0.08202268136607485)
            (5, 0.2509935709276418)
            (10, 0.6426620176083917)
            (15, 1.2911081004771743)
            (20, 1.4033214033636054)
            (25, 1.5781057619680992)
            (30, 1.8305313104750203)
            (31, 1.841829711119998)
        };
        \addlegendentry{Ratio}

    \end{axis}
\end{tikzpicture}
    \caption{
        The average gradient flow (left y-axis) and the ratio between the effective and the ineffective (right y-axis) under each layer of Llama2-7b on MATH.
    }
    \label{fig:grad_flow_llama2_7b_math}
\end{figure}
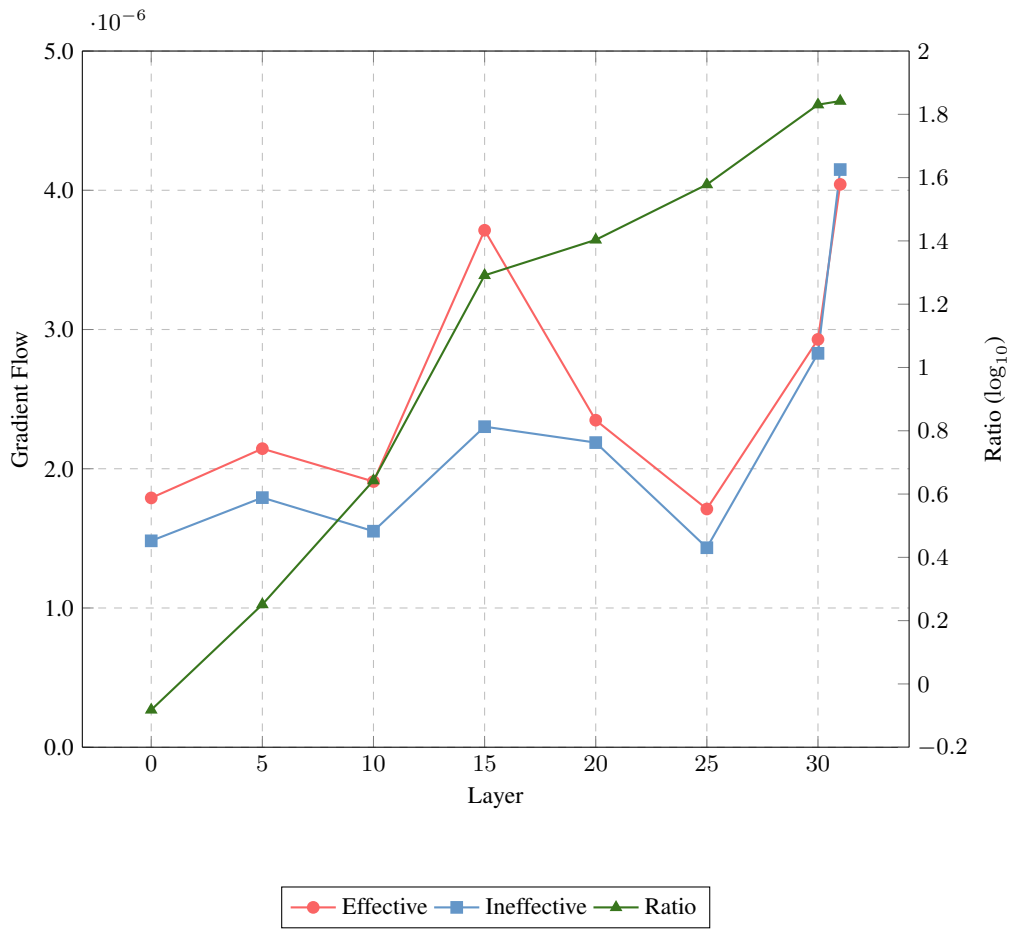

\begin{figure}
    \small
    \centering
    \begin{tikzpicture}
    \begin{axis}[
        % --- 左侧 Y 轴 ---
        axis y line*=left,
        xlabel={Layer},
        ylabel={Gradient Flow}, 
        ymin=0, ymax=3e-6, % 调整 Ymax
        width=0.9\linewidth,
        grid=major,
        grid style=dashed,
        yticklabels={0.0, 1.0, 2.0, 3.0, 4.0, 5.0},
        yticklabel style={/pgf/number format/fixed, /pgf/number format/precision=1},
    ]
        % 更新 "Effective" 数据
        \addplot[softred, mark=*, sharp plot, thick] coordinates {
            (0, 7.910631098725826e-07)
            (5, 7.25574004984898e-07)
            (10, 7.264505983068133e-07)
            (15, 6.344321525564341e-07)
            (20, 9.09511438279328e-07)
            (25, 8.241421544912606e-07)
            (30, 2.3395638938912292e-06)
            (31, 1.6431432402064157e-06)
        };
        % 更新 "Ineffective" 数据
        \addplot[softblue, mark=square*, sharp plot, thick] coordinates {
            (0, 7.432507362163202e-07)
            (5, 6.636555510932808e-07)
            (10, 6.832807017291763e-07)
            (15, 6.005350132866307e-07)
            (20, 8.220615971378141e-07)
            (25, 7.987858537310812e-07)
            (30, 2.2676736394390846e-06)
            (31, 1.5035797094390435e-06)
        };
    \end{axis}

    \begin{axis}[
        % --- 右侧 Y 轴 ---
        axis y line*=right,
        axis x line=none, % 不重复绘制X轴
        ylabel={Ratio ($\log_{10}$)},
        ymin=0, ymax=1.0, % 显著调整 Ymax
        width=0.9\linewidth,
        % --- 图例 ---
        legend style={
            at={(0.5, -0.2)},
            anchor=north,
            legend columns=3,
            legend cell align=left,
        }
    ]
        % 占位图
        \addplot[draw=none, forget plot] coordinates {(0,0)};
        
        % 图例条目
        \addlegendimage{softred, mark=*, sharp plot, thick}
        \addlegendentry{Effective}
        
        \addlegendimage{softblue, mark=square*, sharp plot, thick}
        \addlegendentry{Ineffective}
        
        % 更新 "Ratio" 数据
        \addplot[softgreen, mark=triangle*, sharp plot, thick] coordinates {
            (0, 0.027075784205101688)
            (5, 0.20347356680956363)
            (10, 0.3358453717780459)
            (15, 0.47510483256087876)
            (20, 0.5981795243383639)
            (25, 0.6599355698457758)
            (30, 0.7707588492022568)
            (31, 0.8093078173474558)
        };
        \addlegendentry{Ratio}

    \end{axis}
\end{tikzpicture}
    \caption{
        The average gradient flow (left y-axis) and the ratio between the effective and the ineffective (right y-axis) under each layer of Llama2-7b on ARC-Challenge.
    }
    \label{fig:grad_flow_llama2_7b_arc}
\end{figure}
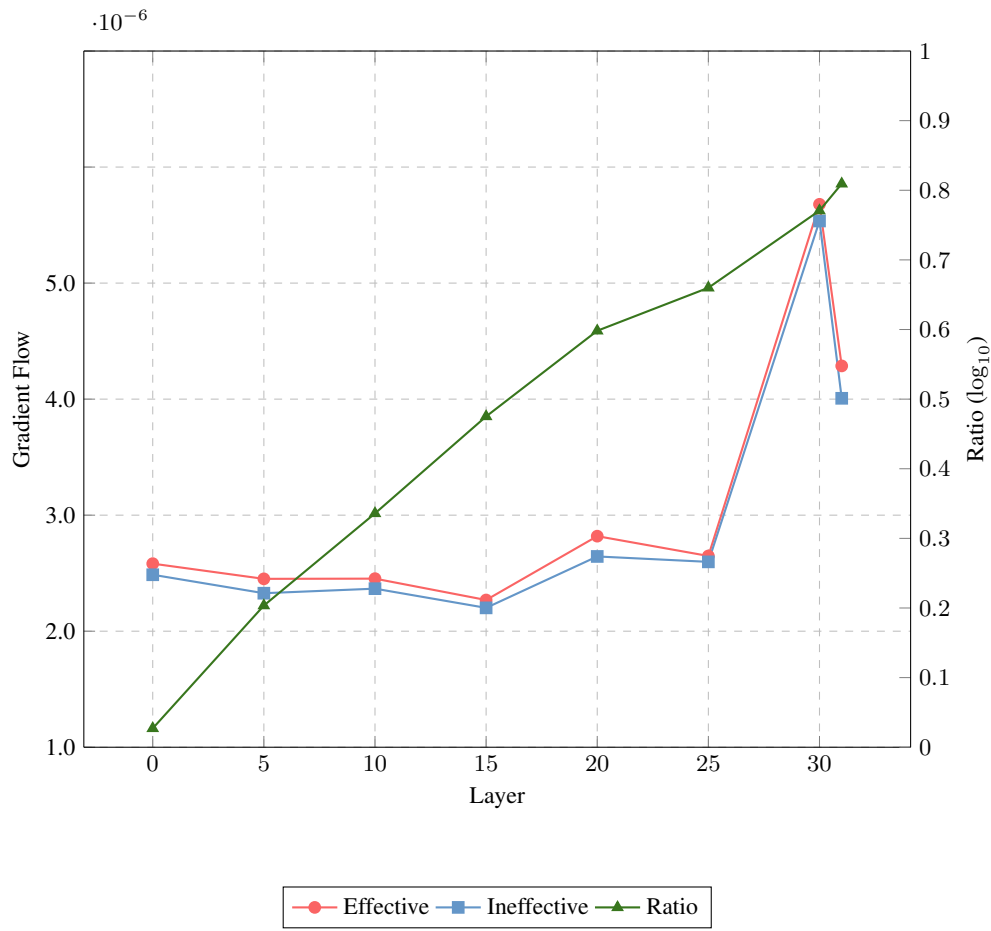

\begin{figure}
    \small
    \centering
    \begin{tikzpicture}
    \begin{axis}[
        % --- 左侧 Y 轴 ---
        axis y line*=left,
        xlabel={Layer},
        ylabel={Gradient Flow}, 
        ymin=0, ymax=5e-6, % 调整 Ymax
        width=0.9\linewidth,
        grid=major,
        grid style=dashed,
        % 调整刻度
        ytick={0, 1e-6, 2e-6, 3e-6, 4e-6, 5e-6},
        yticklabels={0.0, 1.0, 2.0, 3.0, 4.0, 5.0},
        yticklabel style={/pgf/number format/fixed, /pgf/number format/precision=1},
    ]
        % 更新 "Effective" 数据
        \addplot[softred, mark=*, sharp plot, thick] coordinates {
            (0, 1.7904324077716114e-06)
            (5, 2.1442633908463904e-06)
            (10, 1.9083678211018196e-06)
            (15, 3.7126875298554296e-06)
            (20, 2.348637593740932e-06)
            (25, 1.7109578800363486e-06)
            (30, 2.9289985832292587e-06)
            (31, 4.042236412260536e-06)
        };
        % 更新 "Ineffective" 数据
        \addplot[softblue, mark=square*, sharp plot, thick] coordinates {
            (0, 1.4822970655359313e-06)
            (5, 1.7926364646304552e-06)
            (10, 1.5515075107780184e-06)
            (15, 2.302106618761019e-06)
            (20, 2.1878700985358724e-06)
            (25, 1.432831680858726e-06)
            (30, 2.8288279278272e-06)
            (31, 4.1487771787582326e-06)
        };
    \end{axis}

    \begin{axis}[
        % --- 右侧 Y 轴 ---
        axis y line*=right,
        axis x line=none, % 不重复绘制X轴
        ylabel={Ratio ($\log_{10}$)},
        ymin=-0.2, ymax=2.0, % 显著调整 Ymax
        width=0.9\linewidth,
        % --- 图例 ---
        legend style={
            at={(0.5, -0.2)},
            anchor=north,
            legend columns=3,
            legend cell align=left,
        }
    ]
        % 占位图
        \addplot[draw=none, forget plot] coordinates {(0,0)};
        
        % 图例条目
        \addlegendimage{softred, mark=*, sharp plot, thick}
        \addlegendentry{Effective}
        
        \addlegendimage{softblue, mark=square*, sharp plot, thick}
        \addlegendentry{Ineffective}
        
        % 更新 "Ratio" 数据
        \addplot[softgreen, mark=triangle*, sharp plot, thick] coordinates {
            (0, -0.08202268136607485)
            (5, 0.2509935709276418)
            (10, 0.6426620176083917)
            (15, 1.2911081004771743)
            (20, 1.4033214033636054)
            (25, 1.5781057619680992)
            (30, 1.8305313104750203)
            (31, 1.841829711119998)
        };
        \addlegendentry{Ratio}

    \end{axis}
\end{tikzpicture}
    \caption{
        The average gradient flow (left y-axis) and the ratio between the effective and the ineffective (right y-axis) under each layer of Llama2-7b on MMLU-Pro.
    }
    \label{fig:grad_flow_llama2_7b_mmlu}
\end{figure}
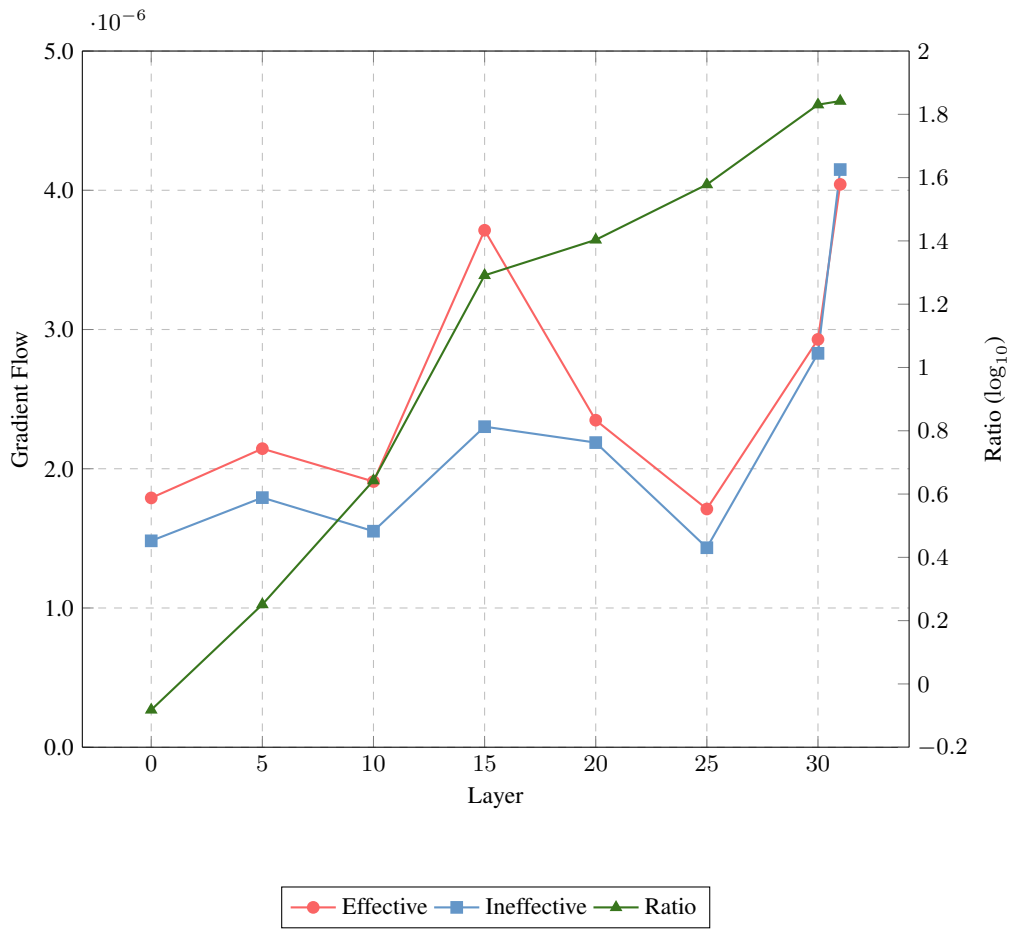

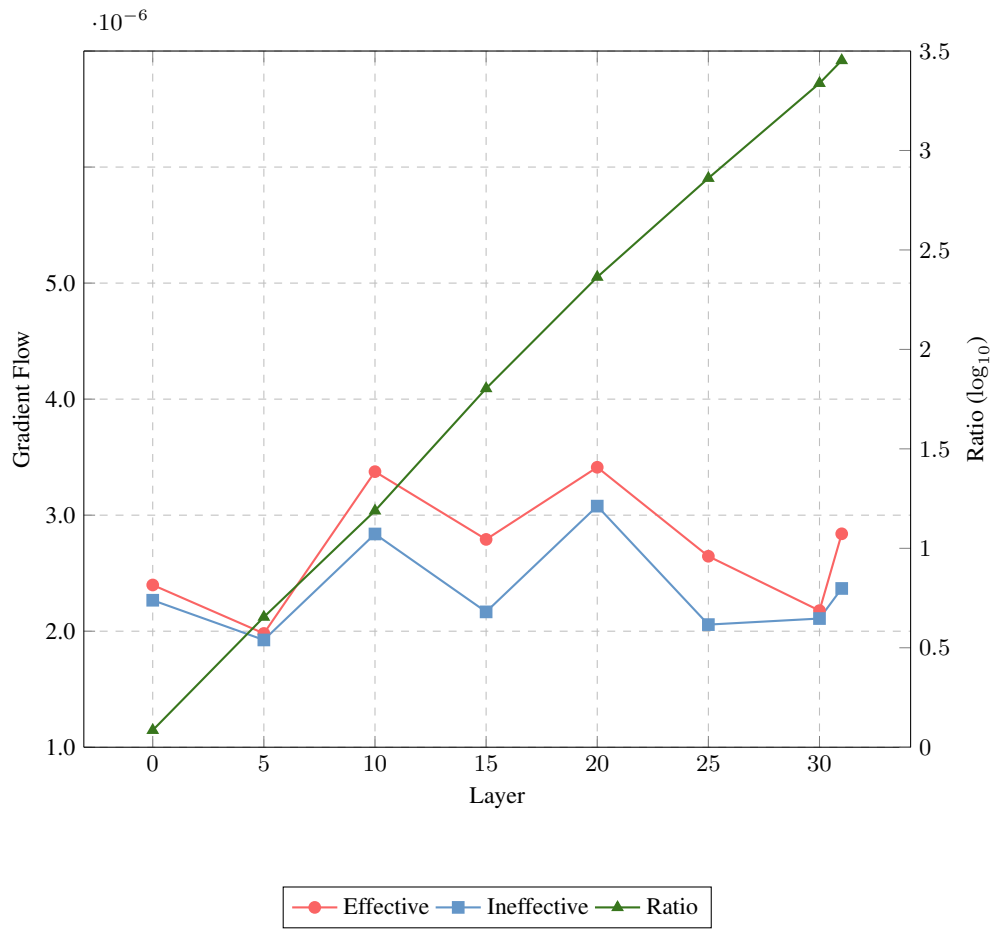
\begin{figure}
    \small
    \centering
    \begin{tikzpicture}
    \begin{axis}[
        % --- 左侧 Y 轴 ---
        axis y line*=left,
        xlabel={Layer},
        ylabel={Gradient Flow}, 
        ymin=0, ymax=3e-6, % 调整 Ymax
        width=0.9\linewidth,
        grid=major,
        grid style=dashed,
        yticklabels={0.0, 1.0, 2.0, 3.0, 4.0, 5.0},
        yticklabel style={/pgf/number format/fixed, /pgf/number format/precision=1},
    ]
        % 更新 "Effective" 数据
        \addplot[softred, mark=*, sharp plot, thick] coordinates {
            (0, 6.991558262353564e-07)
            (5, 4.902734538815349e-07)
            (10, 1.1875337174902405e-06)
            (15, 8.955186074559819e-07)
            (20, 1.2063895018860949e-06)
            (25, 8.229383509609982e-07)
            (30, 5.88662592774551e-07)
            (31, 9.196996784261183e-07)
        };
        % 更新 "Ineffective" 数据
        \addplot[softblue, mark=square*, sharp plot, thick] coordinates {
            (0, 6.333795120196472e-07)
            (5, 4.6226776252814887e-07)
            (10, 9.189283987673173e-07)
            (15, 5.832312885912014e-07)
            (20, 1.038983290645893e-06)
            (25, 5.284441589389526e-07)
            (30, 5.545124204721145e-07)
            (31, 6.841462333102533e-07)
        };
    \end{axis}

    \begin{axis}[
        % --- 右侧 Y 轴 ---
        axis y line*=right,
        axis x line=none, % 不重复绘制X轴
        ylabel={Ratio ($\log_{10}$)},
        ymin=0, ymax=3.5, % 显著调整 Ymax
        width=0.9\linewidth,
        % --- 图例 ---
        legend style={
            at={(0.5, -0.2)},
            anchor=north,
            legend columns=3,
            legend cell align=left,
        }
    ]
        % 占位图
        \addplot[draw=none, forget plot] coordinates {(0,0)};
        
        % 图例条目
        \addlegendimage{softred, mark=*, sharp plot, thick}
        \addlegendentry{Effective}
        
        \addlegendimage{softblue, mark=square*, sharp plot, thick}
        \addlegendentry{Ineffective}
        
        % 更新 "Ratio" 数据
        \addplot[softgreen, mark=triangle*, sharp plot, thick] coordinates {
            (0, 0.08536698095260085)
            (5, 0.6543650437223602)
            (10, 1.1885924616229941)
            (15, 1.8035231378130607)
            (20, 2.3635347644151836)
            (25, 2.8606716992143117)
            (30, 3.3379622760821444)
            (31, 3.4527037592800776)
        };
        \addlegendentry{Ratio}

    \end{axis}
\end{tikzpicture}
    \caption{
        The average gradient flow (left y-axis) and the ratio between the effective and the ineffective (right y-axis) under each layer of Llama2-7b on Amazon Review.
    }
    \label{fig:grad_flow_llama2_7b_amazon}
\end{figure}

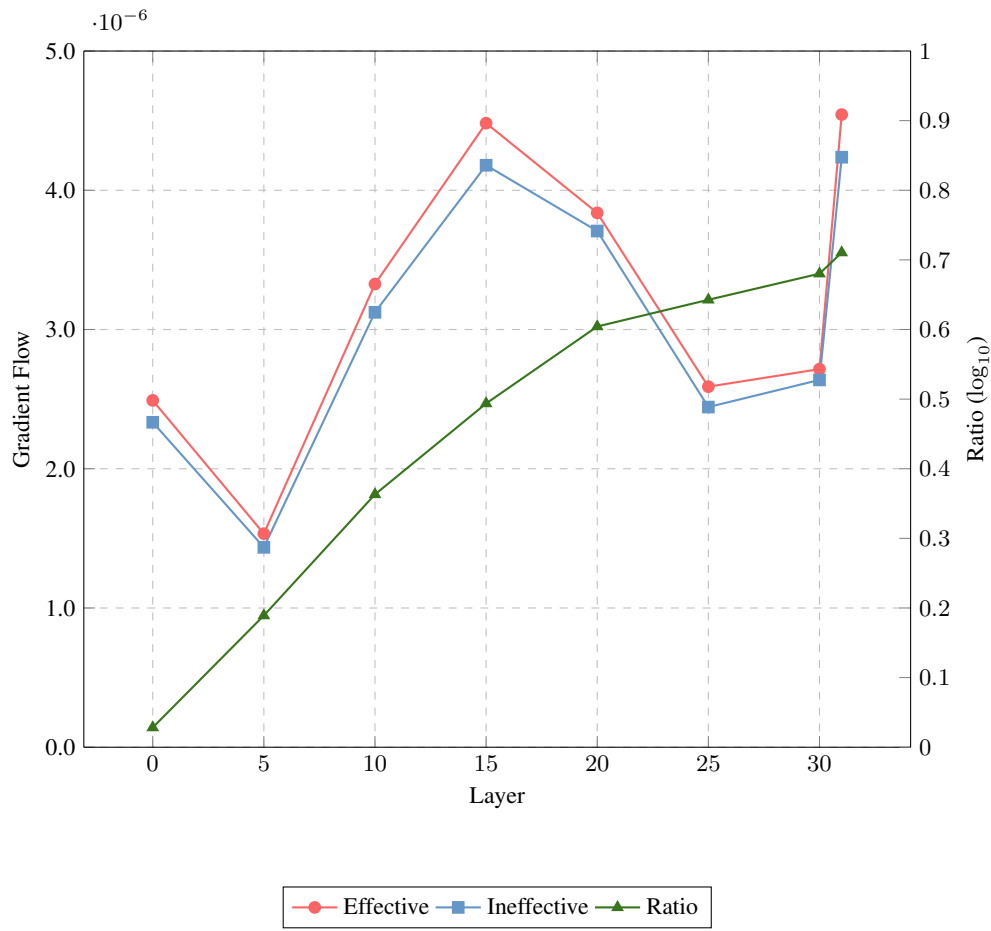
\begin{figure}
    \small
    \centering
    \begin{tikzpicture}
    \begin{axis}[
        % --- 左侧 Y 轴 ---
        axis y line*=left,
        xlabel={Layer},
        ylabel={Gradient Flow}, 
        ymin=0, ymax=5e-6, % 调整 Ymax
        width=0.9\linewidth,
        grid=major,
        grid style=dashed,
        % 调整刻度
        ytick={0, 1e-6, 2e-6, 3e-6, 4e-6, 5e-6},
        yticklabels={0.0, 1.0, 2.0, 3.0, 4.0, 5.0},
        yticklabel style={/pgf/number format/fixed, /pgf/number format/precision=1},
    ]
        % 更新 "Effective" 数据
        \addplot[softred, mark=*, sharp plot, thick] coordinates {
            (0, 2.4908619571725958e-06)
            (5, 1.5336684511770325e-06)
            (10, 3.326613539437281e-06)
            (15, 4.481976782736515e-06)
            (20, 3.8376467006610956e-06)
            (25, 2.5894503291003557e-06)
            (30, 2.715486569553167e-06)
            (31, 4.543845484214706e-06)
        };
        % 更新 "Ineffective" 数据
        \addplot[softblue, mark=square*, sharp plot, thick] coordinates {
            (0, 2.33377684415765e-06)
            (5, 1.435650931723233e-06)
            (10, 3.1237244846644378e-06)
            (15, 4.179725745448845e-06)
            (20, 3.7073625811618036e-06)
            (25, 2.4430667865869637e-06)
            (30, 2.6378691952036855e-06)
            (31, 4.237533377569153e-06)
        };
    \end{axis}

    \begin{axis}[
        % --- 右侧 Y 轴 ---
        axis y line*=right,
        axis x line=none, % 不重复绘制X轴
        ylabel={Ratio ($\log_{10}$)},
        ymin=0, ymax=1, % 显著调整 Ymax
        width=0.9\linewidth,
        % --- 图例 ---
        legend style={
            at={(0.5, -0.2)},
            anchor=north,
            legend columns=3,
            legend cell align=left,
        }
    ]
        % 占位图
        \addplot[draw=none, forget plot] coordinates {(0,0)};
        
        % 图例条目
        \addlegendimage{softred, mark=*, sharp plot, thick}
        \addlegendentry{Effective}
        
        \addlegendimage{softblue, mark=square*, sharp plot, thick}
        \addlegendentry{Ineffective}
        
        % 更新 "Ratio" 数据
        \addplot[softgreen, mark=triangle*, sharp plot, thick] coordinates {
            (0, 0.028290333295065813)
            (5, 0.18915078285115575)
            (10, 0.36309581900512294)
            (15, 0.4937069546975641)
            (20, 0.6043008984624403)
            (25, 0.6426674797957861)
            (30, 0.6801826906623907)
            (31, 0.7104931132004556)
        };
        \addlegendentry{Ratio}

    \end{axis}
\end{tikzpicture}
    \caption{
        The average gradient flow (left y-axis) and the ratio between the effective and the ineffective (right y-axis) under each layer of Llama3.1-8b on GSM8K.
    }
    \label{fig:grad_flow_llama31_8b_gsm8k}
\end{figure}

\begin{figure}
    \small
    \centering
    \begin{tikzpicture}
    \begin{axis}[
        % --- 左侧 Y 轴 ---
        axis y line*=left,
        xlabel={Layer},
        ylabel={Gradient Flow}, 
        ymin=0, ymax=5e-6, % 调整 Ymax
        width=0.9\linewidth,
        grid=major,
        grid style=dashed,
        % 调整刻度
        ytick={0, 1e-6, 2e-6, 3e-6, 4e-6, 5e-6},
        yticklabels={0.0, 1.0, 2.0, 3.0, 4.0, 5.0},
        yticklabel style={/pgf/number format/fixed, /pgf/number format/precision=1},
    ]
        % 更新 "Effective" 数据
        \addplot[softred, mark=*, sharp plot, thick] coordinates {
            (0, 2.288379695115435e-06)
            (5, 1.6302775265455926e-06)
            (10, 1.6497437389155178e-06)
            (15, 3.4245672668243878e-06)
            (20, 3.359320068776302e-06)
            (25, 2.5811286936728647e-06)
            (30, 3.1142083511143453e-06)
            (31, 3.566694489359643e-06)
        };
        % 更新 "Ineffective" 数据
        \addplot[softblue, mark=square*, sharp plot, thick] coordinates {
            (0, 1.814031406933329e-06)
            (5, 1.1362805823880782e-06)
            (10, 1.2324295107269162e-06)
            (15, 2.4168130834200534e-06)
            (20, 2.3570933755765847e-06)
            (25, 1.9786925198360044e-06)
            (30, 2.3923096740963526e-06)
            (31, 2.689711717132996e-06)
        };
    \end{axis}

    \begin{axis}[
        % --- 右侧 Y 轴 ---
        axis y line*=right,
        axis x line=none, % 不重复绘制X轴
        ylabel={Ratio ($\log_{10}$)},
        ymin=0, ymax=5, % 显著调整 Ymax
        width=0.9\linewidth,
        % --- 图例 ---
        legend style={
            at={(0.5, -0.2)},
            anchor=north,
            legend columns=3,
            legend cell align=left,
        }
    ]
        % 占位图
        \addplot[draw=none, forget plot] coordinates {(0,0)};
        
        % 图例条目
        \addlegendimage{softred, mark=*, sharp plot, thick}
        \addlegendentry{Effective}
        
        \addlegendimage{softblue, mark=square*, sharp plot, thick}
        \addlegendentry{Ineffective}
        
        % 更新 "Ratio" 数据
        \addplot[softgreen, mark=triangle*, sharp plot, thick] coordinates {
            (0, 0.10088328372342513)
            (5, 0.7780386403899239)
            (10, 1.521870645972826)
            (15, 2.2796571316920313)
            (20, 2.9162755430355047)
            (25, 3.46634574234446)
            (30, 4.060125363244765)
            (31, 4.182685539137614)
        };
        \addlegendentry{Ratio}

    \end{axis}
\end{tikzpicture}
    \caption{
        The average gradient flow (left y-axis) and the ratio between the effective and the ineffective (right y-axis) under each layer of Llama3.1-8b on MATH.
    }
    \label{fig:grad_flow_llama31_8b_math}
\end{figure}
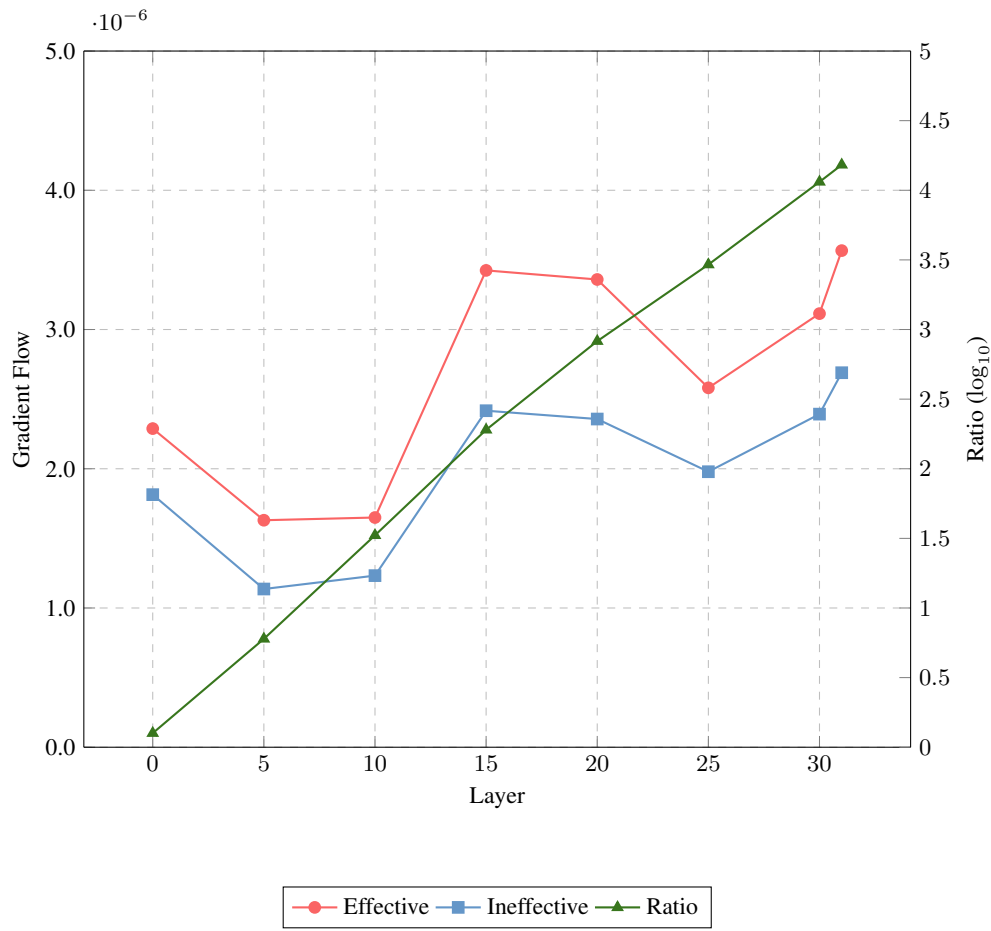

\begin{figure}
    \small
    \centering
    \begin{tikzpicture}
    \begin{axis}[
        % --- 左侧 Y 轴 ---
        axis y line*=left,
        xlabel={Layer},
        ylabel={Gradient Flow}, 
        ymin=0, ymax=1e-5, % 调整 Ymax
        width=0.9\linewidth,
        grid=major,
        grid style=dashed,
        yticklabel style={/pgf/number format/fixed, /pgf/number format/precision=1},
    ]
        % 更新 "Effective" 数据
        \addplot[softred, mark=*, sharp plot, thick] coordinates {
            (0, 5.5138033142714635e-06)
            (5, 4.095666032301195e-06)
            (10, 4.710304259743199e-06)
            (15, 9.04877481979006e-06)
            (20, 6.7356586361308116e-06)
            (25, 6.804783721828193e-06)
            (30, 4.876363348939604e-06)
            (31, 6.538416509432143e-06)
        };
        % 更新 "Ineffective" 数据
        \addplot[softblue, mark=square*, sharp plot, thick] coordinates {
            (0, 4.713993927569306e-06)
            (5, 3.4338856572281486e-06)
            (10, 3.3113449976448212e-06)
            (15, 6.808295927649577e-06)
            (20, 6.275446036513956e-06)
            (25, 5.249846385912713e-06)
            (30, 3.9478824945479915e-06)
            (31, 4.929274545046673e-06)
        };
    \end{axis}

    \begin{axis}[
        % --- 右侧 Y 轴 ---
        axis y line*=right,
        axis x line=none, % 不重复绘制X轴
        ylabel={Ratio ($\log_{10}$)},
        ymin=0, ymax=4, % 显著调整 Ymax
        width=0.9\linewidth,
        % --- 图例 ---
        legend style={
            at={(0.5, -0.2)},
            anchor=north,
            legend columns=3,
            legend cell align=left,
        }
    ]
        % 占位图
        \addplot[draw=none, forget plot] coordinates {(0,0)};
        
        % 图例条目
        \addlegendimage{softred, mark=*, sharp plot, thick}
        \addlegendentry{Effective}
        
        \addlegendimage{softblue, mark=square*, sharp plot, thick}
        \addlegendentry{Ineffective}
        
        % 更新 "Ratio" 数据
        \addplot[softgreen, mark=triangle*, sharp plot, thick] coordinates {
            (0, 0.06806225136000504)
            (5, 0.5556676952896547)
            (10, 1.2183558279392124)
            (15, 1.805487452125645)
            (20, 2.160296188465829)
            (25, 2.766313873286559)
            (30, 3.1940051239931946)
            (31, 3.316694698832896)
        };
        \addlegendentry{Ratio}

    \end{axis}
\end{tikzpicture}
    \caption{
        The average gradient flow (left y-axis) and the ratio between the effective and the ineffective (right y-axis) under each layer of Llama3.1-8b on ARC-Challenge.
    }
    \label{fig:grad_flow_llama31_8b_arc}
\end{figure}
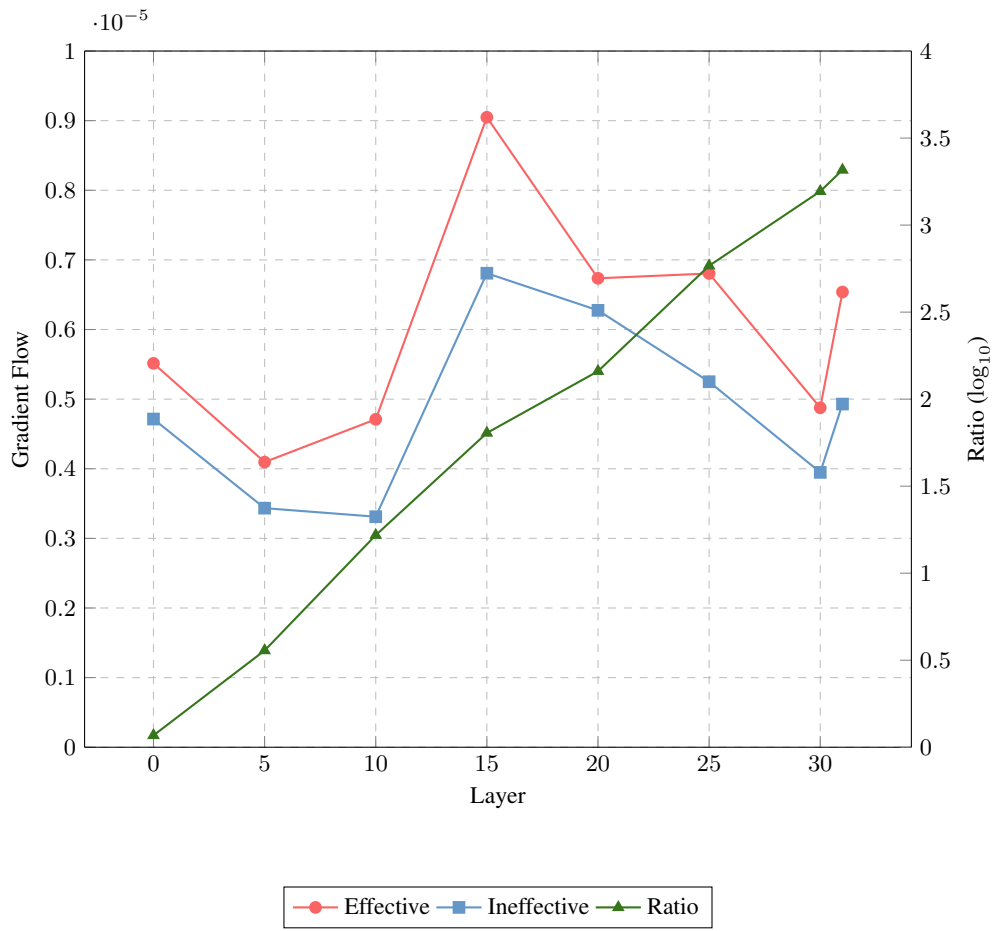

\begin{figure}
    \small
    \centering
    \begin{tikzpicture}
    \begin{axis}[
        % --- 左侧 Y 轴 ---
        axis y line*=left,
        xlabel={Layer},
        ylabel={Gradient Flow}, 
        ymin=0, ymax=3e-6, % 调整 Ymax
        width=0.9\linewidth,
        grid=major,
        grid style=dashed,
        yticklabel style={/pgf/number format/fixed, /pgf/number format/precision=1},
    ]
        % 更新 "Effective" 数据
        \addplot[softred, mark=*, sharp plot, thick] coordinates {
            (0, 8.622542010184261e-07)
            (5, 8.43206024811417e-07)
            (10, 9.686898346235853e-07)
            (15, 2.6777549720877002e-06)
            (20, 1.951642535597749e-06)
            (25, 1.872582275432013e-06)
            (30, 1.5790667928144805e-06)
            (31, 1.94032457064471e-06)
        };
        % 更新 "Ineffective" 数据
        \addplot[softblue, mark=square*, sharp plot, thick] coordinates {
            (0, 8.735933790804395e-07)
            (5, 8.342910327585606e-07)
            (10, 9.360352931642169e-07)
            (15, 2.629016484512192e-06)
            (20, 1.8816121096776328e-06)
            (25, 1.824609418221712e-06)
            (30, 1.5631092868950702e-06)
            (31, 1.8682200169095609e-06)
        };
    \end{axis}

    \begin{axis}[
        % --- 右侧 Y 轴 ---
        axis y line*=right,
        axis x line=none, % 不重复绘制X轴
        ylabel={Ratio ($\log_{10}$)},
        ymin=-0.05, ymax=0.25, % 显著调整 Ymax
        width=0.9\linewidth,
        % --- 图例 ---
        legend style={
            at={(0.5, -0.2)},
            anchor=north,
            legend columns=3,
            legend cell align=left,
        }
    ]
        % 占位图
        \addplot[draw=none, forget plot] coordinates {(0,0)};
        
        % 图例条目
        \addlegendimage{softred, mark=*, sharp plot, thick}
        \addlegendentry{Effective}
        
        \addlegendimage{softblue, mark=square*, sharp plot, thick}
        \addlegendentry{Ineffective}
        
        % 更新 "Ratio" 数据
        \addplot[softgreen, mark=triangle*, sharp plot, thick] coordinates {
            (0, -0.005674014891754471)
            (5, 0.04551175626225638)
            (10, 0.05266352461950259)
            (15, 0.05958451309277709)
            (20, 0.0854199369305946)
            (25, 0.10927250511227939)
            (30, 0.1876881062603129)
            (31, 0.20413446852952014)
        };
        \addlegendentry{Ratio}

    \end{axis}
\end{tikzpicture}
    \caption{
        The average gradient flow (left y-axis) and the ratio between the effective and the ineffective (right y-axis) under each layer of Llama3.1-8b on MMLU-Pro.
    }
    \label{fig:grad_flow_llama31_8b_mmlu}
\end{figure}
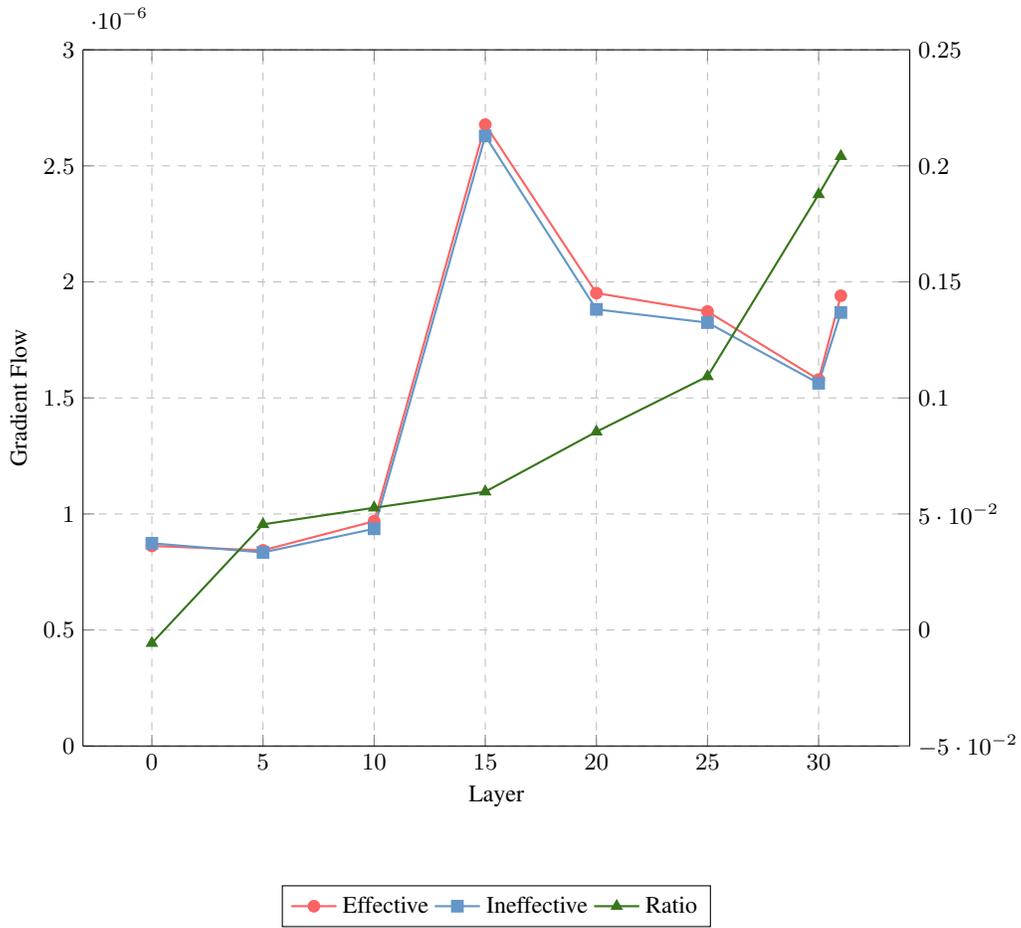

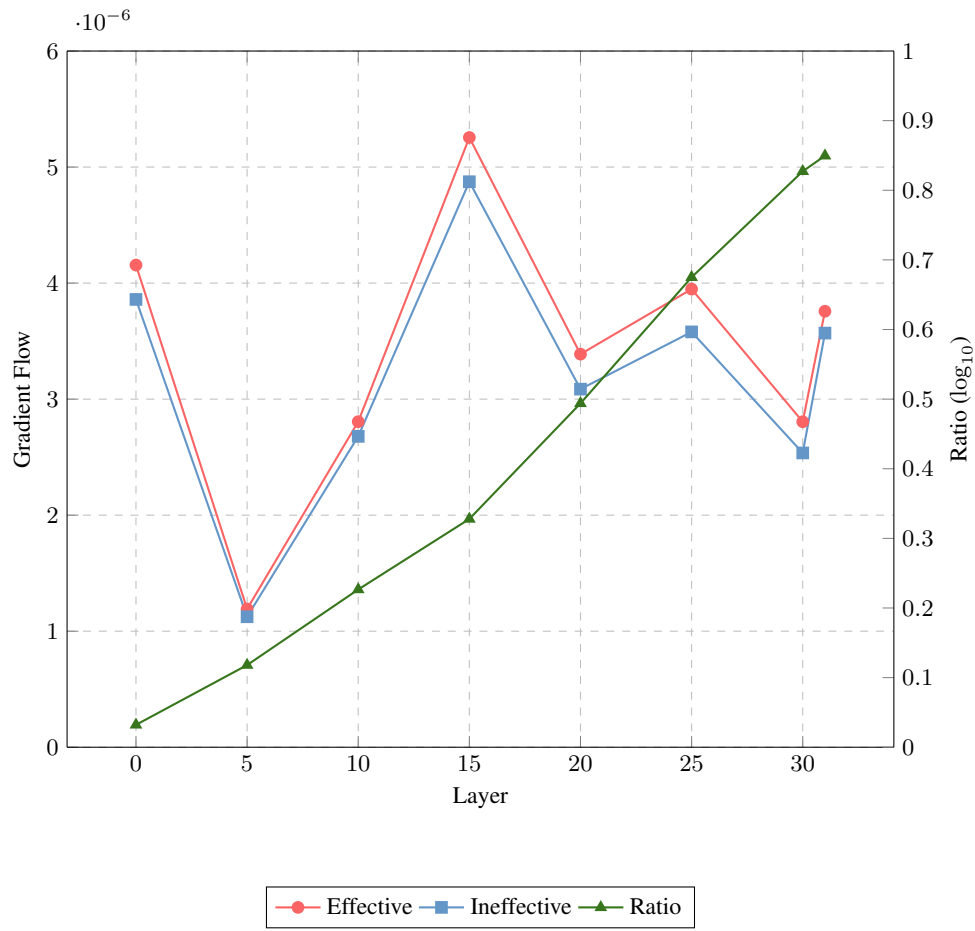
\begin{figure}
    \small
    \centering
    \begin{tikzpicture}
    \begin{axis}[
        % --- 左侧 Y 轴 ---
        axis y line*=left,
        xlabel={Layer},
        ylabel={Gradient Flow}, 
        ymin=0, ymax=6e-6, % 调整 Ymax
        width=0.9\linewidth,
        grid=major,
        grid style=dashed,
        yticklabel style={/pgf/number format/fixed, /pgf/number format/precision=1},
    ]
        % 更新 "Effective" 数据
        \addplot[softred, mark=*, sharp plot, thick] coordinates {
            (0, 4.155530488193248e-06)
            (5, 1.1916050633213224e-06)
            (10, 2.8059177387044306e-06)
            (15, 5.255003324796585e-06)
            (20, 3.3877415389200064e-06)
            (25, 3.948194423978398e-06)
            (30, 2.8053653999485014e-06)
            (31, 3.757498273688935e-06)
        };
        % 更新 "Ineffective" 数据
        \addplot[softblue, mark=square*, sharp plot, thick] coordinates {
            (0, 3.8590870945559035e-06)
            (5, 1.1244742079740242e-06)
            (10, 2.6791903173998877e-06)
            (15, 4.8735879270298615e-06)
            (20, 3.086180473958173e-06)
            (25, 3.579470643444438e-06)
            (30, 2.5362311132979693e-06)
            (31, 3.569037703197952e-06)
        };
    \end{axis}

    \begin{axis}[
        % --- 右侧 Y 轴 ---
        axis y line*=right,
        axis x line=none, % 不重复绘制X轴
        ylabel={Ratio ($\log_{10}$)},
        ymin=0, ymax=1, % 显著调整 Ymax
        width=0.9\linewidth,
        % --- 图例 ---
        legend style={
            at={(0.5, -0.2)},
            anchor=north,
            legend columns=3,
            legend cell align=left,
        }
    ]
        % 占位图
        \addplot[draw=none, forget plot] coordinates {(0,0)};
        
        % 图例条目
        \addlegendimage{softred, mark=*, sharp plot, thick}
        \addlegendentry{Effective}
        
        \addlegendimage{softblue, mark=square*, sharp plot, thick}
        \addlegendentry{Ineffective}
        
        % 更新 "Ratio" 数据
        \addplot[softgreen, mark=triangle*, sharp plot, thick] coordinates {
            (0, 0.032141892837793104)
            (5, 0.11798892007224213)
            (10, 0.22655808688647985)
            (15, 0.3278458025114596)
            (20, 0.493881817674351)
            (25, 0.6750205229197838)
            (30, 0.8272038296610562)
            (31, 0.8495514835390708)
        };
        \addlegendentry{Ratio}

    \end{axis}
\end{tikzpicture}
    \caption{
        The average gradient flow (left y-axis) and the ratio between the effective and the ineffective (right y-axis) under each layer of Llama3.1-8b on Amazon Review.
    }
    \label{fig:grad_flow_llama31_8b_amazon}
\end{figure}

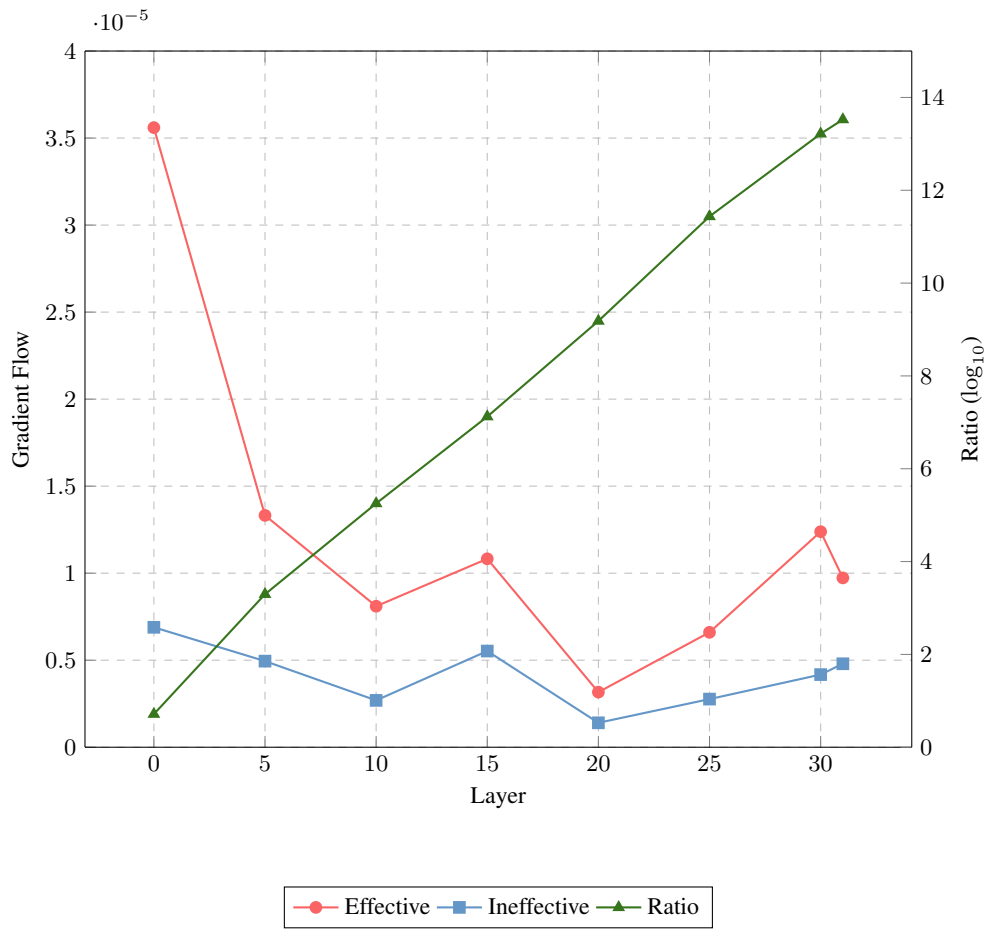
\begin{figure}
    \small
    \centering
    \begin{tikzpicture}
    \begin{axis}[
        % --- 左侧 Y 轴 ---
        axis y line*=left,
        xlabel={Layer},
        ylabel={Gradient Flow}, 
        ymin=0, ymax=4e-5, % 调整 Ymax
        width=0.9\linewidth,
        grid=major,
        grid style=dashed,
        yticklabel style={/pgf/number format/fixed, /pgf/number format/precision=1},
    ]
        % 更新 "Effective" 数据
        \addplot[softred, mark=*, sharp plot, thick] coordinates {
            (0, 3.559993638191372e-05)
            (5, 1.3319908248377033e-05)
            (10, 8.099747901724186e-06)
            (15, 1.0825474419107195e-05)
            (20, 3.1652778034185758e-06)
            (25, 6.60180876366212e-06)
            (30, 1.2385087757138535e-05)
            (31, 9.728507393447217e-06)
        };
        % 更新 "Ineffective" 数据
        \addplot[softblue, mark=square*, sharp plot, thick] coordinates {
            (0, 6.890828362057536e-06)
            (5, 4.946653291426628e-06)
            (10, 2.691352888591196e-06)
            (15, 5.525402994780134e-06)
            (20, 1.4034774577257023e-06)
            (25, 2.764703708635352e-06)
            (30, 4.173642744056203e-06)
            (31, 4.7955660523836165e-06)
        };
    \end{axis}

    \begin{axis}[
        % --- 右侧 Y 轴 ---
        axis y line*=right,
        axis x line=none, % 不重复绘制X轴
        ylabel={Ratio ($\log_{10}$)},
        ymin=0, ymax=15, % 显著调整 Ymax
        width=0.9\linewidth,
        % --- 图例 ---
        legend style={
            at={(0.5, -0.2)},
            anchor=north,
            legend columns=3,
            legend cell align=left,
        }
    ]
        % 占位图
        \addplot[draw=none, forget plot] coordinates {(0,0)};
        
        % 图例条目
        \addlegendimage{softred, mark=*, sharp plot, thick}
        \addlegendentry{Effective}
        
        \addlegendimage{softblue, mark=square*, sharp plot, thick}
        \addlegendentry{Ineffective}
        
        % 更新 "Ratio" 数据
        \addplot[softgreen, mark=triangle*, sharp plot, thick] coordinates {
            (0, 0.7131777893096537)
            (5, 3.295874768246594)
            (10, 5.251710220136274)
            (15, 7.12282099331919)
            (20, 9.183073255169125)
            (25, 11.435724405141231)
            (30, 13.21659317262774)
            (31, 13.523799508733267)
        };
        \addlegendentry{Ratio}

    \end{axis}
\end{tikzpicture}
    \caption{
        The average gradient flow (left y-axis) and the ratio between the effective and the ineffective (right y-axis) under each layer of Llama-R1-8b on GSM8K.
    }
    \label{fig:grad_flow_llama_r1_8b_gsm8k}
\end{figure}

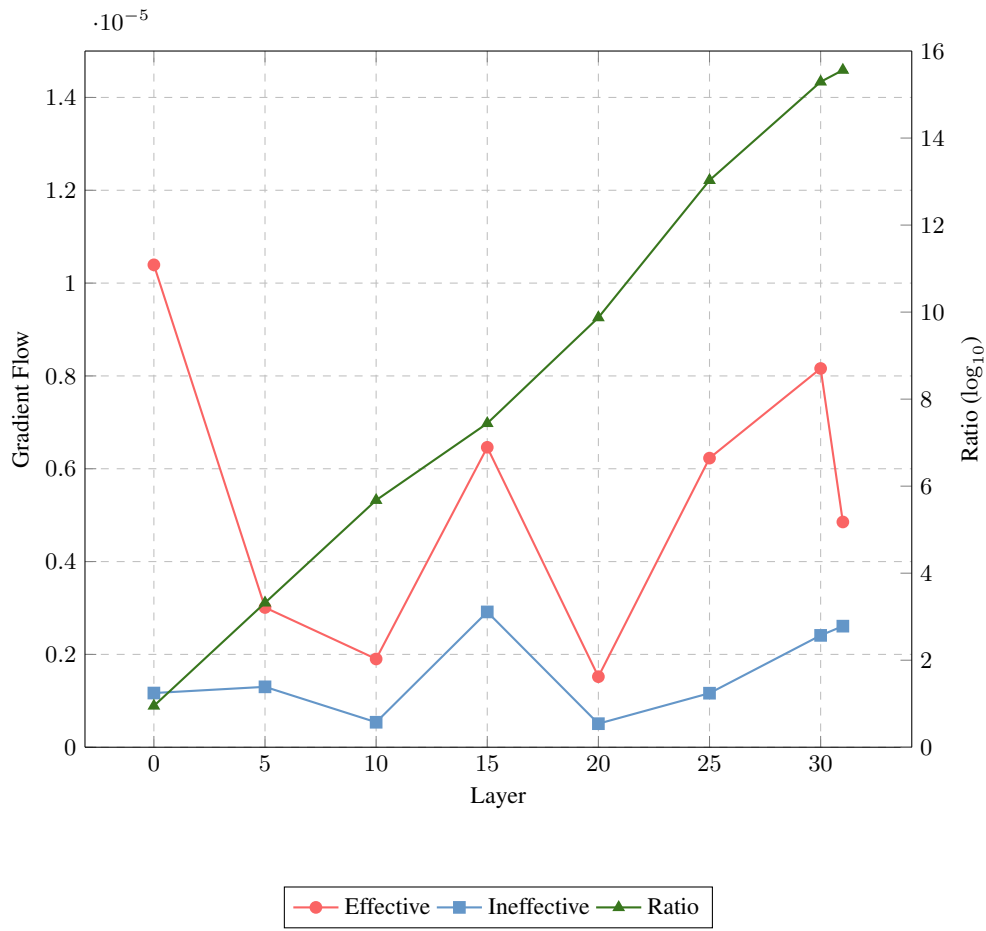
\begin{figure}
    \small
    \centering
    \begin{tikzpicture}
    \begin{axis}[
        % --- 左侧 Y 轴 ---
        axis y line*=left,
        xlabel={Layer},
        ylabel={Gradient Flow}, 
        ymin=0, ymax=1.5e-5, % 调整 Ymax
        width=0.9\linewidth,
        grid=major,
        grid style=dashed,
        yticklabel style={/pgf/number format/fixed, /pgf/number format/precision=1},
    ]
        % 更新 "Effective" 数据
        \addplot[softred, mark=*, sharp plot, thick] coordinates {
            (0, 1.0393529336738538e-05)
            (5, 3.0105030646154773e-06)
            (10, 1.9018725652131252e-06)
            (15, 6.461676861135857e-06)
            (20, 1.5200527201386649e-06)
            (25, 6.228986322298624e-06)
            (30, 8.160280685842736e-06)
            (31, 4.8531768754855875e-06)
        };
        % 更新 "Ineffective" 数据
        \addplot[softblue, mark=square*, sharp plot, thick] coordinates {
            (0, 1.1676701205942663e-06)
            (5, 1.3012271438128664e-06)
            (10, 5.381868390941236e-07)
            (15, 2.915219283750048e-06)
            (20, 5.073757733953244e-07)
            (25, 1.1642271147138672e-06)
            (30, 2.4096862034639344e-06)
            (31, 2.6086227080668323e-06)
        };
    \end{axis}

    \begin{axis}[
        % --- 右侧 Y 轴 ---
        axis y line*=right,
        axis x line=none, % 不重复绘制X轴
        ylabel={Ratio ($\log_{10}$)},
        ymin=0, ymax=16, % 显著调整 Ymax
        width=0.9\linewidth,
        % --- 图例 ---
        legend style={
            at={(0.5, -0.2)},
            anchor=north,
            legend columns=3,
            legend cell align=left,
        }
    ]
        % 占位图
        \addplot[draw=none, forget plot] coordinates {(0,0)};
        
        % 图例条目
        \addlegendimage{softred, mark=*, sharp plot, thick}
        \addlegendentry{Effective}
        
        \addlegendimage{softblue, mark=square*, sharp plot, thick}
        \addlegendentry{Ineffective}
        
        % 更新 "Ratio" 数据
        \addplot[softgreen, mark=triangle*, sharp plot, thick] coordinates {
            (0, 0.949442879000963)
            (5, 3.3188410658198806)
            (10, 5.676697285545097)
            (15, 7.443586144467138)
            (20, 9.874626321276821)
            (25, 13.029105633673678)
            (30, 15.292316347822798)
            (31, 15.561931196925013)
        };
        \addlegendentry{Ratio}

    \end{axis}
\end{tikzpicture}
    \caption{
        The average gradient flow (left y-axis) and the ratio between the effective and the ineffective (right y-axis) under each layer of Llama-R1-8b on MATH.
    }
    \label{fig:grad_flow_llama_r1_8b_math}
\end{figure}

\begin{figure}
    \small
    \centering
    \begin{tikzpicture}
    \begin{axis}[
        % --- 左侧 Y 轴 ---
        axis y line*=left,
        xlabel={Layer},
        ylabel={Gradient Flow}, 
        ymin=0, ymax=1e-4, % 调整 Ymax
        width=0.9\linewidth,
        grid=major,
        grid style=dashed,
        yticklabel style={/pgf/number format/fixed, /pgf/number format/precision=1},
    ]
        % 更新 "Effective" 数据
        \addplot[softred, mark=*, sharp plot, thick] coordinates {
            (0, 7.427738065382578e-05)
            (5, 9.737345874770666e-05)
            (10, 2.0482002357892257e-05)
            (15, 1.6381442455132795e-05)
            (20, 3.6071368757625158e-06)
            (25, 8.297749388699546e-06)
            (30, 3.8695667484858714e-05)
            (31, 1.3594102665666623e-05)
        };
        % 更新 "Ineffective" 数据
        \addplot[softblue, mark=square*, sharp plot, thick] coordinates {
            (0, 6.971304069102425e-05)
            (5, 9.213169202033588e-05)
            (10, 2.032785147475502e-05)
            (15, 1.6818193536906852e-05)
            (20, 3.6067861888255948e-06)
            (25, 7.920144081028459e-06)
            (30, 3.593053172768821e-05)
            (31, 1.4386049543848612e-05)
        };
    \end{axis}

    \begin{axis}[
        % --- 右侧 Y 轴 ---
        axis y line*=right,
        axis x line=none, % 不重复绘制X轴
        ylabel={Ratio ($\log_{10}$)},
        ymin=0, ymax=0.6, % 显著调整 Ymax
        width=0.9\linewidth,
        % --- 图例 ---
        legend style={
            at={(0.5, -0.2)},
            anchor=north,
            legend columns=3,
            legend cell align=left,
        }
    ]
        % 占位图
        \addplot[draw=none, forget plot] coordinates {(0,0)};
        
        % 图例条目
        \addlegendimage{softred, mark=*, sharp plot, thick}
        \addlegendentry{Effective}
        
        \addlegendimage{softblue, mark=square*, sharp plot, thick}
        \addlegendentry{Ineffective}
        
        % 更新 "Ratio" 数据
        \addplot[softgreen, mark=triangle*, sharp plot, thick] coordinates {
            (0, 0.006990601516640581)
            (5, 0.11139125751362529)
            (10, 0.12272977323325115)
            (15, 0.17837783387551182)
            (20, 0.3033949636794775)
            (25, 0.38301391549851416)
            (30, 0.49738804509631057)
            (31, 0.5121421866551232)
        };
        \addlegendentry{Ratio}

    \end{axis}
\end{tikzpicture}
    \caption{
        The average gradient flow (left y-axis) and the ratio between the effective and the ineffective (right y-axis) under each layer of Llama-R1-8b on ARC-Challenge.
    }
    \label{fig:grad_flow_llama_r1_8b_arc}
\end{figure}
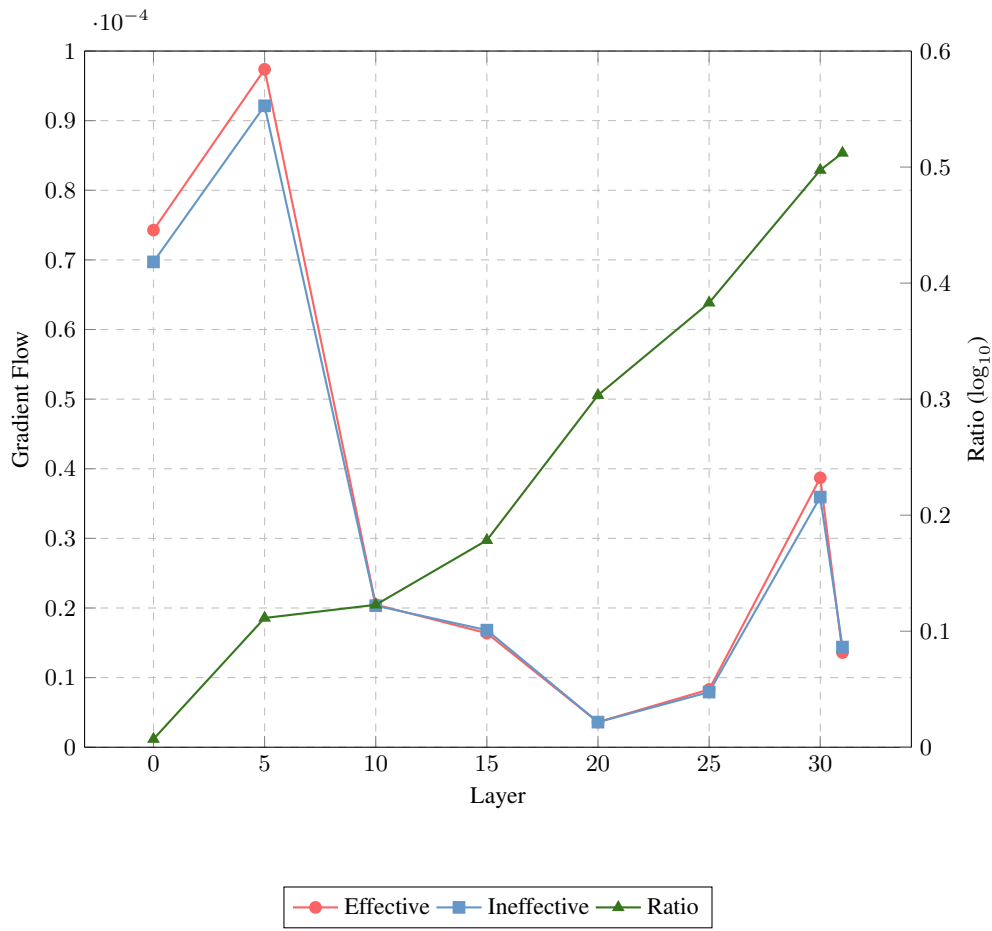

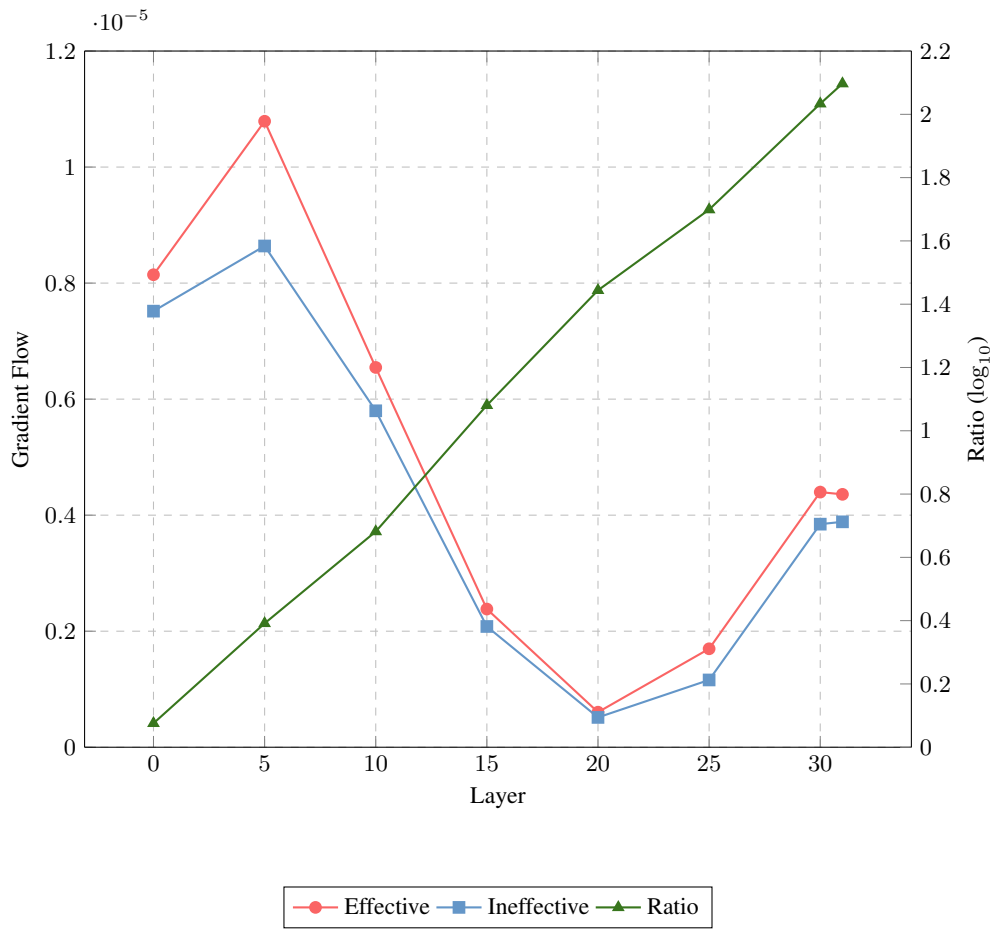
\begin{figure}
    \small
    \centering
    \begin{tikzpicture}
    \begin{axis}[
        % --- 左侧 Y 轴 ---
        axis y line*=left,
        xlabel={Layer},
        ylabel={Gradient Flow}, 
        ymin=0, ymax=1.2e-5, % 调整 Ymax
        width=0.9\linewidth,
        grid=major,
        grid style=dashed,
        yticklabel style={/pgf/number format/fixed, /pgf/number format/precision=1},
    ]
        % 更新 "Effective" 数据
        \addplot[softred, mark=*, sharp plot, thick] coordinates {
            (0, 8.144178810936956e-06)
            (5, 1.078973177696204e-05)
            (10, 6.5461398879216255e-06)
            (15, 2.3817613578212507e-06)
            (20, 6.042468215769125e-07)
            (25, 1.6967908909561784e-06)
            (30, 4.397413586127313e-06)
            (31, 4.358927692236738e-06)
        };
        % 更新 "Ineffective" 数据
        \addplot[softblue, mark=square*, sharp plot, thick] coordinates {
            (0, 7.5166148212486555e-06)
            (5, 8.639674439600542e-06)
            (10, 5.799091222372733e-06)
            (15, 2.0812147193309135e-06)
            (20, 5.138124100850662e-07)
            (25, 1.1595492906874398e-06)
            (30, 3.845029752096807e-06)
            (31, 3.88524537656296e-06)
        };
    \end{axis}

    \begin{axis}[
        % --- 右侧 Y 轴 ---
        axis y line*=right,
        axis x line=none, % 不重复绘制X轴
        ylabel={Ratio ($\log_{10}$)},
        ymin=0, ymax=2.2, % 显著调整 Ymax
        width=0.9\linewidth,
        % --- 图例 ---
        legend style={
            at={(0.5, -0.2)},
            anchor=north,
            legend columns=3,
            legend cell align=left,
        }
    ]
        % 占位图
        \addplot[draw=none, forget plot] coordinates {(0,0)};
        
        % 图例条目
        \addlegendimage{softred, mark=*, sharp plot, thick}
        \addlegendentry{Effective}
        
        \addlegendimage{softblue, mark=square*, sharp plot, thick}
        \addlegendentry{Ineffective}
        
        % 更新 "Ratio" 数据
        \addplot[softgreen, mark=triangle*, sharp plot, thick] coordinates {
            (0, 0.07559744025068595)
            (5, 0.391355852798011)
            (10, 0.6816184726840433)
            (15, 1.0804862346325872)
            (20, 1.4439223551789495)
            (25, 1.698938675409177)
            (30, 2.0335079036839607)
            (31, 2.096985713312038)
        };
        \addlegendentry{Ratio}

    \end{axis}
\end{tikzpicture}
    \caption{
        The average gradient flow (left y-axis) and the ratio between the effective and the ineffective (right y-axis) under each layer of Llama-R1-8b on MMLU-Pro.
    }
    \label{fig:grad_flow_llama_r1_8b_mmlu}
\end{figure}

\begin{figure}
    \small
    \centering
    \begin{tikzpicture}
    \begin{axis}[
        % --- 左侧 Y 轴 ---
        axis y line*=left,
        xlabel={Layer},
        ylabel={Gradient Flow}, 
        ymin=0, ymax=3e-5, % 调整 Ymax
        width=0.9\linewidth,
        grid=major,
        grid style=dashed,
        yticklabel style={/pgf/number format/fixed, /pgf/number format/precision=1},
    ]
        % 更新 "Effective" 数据
        \addplot[softred, mark=*, sharp plot, thick] coordinates {
            (0, 2.5715761391135555e-05)
            (5, 5.024336511875542e-06)
            (10, 1.1002157407347113e-05)
            (15, 1.0223326777728895e-05)
            (20, 2.674984784789558e-06)
            (25, 7.4532843730897485e-06)
            (30, 6.4920912498356e-06)
            (31, 6.4989611322137835e-06)
        };
        % 更新 "Ineffective" 数据
        \addplot[softblue, mark=square*, sharp plot, thick] coordinates {
            (0, 2.2542473910410384e-05)
            (5, 4.8979116096840395e-06)
            (10, 7.365406213466486e-06)
            (15, 7.5614245740022545e-06)
            (20, 2.3003995333207666e-06)
            (25, 5.464271574358767e-06)
            (30, 6.094948585592889e-06)
            (31, 5.94315174945829e-06)
        };
    \end{axis}

    \begin{axis}[
        % --- 右侧 Y 轴 ---
        axis y line*=right,
        axis x line=none, % 不重复绘制X轴
        ylabel={Ratio ($\log_{10}$)},
        ymin=0, ymax=3, % 显著调整 Ymax
        width=0.9\linewidth,
        % --- 图例 ---
        legend style={
            at={(0.5, -0.2)},
            anchor=north,
            legend columns=3,
            legend cell align=left,
        }
    ]
        % 占位图
        \addplot[draw=none, forget plot] coordinates {(0,0)};
        
        % 图例条目
        \addlegendimage{softred, mark=*, sharp plot, thick}
        \addlegendentry{Effective}
        
        \addlegendimage{softblue, mark=square*, sharp plot, thick}
        \addlegendentry{Ineffective}
        
        % 更新 "Ratio" 数据
        \addplot[softgreen, mark=triangle*, sharp plot, thick] coordinates {
            (0, 0.05719781170391028)
            (5, 0.4102491491557467)
            (10, 0.9982036509866486)
            (15, 1.6222975970002749)
            (20, 2.0223326976409646)
            (25, 2.3315250698639804)
            (30, 2.571829190041956)
            (31, 2.6106563102111497)
        };
        \addlegendentry{Ratio}

    \end{axis}
\end{tikzpicture}
    \caption{
        The average gradient flow (left y-axis) and the ratio between the effective and the ineffective (right y-axis) under each layer of Llama-R1-8b on Amazon Review.
    }
    \label{fig:grad_flow_llama_r1_8b_amazon}
\end{figure}
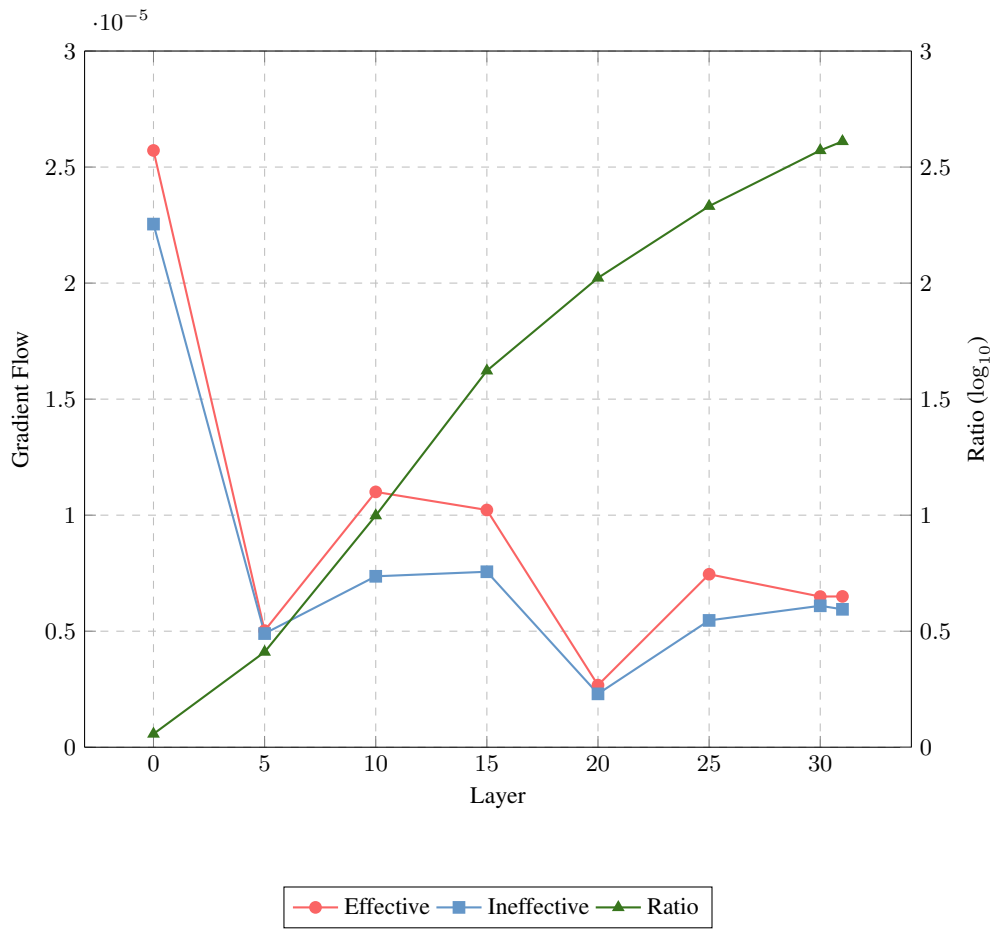

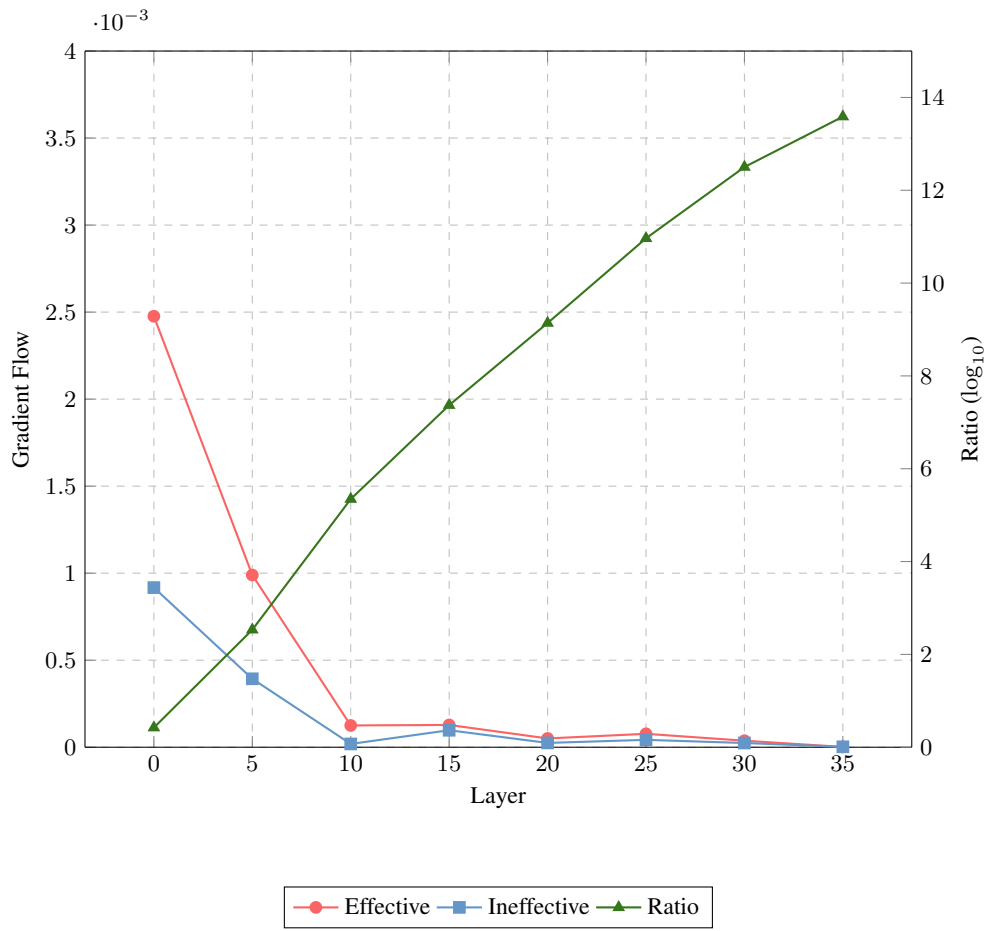
\begin{figure}
    \small
    \centering
    \begin{tikzpicture}
    \begin{axis}[
        % --- 左侧 Y 轴 ---
        axis y line*=left,
        xlabel={Layer},
        ylabel={Gradient Flow}, 
        ymin=0, ymax=4e-3, % 调整 Ymax
        width=0.9\linewidth,
        grid=major,
        grid style=dashed,
        yticklabel style={/pgf/number format/fixed, /pgf/number format/precision=1},
    ]
        % 更新 "Effective" 数据
        \addplot[softred, mark=*, sharp plot, thick] coordinates {
            (0, 0.00247630238401893)
            (5, 0.0009891603581776642)
            (10, 0.00012510979536407653)
            (15, 0.0001279727905790574)
            (20, 5.046153055338215e-05)
            (25, 7.704257838385447e-05)
            (30, 3.745685418029884e-05)
            (35, 1.7538671730375785e-06)
        };
        % 更新 "Ineffective" 数据
        \addplot[softblue, mark=square*, sharp plot, thick] coordinates {
            (0, 0.0009171846825655966)
            (5, 0.00039322260389232486)
            (10, 1.8862134355834353e-05)
            (15, 9.756627760035154e-05)
            (20, 2.4747284536043363e-05)
            (25, 4.237393257564849e-05)
            (30, 2.423390071788103e-05)
            (35, 2.4526773482485015e-06)
        };
    \end{axis}

    \begin{axis}[
        % --- 右侧 Y 轴 ---
        axis y line*=right,
        axis x line=none, % 不重复绘制X轴
        ylabel={Ratio ($\log_{10}$)},
        ymin=0, ymax=15, % 显著调整 Ymax
        width=0.9\linewidth,
        % --- 图例 ---
        legend style={
            at={(0.5, -0.2)},
            anchor=north,
            legend columns=3,
            legend cell align=left,
        }
    ]
        % 占位图
        \addplot[draw=none, forget plot] coordinates {(0,0)};
        
        % 图例条目
        \addlegendimage{softred, mark=*, sharp plot, thick}
        \addlegendentry{Effective}
        
        \addlegendimage{softblue, mark=square*, sharp plot, thick}
        \addlegendentry{Ineffective}
        
        % 更新 "Ratio" 数据
        \addplot[softgreen, mark=triangle*, sharp plot, thick] coordinates {
            (0, 0.42212091458735246)
            (5, 2.5308391117617943)
            (10, 5.345791213108586)
            (15, 7.369274465276719)
            (20, 9.138776598804906)
            (25, 10.96458140657208)
            (30, 12.499992681822178)
            (35, 13.586035867790612)
        };
        \addlegendentry{Ratio}

    \end{axis}
\end{tikzpicture}
    \caption{
        The average gradient flow (left y-axis) and the ratio between the effective and the ineffective (right y-axis) under each layer of Qwen3-8b on GSM8K.
    }
    \label{fig:grad_flow_qwen3_8b_gsm8k}
\end{figure}

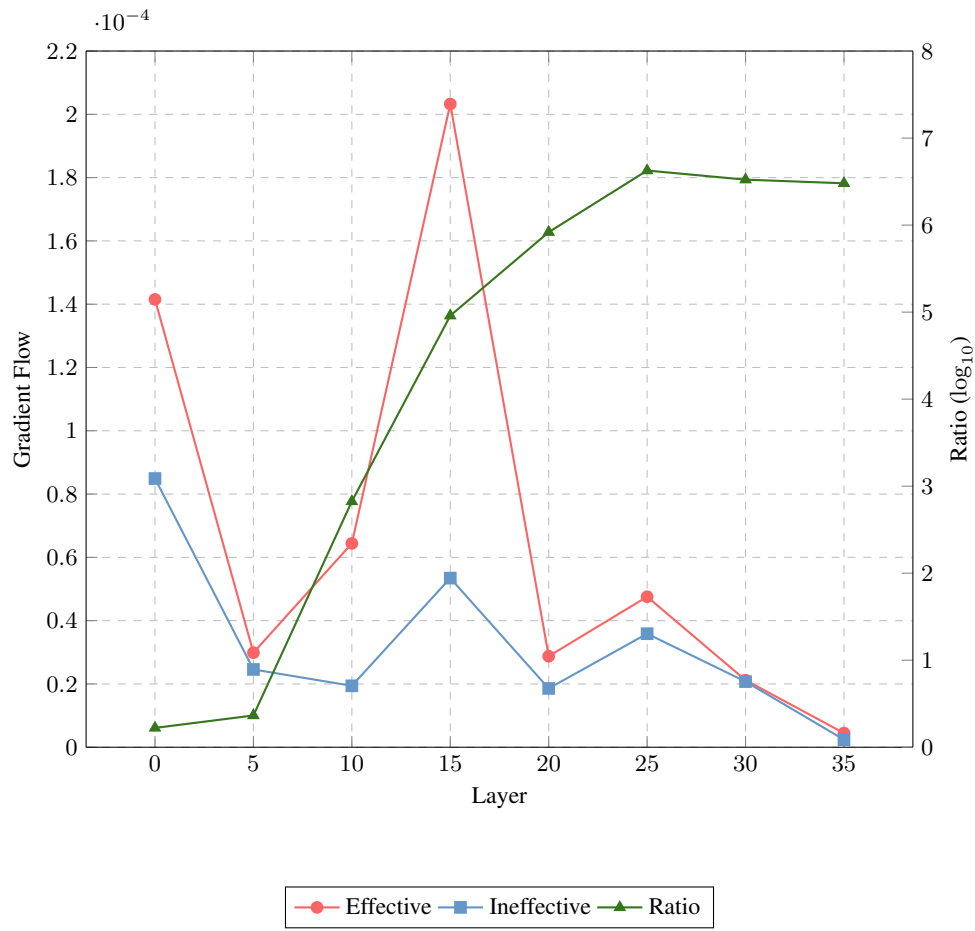
\begin{figure}
    \small
    \centering
    \begin{tikzpicture}
    \begin{axis}[
        % --- 左侧 Y 轴 ---
        axis y line*=left,
        xlabel={Layer},
        ylabel={Gradient Flow}, 
        ymin=0, ymax=2.2e-4, % 调整 Ymax
        width=0.9\linewidth,
        grid=major,
        grid style=dashed,
        yticklabel style={/pgf/number format/fixed, /pgf/number format/precision=1},
    ]
        % 更新 "Effective" 数据
        \addplot[softred, mark=*, sharp plot, thick] coordinates {
            (0, 0.0001415163278579712)
            (5, 2.987310290336609e-05)
            (10, 6.441306322813034e-05)
            (15, 0.0002032555639743805)
            (20, 2.876296639442444e-05)
            (25, 4.7557055950164795e-05)
            (30, 2.1189451217651367e-05)
            (35, 4.4668559004857705e-06)
        };
        % 更新 "Ineffective" 数据
        \addplot[softblue, mark=square*, sharp plot, thick] coordinates {
            (0, 8.490838502582751e-05)
            (5, 2.4571524638878672e-05)
            (10, 1.941013493035969e-05)
            (15, 5.3441421569962253e-05)
            (20, 1.857895404100418e-05)
            (25, 3.587454557418823e-05)
            (30, 2.07439849251195e-05)
            (35, 2.4000892871730635e-06)
        };
    \end{axis}

    \begin{axis}[
        % --- 右侧 Y 轴 ---
        axis y line*=right,
        axis x line=none, % 不重复绘制X轴
        ylabel={Ratio ($\log_{10}$)},
        ymin=0, ymax=8, % 显著调整 Ymax
        width=0.9\linewidth,
        % --- 图例 ---
        legend style={
            at={(0.5, -0.2)},
            anchor=north,
            legend columns=3,
            legend cell align=left,
        }
    ]
        % 占位图
        \addplot[draw=none, forget plot] coordinates {(0,0)};
        
        % 图例条目
        \addlegendimage{softred, mark=*, sharp plot, thick}
        \addlegendentry{Effective}
        
        \addlegendimage{softblue, mark=square*, sharp plot, thick}
        \addlegendentry{Ineffective}
        
        % 更新 "Ratio" 数据
        \addplot[softgreen, mark=triangle*, sharp plot, thick] coordinates {
            (0, 0.22185597014736355)
            (5, 0.3650113960124942)
            (10, 2.824964517967911)
            (15, 4.959336259533357)
            (20, 5.918810034902457)
            (25, 6.627009934881028)
            (30, 6.522198796905296)
            (35, 6.4804927609996135)
        };
        \addlegendentry{Ratio}

    \end{axis}
\end{tikzpicture}
    \caption{
        The average gradient flow (left y-axis) and the ratio between the effective and the ineffective (right y-axis) under each layer of Qwen3-8b on MATH.
    }
    \label{fig:grad_flow_qwen3_8b_math}
\end{figure}

\begin{figure}
    \small
    \centering
    \begin{tikzpicture}
    \begin{axis}[
        % --- 左侧 Y 轴 ---
        axis y line*=left,
        xlabel={Layer},
        ylabel={Gradient Flow}, 
        ymin=0, ymax=6e-4, % 调整 Ymax
        width=0.9\linewidth,
        grid=major,
        grid style=dashed,
        yticklabel style={/pgf/number format/fixed, /pgf/number format/precision=1},
    ]
        % 更新 "Effective" 数据
        \addplot[softred, mark=*, sharp plot, thick] coordinates {
            (0, 0.0005901963112716079)
            (5, 0.0001767560438870739)
            (10, 7.1088618753944e-06)
            (15, 1.5492140160126716e-05)
            (20, 7.377311542237667e-06)
            (25, 1.2435501777345466e-05)
            (30, 1.3638686036607985e-05)
            (35, 3.2319160219153774e-06)
        };
        % 更新 "Ineffective" 数据
        \addplot[softblue, mark=square*, sharp plot, thick] coordinates {
            (0, 0.00021094213488887753)
            (5, 6.504537589892476e-05)
            (10, 5.327973830190505e-06)
            (15, 2.2023067775021688e-05)
            (20, 5.436964914348355e-06)
            (25, 7.121356262018708e-06)
            (30, 7.522885153678414e-06)
            (35, 3.0800063637315134e-06)
        };
    \end{axis}

    \begin{axis}[
        % --- 右侧 Y 轴 ---
        axis y line*=right,
        axis x line=none, % 不重复绘制X轴
        ylabel={Ratio ($\log_{10}$)},
        ymin=0, ymax=7, % 显著调整 Ymax
        width=0.9\linewidth,
        % --- 图例 ---
        legend style={
            at={(0.5, -0.2)},
            anchor=north,
            legend columns=3,
            legend cell align=left,
        }
    ]
        % 占位图
        \addplot[draw=none, forget plot] coordinates {(0,0)};
        
        % 图例条目
        \addlegendimage{softred, mark=*, sharp plot, thick}
        \addlegendentry{Effective}
        
        \addlegendimage{softblue, mark=square*, sharp plot, thick}
        \addlegendentry{Ineffective}
        
        % 更新 "Ratio" 数据
        \addplot[softgreen, mark=triangle*, sharp plot, thick] coordinates {
            (0, 0.4452952482699798)
            (5, 2.5744096680291593)
            (10, 3.8896756327409303)
            (15, 4.385342903191488)
            (20, 5.1573986184414835)
            (25, 5.6487779463464065)
            (30, 5.864999687152827)
            (35, 6.145501435782862)
        };
        \addlegendentry{Ratio}

    \end{axis}
\end{tikzpicture}
    \caption{
        The average gradient flow (left y-axis) and the ratio between the effective and the ineffective (right y-axis) under each layer of Qwen3-8b on ARC-Challenge.
    }
    \label{fig:grad_flow_qwen3_8b_arc}
\end{figure}
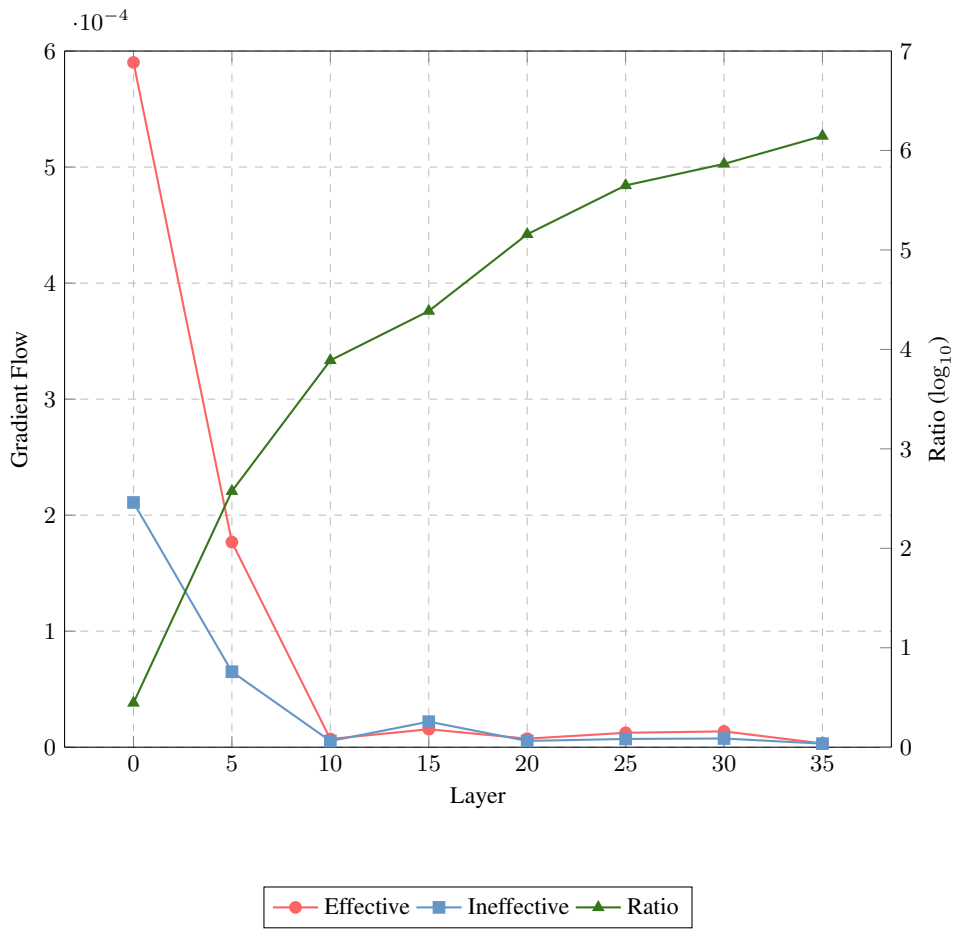

\begin{figure}
    \small
    \centering
    \begin{tikzpicture}
    \begin{axis}[
        % --- 左侧 Y 轴 ---
        axis y line*=left,
        xlabel={Layer},
        ylabel={Gradient Flow}, 
        ymin=0, ymax=1e-4, % 调整 Ymax
        width=0.9\linewidth,
        grid=major,
        grid style=dashed,
        yticklabel style={/pgf/number format/fixed, /pgf/number format/precision=1},
    ]
        % 更新 "Effective" 数据
        \addplot[softred, mark=*, sharp plot, thick] coordinates {
            (0, 8.716286055026992e-05)
            (5, 2.460567184468929e-05)
            (10, 6.6096917744274234e-06)
            (15, 2.1160622419532202e-05)
            (20, 5.924348888869363e-06)
            (25, 9.412022311147498e-06)
            (30, 9.69682972522732e-06)
            (35, 7.811669041245893e-07)
        };
        % 更新 "Ineffective" 数据
        \addplot[softblue, mark=square*, sharp plot, thick] coordinates {
            (0, 6.652196909619684e-05)
            (5, 1.6077541394287117e-05)
            (10, 6.458594324233674e-06)
            (15, 1.3701488497036109e-05)
            (20, 4.02882487842875e-06)
            (25, 8.13451932617942e-06)
            (30, 9.603965284519394e-06)
            (35, 6.550436120641106e-07)
        };
    \end{axis}

    \begin{axis}[
        % --- 右侧 Y 轴 ---
        axis y line*=right,
        axis x line=none, % 不重复绘制X轴
        ylabel={Ratio ($\log_{10}$)},
        ymin=0, ymax=2, % 显著调整 Ymax
        width=0.9\linewidth,
        % --- 图例 ---
        legend style={
            at={(0.5, -0.2)},
            anchor=north,
            legend columns=3,
            legend cell align=left,
        }
    ]
        % 占位图
        \addplot[draw=none, forget plot] coordinates {(0,0)};
        
        % 图例条目
        \addlegendimage{softred, mark=*, sharp plot, thick}
        \addlegendentry{Effective}
        
        \addlegendimage{softblue, mark=square*, sharp plot, thick}
        \addlegendentry{Ineffective}
        
        % 更新 "Ratio" 数据
        \addplot[softgreen, mark=triangle*, sharp plot, thick] coordinates {
            (0, 0.09428224693973282)
            (5, 0.12880328871176328)
            (10, 0.3750752103518097)
            (15, 0.8397080191767384)
            (20, 0.882746012275665)
            (25, 0.8925407835487162)
            (30, 1.057101842405872)
            (31, 1.5787971032449564)
        };
        \addlegendentry{Ratio}

    \end{axis}
\end{tikzpicture}
    \caption{
        The average gradient flow (left y-axis) and the ratio between the effective and the ineffective (right y-axis) under each layer of Qwen3-8b on MMLU-Pro.
    }
    \label{fig:grad_flow_qwen3_8b_mmlu}
\end{figure}
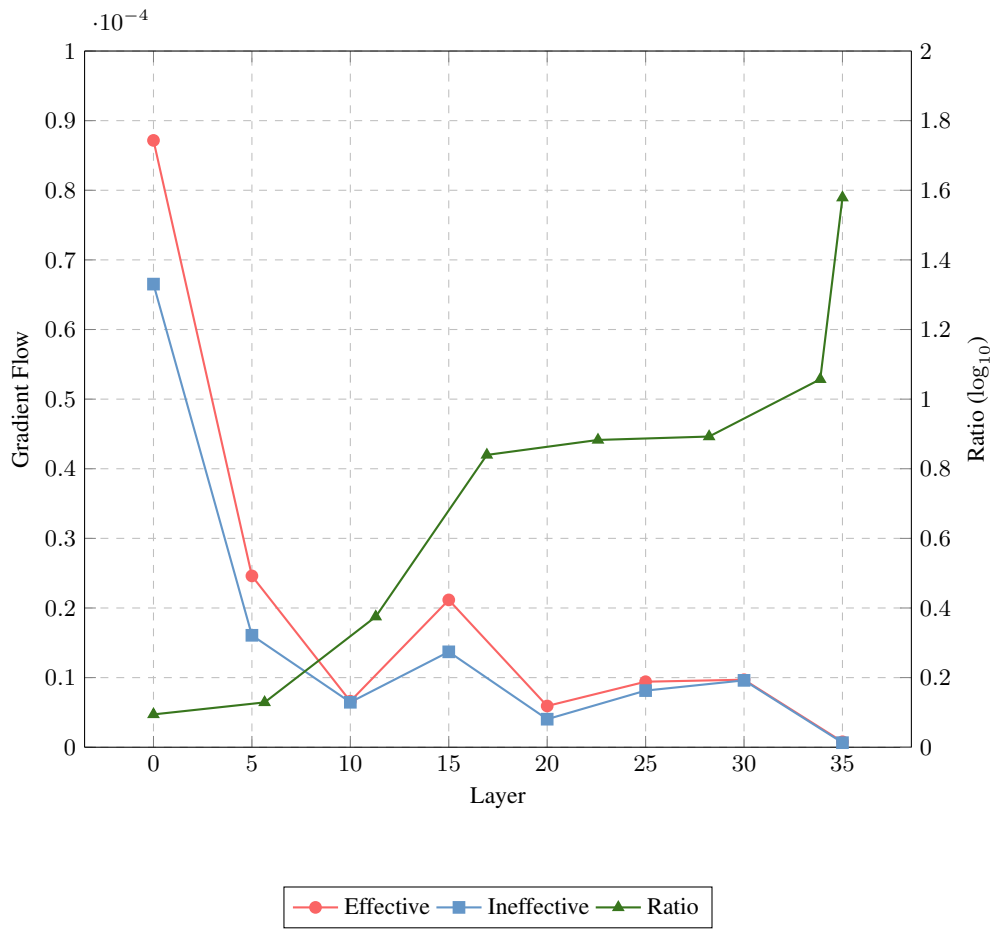

\begin{figure}
    \small
    \centering
    \begin{tikzpicture}
    \begin{axis}[
        % --- 左侧 Y 轴 ---
        axis y line*=left,
        xlabel={Layer},
        ylabel={Gradient Flow}, 
        ymin=0, ymax=2e-3, % 调整 Ymax
        width=0.9\linewidth,
        grid=major,
        grid style=dashed,
        yticklabel style={/pgf/number format/fixed, /pgf/number format/precision=1},
    ]
        % 更新 "Effective" 数据
        \addplot[softred, mark=*, sharp plot, thick] coordinates {
            (0, 0.0019677791732198324)
            (5, 0.000527436042968649)
            (10, 9.779280814401116e-06)
            (15, 1.7270328756170882e-05)
            (20, 1.4625651727847448e-05)
            (25, 0.0001279744736196266)
            (30, 8.49318037103921e-05)
            (35, 7.984481560198732e-06)
        };
        % 更新 "Ineffective" 数据
        \addplot[softblue, mark=square*, sharp plot, thick] coordinates {
            (0, 0.0006089733925047925)
            (5, 0.00016586492354828965)
            (10, 6.137218249698769e-06)
            (15, 7.829924354721046e-06)
            (20, 1.196829939748736e-05)
            (25, 3.2535444198448755e-05)
            (30, 3.8911870255422244e-05)
            (35, 3.505593628562031e-06)
        };
    \end{axis}

    \begin{axis}[
        % --- 右侧 Y 轴 ---
        axis y line*=right,
        axis x line=none, % 不重复绘制X轴
        ylabel={Ratio ($\log_{10}$)},
        ymin=0, ymax=14, % 显著调整 Ymax
        width=0.9\linewidth,
        % --- 图例 ---
        legend style={
            at={(0.5, -0.2)},
            anchor=north,
            legend columns=3,
            legend cell align=left,
        }
    ]
        % 占位图
        \addplot[draw=none, forget plot] coordinates {(0,0)};
        
        % 图例条目
        \addlegendimage{softred, mark=*, sharp plot, thick}
        \addlegendentry{Effective}
        
        \addlegendimage{softblue, mark=square*, sharp plot, thick}
        \addlegendentry{Ineffective}
        
        % 更新 "Ratio" 数据
        \addplot[softgreen, mark=triangle*, sharp plot, thick] coordinates {
            (0, 0.5049838774838994)
            (5, 3.0498675742468575)
            (10, 5.044440393830168)
            (15, 6.909238000911148)
            (20, 8.675892276479635)
            (25, 10.295623258208607)
            (30, 11.875869947572582)
            (35, 12.919927747871924)
        };
        \addlegendentry{Ratio}

    \end{axis}
\end{tikzpicture}
    \caption{
        The average gradient flow (left y-axis) and the ratio between the effective and the ineffective (right y-axis) under each layer of Qwen3-8b on Amazon Review.
    }
    \label{fig:grad_flow_qwen3_8b_amazon}
\end{figure}
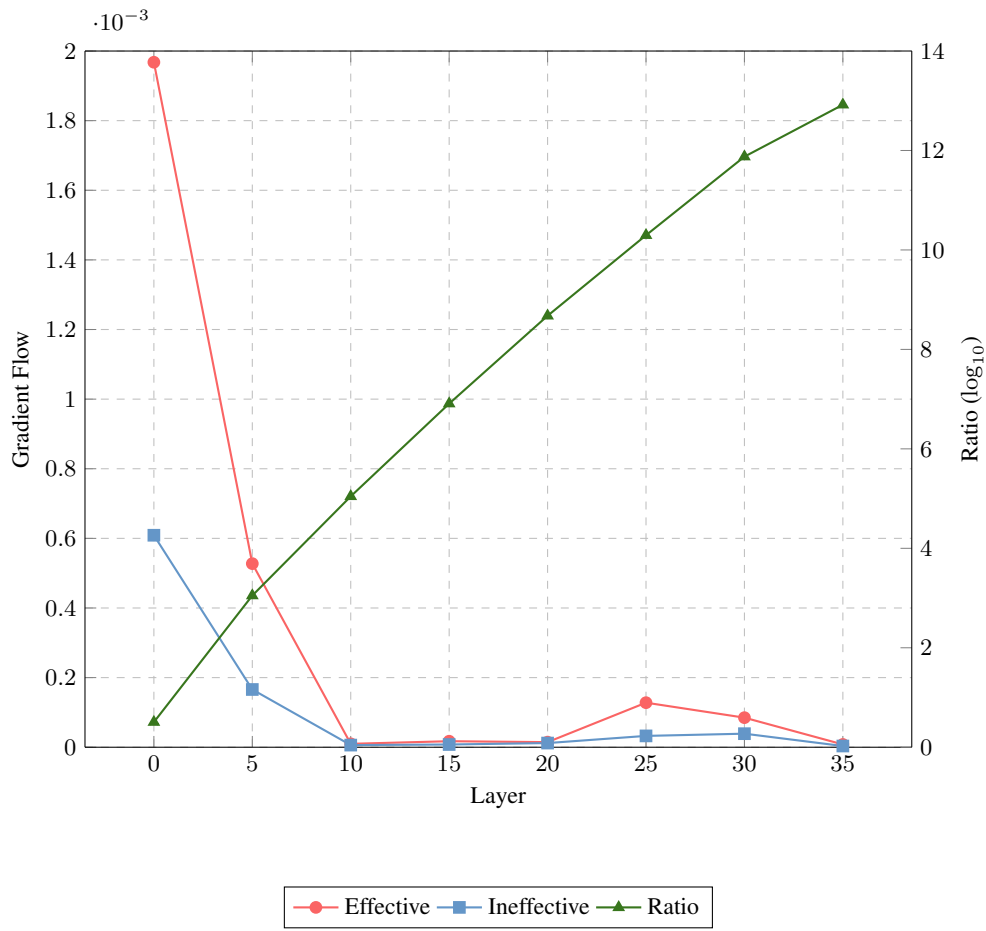

        In this part, we present the gradient flow under each setting from Figure~\ref{fig:grad_flow_llama2_7b_gsm8k} to Figure~\ref{fig:grad_flow_qwen3_8b_amazon}.

\end{document}